\documentclass{article} %
\ifdefined\pdfminorversion
\fi
\usepackage{iclr2026_conference,times}
\usepackage{needspace}

\usepackage{amsmath,amsfonts,bm}

\def\eqref#1{equation~\ref{#1}}

\def\1{\bm{1}}

\DeclareMathAlphabet{\mathsfit}{\encodingdefault}{\sfdefault}{m}{sl}
\SetMathAlphabet{\mathsfit}{bold}{\encodingdefault}{\sfdefault}{bx}{n}

\definecolor{hyperlinkblue}{RGB}{70,130,180}
\usepackage{url}
\usepackage{xspace}
\usepackage{booktabs}
\usepackage{graphicx}
\usepackage{float}
\usepackage{placeins}
\usepackage{multirow}
\usepackage{colortbl}
\usepackage{tabularx}
\usepackage{varwidth}
\usepackage[colorlinks=true,allcolors=hyperlinkblue]{hyperref}

\newcommand{\methodname}{Agima\xspace}
\newcommand{\modelname}{Agima-A7B\xspace}
\newcommand{\myparagraph}[1]{\par\noindent\textbf{#1}\ }

\title{Train an Imaginative Agent}

\author{%
\textbf{Jiaxin Ge}$^{1,*}$ \quad
\textbf{Junyi Zhang}$^{1,*}$ \quad
\textbf{Tianning Chai}$^{1,*}$ \quad
\textbf{Hanjun Yoo}$^{1}$ \\
\textbf{Chenyang Wang}$^{2}$ \quad
\textbf{Xiuyu Li}$^{2}$ \quad
\textbf{Weijia Shi}$^{3}$ \quad
\quad
\textbf{XuDong Wang}$^{1,4}$ \\
\textbf{Sewon Min}$^{1}$ \quad
\textbf{Luke Zettlemoyer}$^{3}$ \quad
\textbf{Trevor Darrell}$^{1,\dagger}$ \quad
\textbf{Haiwen Feng}$^{1,2,\dagger}$ \\[3pt]
\small $^{1}$University of California, Berkeley \quad $^{2}$Impossible Research \\
\small $^{3}$University of Washington \quad $^{4}$Duke University
}
\hypersetup{pdftitle={Train an Imaginative Agent},
pdfauthor={Jiaxin Ge, Junyi Zhang, Tianning Chai, Hanjun Yoo, Chenyang Wang, Xiuyu Li, Weijia Shi, XuDong Wang, Sewon Min, Luke Zettlemoyer, Trevor Darrell, Haiwen Feng}}
\iclrarxivcopy %

\begin{document}
\raggedbottom
\setlength{\floatsep}{8pt plus 2pt minus 2pt}
\setlength{\textfloatsep}{12pt plus 2pt minus 2pt}
\maketitle
\begingroup
\renewcommand{\thefootnote}{\ensuremath{*,\dagger}}
\footnotetext{: Equal contribution.}
\endgroup
\suppressfloats[t]
\begin{abstract}
A seemingly reasonable push in Sokoban can make the puzzle unsolvable. 
To avoid such mistakes, an agent has to anticipate the consequences of its actions before execution, especially when feedback from the real environment is sparse or actions are irreversible.
We introduce Agentic Imagination (Agima), a training framework that trains an agent to imagine before acting, i.e. to be imaginative. 
Agima trains a single omni model to propose actions, generate images of their outcomes, and then reason over these imagined futures to decide what to execute in the real environment. 
Rather than committing to a single plan, the agent can simulate a sequence of actions, explore alternatives, and compare possible futures.
We first use supervised fine-tuning (SFT) to unify policy and world modeling within a single omni model. This enables the agent to predict future outcomes and perform basic forms of imagination, such as forward thinking and counterfactual branching. 
We then use agentic reinforcement learning (RL) to teach the agent to reason with its imagined futures and use them effectively for planning.
Across four environments spanning Maze2D, Sokoban, Pacman, and PushT, Agima generalizes to harder settings unseen during training. 
Trained with SFT alone on two-box Sokoban, Agima solves 40\% of four-box puzzles while seeing the real board only once every five moves, compared with an action-only planner that only solves 2\%. %
After RL, new behaviors emerge in how Agima uses its imagination: it learns a new unified format that was unseen during the SFT stage, learns to conduct forward thinking only when necessary, and branches more frequently at critical decision points.
An agent that can picture where its actions lead can learn when and where to look ahead, that is, to plan. Project page: \url{https://agima.net/}.
\end{abstract}

\begin{figure}[h]
    \centering
    \includegraphics[width=1\linewidth,trim=20 11 20 14,clip]{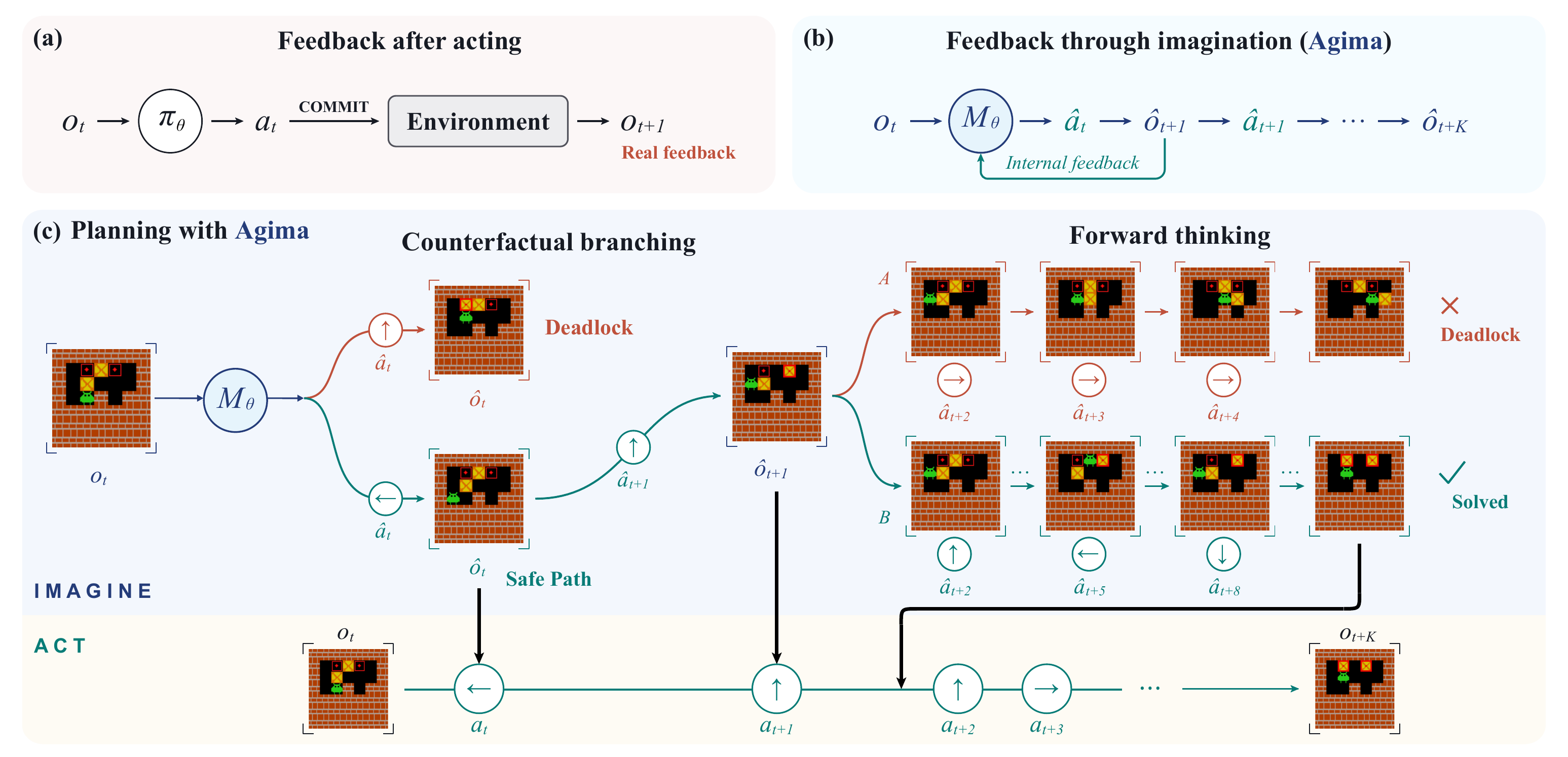}
    \vspace{-2em}
    \caption{
        \textbf{Planning with agentic imagination.}
        \textbf{(a)} A conventional agent receives feedback only after acting in the
        environment.
        \textbf{(b)} \methodname uses a single model to propose actions and predict the consequences of
        them. Each prediction serves as an internal observation for
        the next decision.
        \textbf{(c)} Planning with \methodname in Sokoban: counterfactual branching rejects
        a deadlocking action, and forward thinking compares candidate plans before
        execution. The executed actions reach $o_{t+K}$, which grounds the next round of
        imagination. In (c), $\hat{o}_j$ is the predicted outcome of $\hat{a}_j$. 
    }
    \label{fig:omniact}
\end{figure}

\section{Introduction}
Agents take actions and receive observations from the environment~\citep{sutton2018reinforcement, yao2023react}, but making good decisions often requires anticipating the consequences of actions several steps ahead.
When environment interactions are expensive or actions are irreversible, directly testing alternative actions through execution may not be feasible.
For example, in Sokoban, a seemingly reasonable push can block another box and make the puzzle unsolvable.
To avoid such failures, agents can predict and evaluate the consequences of possible actions before executing them~\citep{weber2017i2a}.

We introduce \emph{Agentic Imagination} (\methodname), a training framework that teaches an agent to imagine future observations of its actions and reason over them to plan before acting in the real environment.
We build \methodname on an omni model that unifies visual generation and language reasoning within the same multimodal context~\citep{zhou2024transfusion, deng2025bagel}.
The same model serves as both the policy model that proposes actions and as the world model that imagines their visual states.
These predictions become observations that the agent can use for further planning: it can extend a rollout, explore counterfactual actions, or compare longer candidate plans before execution (Figure~\ref{fig:omniact}).
Importantly, imagined and real observations share the same interface and can be used interchangeably as inputs to the model.
This allows the agent to use imagined observations in place of real environment feedback to continue planning, while real observations can replace imagined ones after execution to ground the next round of planning.

To teach the agent to imagine futures and plan with them, we construct training trajectories that interleave action proposals, visual predictions, and outcome evaluation.
This allows the agent to unify policy, world modeling, and reward modeling within a single omni model.
We also teach the agent basic ways to reason over its imagined futures for planning.
A \emph{linear rollout} follows a single action sequence, predicting the state after each step.
\emph{Counterfactual branching} explores alternative actions at individual steps and rejects those that lead to failures.
\emph{Forward thinking} imagines and compares longer candidate plans before deciding which one to execute.
Together, these trajectories teach the agent to generate imagined futures and use them to evaluate different actions before execution.
We use these trajectories for supervised fine-tuning (SFT). 

Once the agent can imagine futures, we want it to learn how to reason over them more effectively, rather than simply following predefined reasoning formats.
Recent progress in reinforcement learning (RL) for reasoning and tool use in language models~\citep{ICLR2025_871ac99f, jin2025searchr1trainingllmsreason} shows that models can learn new reasoning behaviors from outcome rewards.
We adapt this idea to \methodname using GRPO~\citep{shao2024deepseekmath}.
During RL, the agent imagines and evaluates candidate plans, executes its selected actions in the real environment, and receives a binary reward based on task success.
We optimize the agent's reasoning in language space and keeps its generated visual observations as context for planning.
Unlike SFT, we do not provide additional expert demonstrations or supervise how the agent should organize its imagination.
This allows the agent to internalize adaptive reasoning strategies.
It learns from task success when to explore another plan, which alternatives to consider, and when to proceed with its current plan.

We evaluate \methodname in Maze2D~\citep{wang2026visgym}, Sokoban and Pacman~\citep{weber2017i2a}, and PushT~\citep{florence2022ibc, chi2023diffusionpolicy}, training on simpler instances and testing on harder settings excluded from training.
With SFT alone, \methodname directly generalizes from two-box to four-box Sokoban without further training.
With real observations every five actions, it achieves 40\% success, compared with 2\% for an action-only planner fine-tuned from the same backbone.
Forward thinking further improves long-horizon planning in Pacman using the same training instances as linear rollouts.
Agentic RL further improves task success and leads to new behaviors in how \methodname uses its imagination.
For forward thinking, the agent learns a new reasoning format unseen during SFT: it first imagines a single plan and only explores another candidate when the first is predicted to fail.
For counterfactual branching, the agent explores more alternatives and increasingly focuses its branches on critical decisions where a wrong action could make the puzzle unsolvable.
These behaviors emerge from task-level rewards without additional supervision on how to organize its imagination.

Our contributions are threefold.
First, we introduce \methodname, a training framework for agentic imagination that teaches agents to explore, evaluate, and revise plans using predicted visual observations before acting.
Second, we show that agentic imagination generalizes to out-of-domain settings across four environments, with further gains from learning to compare imagined futures using the same underlying training instances.
Third, we show that agentic RL teaches \methodname to organize its own imagination through task-success rewards, eliciting adaptive reasoning strategies that determine when to explore another plan and where to examine alternative actions.

\section{Learning Agentic Imagination}
\label{sec:self_simulation}
\begin{figure*}[t]
\centering
\vspace{-2mm}
\includegraphics[width=\textwidth]{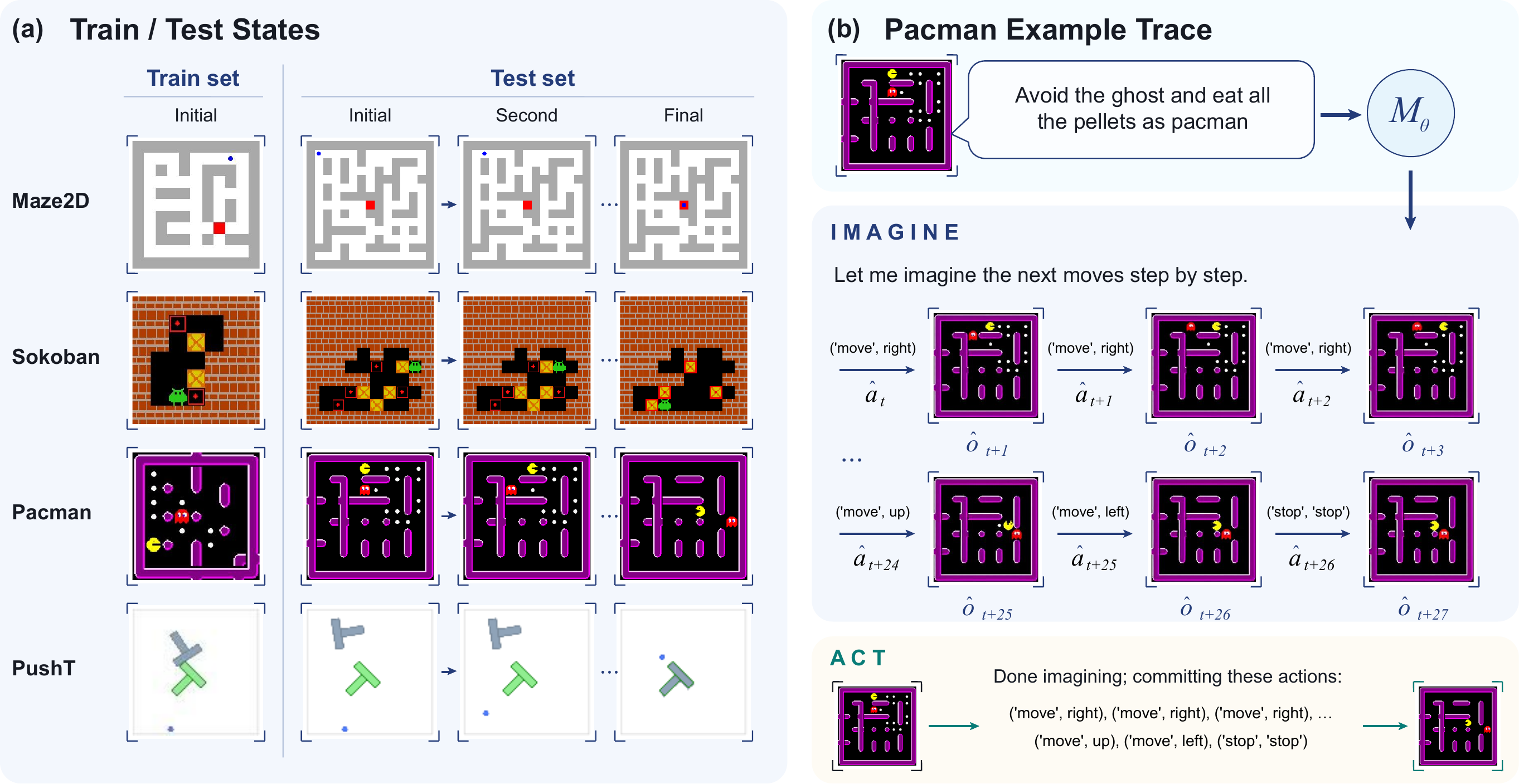}
\par
\vspace{-3mm}
\caption{\textbf{Evaluation environments and imagined visual rollouts.}
(a) Training and test states in four environments.
(b) \methodname interleaves actions and predicted states, then executes the actions.}
\vspace{-2mm}
\label{fig:transition_regimes}
\end{figure*}

Agentic imagination (\methodname) trains an agent to imagine the future observations of its actions and reason over these imagined futures before acting in the environment.
Given the current observation and a goal, the agent proposes an action, generates an image of its predicted outcome, and uses this imagined observation to continue planning.
We build \methodname on a single omni model that unifies language reasoning and visual generation.
This allows the same model to both propose actions and predict their consequences, serving as both the agent policy and the world model.
Our training framework has two stages.
First, supervised fine-tuning (SFT) teaches the agent to unify policy and world modeling within a single model, allowing the agent to imagine future observations and perform basic reasoning over its imagined futures through different planning traces.
Second, reinforcement learning (RL) teaches the agent to use its imagination more effectively for planning, learning when to explore additional futures and which alternatives to consider using only task-level rewards.

\subsection{Planning with Imagined Observations}
\label{sec:imagined_observations}

Consider an agent at time $t$ with visual observation $o_t$ and goal $g$.
Under action chunking~\citep{zhao2023aloha}, an action-only agent proposes a sequence of $K$ actions:
\[
A_t^{(K)}
=
\left(\hat{a}_t,\hat{a}_{t+1},\ldots,\hat{a}_{t+K-1}\right).
\]
Agentic imagination interleaves proposed actions with predicted future observations:
\[
\tau_t^{(K)}
=
\left(
\hat{a}_t,\hat{o}_{t+1},
\hat{a}_{t+1},\hat{o}_{t+2},
\ldots,
\hat{a}_{t+K-1},\hat{o}_{t+K}
\right).
\]
Each predicted state becomes an \emph{internal observation} for subsequent planning.
The agent can therefore propose its next action from imagined consequences without waiting for real feedback.
Once a plan is selected, its proposed actions are executed as $a_{t+j}=\hat{a}_{t+j}$.
The resulting real observation then serves as the starting point for the next round of imagination.
Beyond a single rollout, the agent can also explore alternative actions, return to previous plans, and compare different imagined futures before deciding what to execute.
We next describe how we train the agent to perform these different forms of imagination through supervised fine-tuning.

\subsection{Supervised Learning from Imagination Traces}
\label{sec:planning_formats}

We build \methodname on the BAGEL backbone~\citep{deng2025bagel}, an omni model that can understand and generate both images and text.
The backbone has 14B total parameters and 7B active parameters, denoted by A7B in our model names.
During SFT, we train on interleaved sequences of text and images such as the imagined rollout in Section~\ref{sec:imagined_observations}.
The current observation and goal are provided as input, and the remaining text and images are generated sequentially under teacher forcing.
We apply next-token cross-entropy to text outputs and the image-generation model's MSE objective to image outputs, combining them in the SFT loss:
\[
\mathcal{L}_{\mathrm{CE}}
=
-\sum_{i\in\mathrm{text}}
\log p_\theta(x_i\mid x_{<i}),
\quad
\mathcal{L}_{\mathrm{MSE}}
=
\sum_{j\in\mathrm{image}}
\left\|
\hat{v}_j-v_j
\right\|_2^2,
\quad
\mathcal{L}_{\mathrm{SFT}}
=
\mathcal{L}_{\mathrm{CE}}
+
\lambda_{\mathrm{img}}\mathcal{L}_{\mathrm{MSE}}.
\]
Training on these interleaved sequences teaches the model to both predict future observations and reason over them to decide subsequent actions.
We design three types of imagination traces that differ in how the agent explores alternative futures.
\begin{table*}[t]
\centering
\small
\setlength{\tabcolsep}{2pt}
\renewcommand{\arraystretch}{0.9}
\caption{
\textbf{Performance comparison across Pacman, Maze, Sokoban, and PushT environments.}
$K$ indicates the number of actions generated before feedback. 
Reactive ($K=1$) receives ground-truth feedback after every action.
Pacman, Maze, and Sokoban use the hard splits.
For PushT, we report success accuracy and final coverage on the \textit{far} split.
A7B denotes 7B active parameters (14B total); No Imagine is trained to predict actions without intermediate visual predictions.
Bold indicates the best result at each $K$ (including ties); gains are relative to \modelname \mbox{(No Imagine)}.
}
\vspace{0.5em}
\label{tab:combined_results_updated}
\begin{tabular}{@{}l l l c c c c c@{}}
\toprule
\multirow{2}{*}{\textbf{$K$}}
& \multirow{2}{*}{\textbf{Group}}
& \multirow{2}{*}{\textbf{Method}}
& \multirow{2}{*}{\textbf{Pacman}}
& \multirow{2}{*}{\textbf{Maze}}
& \multirow{2}{*}{\textbf{Sokoban}}
& \multicolumn{2}{c}{\textbf{PushT}} \\
\cmidrule(lr){7-8}
& & & & & & Acc. & Final Cov. \\
\midrule

\rowcolor{black!7}
\cellcolor{white}
&
& Gemini 3.1 Pro
& 0\%
& 38\%
& 3\%
& 0\%
& 0.2\% \\

\rowcolor{black!7}
\cellcolor{white}
& \multirow{-2}{*}{Closed}
& GPT-5.6 Sol
& 0\%
& 48\%
& 32\%
& 0\%
& 1.8\% \\

\multirow{-3}{*}{$1$}
& SFT
& \modelname
& \textbf{84\%}
& \textbf{62\%}
& \textbf{47\%}
& \textbf{42\%}
& \textbf{45.9\%} \\

\midrule
\rowcolor{black!7}
\cellcolor{white}\multirow{6}{*}{$3$}
&
& Gemini 3.1 Pro
& 0\%
& 21\%
& 4\%
& 0\%
& 2.2\% \\

\rowcolor{black!7}
\cellcolor{white}
& \multirow{-2}{*}{Closed}
& GPT-5.6 Sol
& 0\%
& 39\%
& 26\%
& 0\%
& 3.0\% \\

& \multirow{4}{*}{SFT}
& Qwen2.5-VL-7B
& 0\%
& 40\%
& 4\%
& 4\%
& 8.48\% \\

&
& Qwen3-VL-8B
& 0\%
& \textbf{85\%}
& 10\%
& 6\%
& 11.77\% \\

&
& \modelname \mbox{(No Imagine)}
& 25\%
& 54\%
& 10\%
& 12\%
& 18.6\% \\

&
& \modelname
& \textbf{38\%} \gain{+13\%}
& 59\% \gain{+5\%}
& \textbf{53\%} \gain{+43\%}
& \textbf{24\%} \gain{+12\%}
& \textbf{31.7\%} \gain{+13.1\%} \\

\midrule
\rowcolor{black!7}
\cellcolor{white}\multirow{6}{*}{$5$}
&
& Gemini 3.1 Pro
& 0\%
& 20\%
& 6\%
& 0\%
& 1.8\% \\

\rowcolor{black!7}
\cellcolor{white}
& \multirow{-2}{*}{Closed}
& GPT-5.6 Sol
& 1\%
& 34\%
& 20\%
& 1\%
& 8.3\% \\

& \multirow{4}{*}{SFT}
& Qwen2.5-VL-7B
& 0\%
& 28\%
& 0\%
& 2\%
& 4.07\% \\

&
& Qwen3-VL-8B
& 0\%
& 30\%
& 3\%
& 1\%
& 3.41\% \\

&
& \modelname \mbox{(No Imagine)}
& 1\%
& 23\%
& 2\%
& 1\%
& 6.7\% \\

&
& \modelname
& \textbf{37\%} \gain{+36\%}
& \textbf{40\%} \gain{+17\%}
& \textbf{40\%} \gain{+38\%}
& \textbf{18\%} \gain{+17\%}
& \textbf{28.5\%} \gain{+21.8\%} \\

\midrule
\rowcolor{black!7}
\cellcolor{white}\multirow{6}{*}{$10$}
&
& Gemini 3.1 Pro
& 0\%
& 15\%
& 4\%
& 0\%
& 2.0\% \\

\rowcolor{black!7}
\cellcolor{white}
& \multirow{-2}{*}{Closed}
& GPT-5.6 Sol
& 1\%
& \textbf{34\%}
& \textbf{23\%}
& 0\%
& 8.7\% \\

& \multirow{4}{*}{SFT}
& Qwen2.5-VL-7B
& 0\%
& 0\%
& 0\%
& 2\%
& 5.36\% \\

&
& Qwen3-VL-8B
& 0\%
& 0\%
& 0\%
& 0\%
& 5.55\% \\

&
& \modelname \mbox{(No Imagine)}
& 0\%
& 2\%
& 0\%
& 1\%
& 5.3\% \\

&
& \modelname
& \textbf{7\%} \gain{+7\%}
& 21\% \gain{+19\%}
& 15\% \gain{+15\%}
& \textbf{9\%} \gain{+8\%}
& \textbf{22.2\%} \gain{+16.9\%} \\

\midrule
\rowcolor{black!7}
\cellcolor{white}\multirow{6}{*}{$\infty$}
&
& Gemini 3.1 Pro
& 0\%
& 9\%
& 13\%
& 0\%
& 0.8\% \\

\rowcolor{black!7}
\cellcolor{white}
& \multirow{-2}{*}{Closed}
& GPT-5.6 Sol
& 2\%
& \textbf{36\%}
& \textbf{21\%}
& 0\%
& 5.5\% \\

& \multirow{4}{*}{SFT}
& Qwen2.5-VL-7B
& 0\%
& 0\%
& 0\%
& 2\%
& 5.60\% \\

&
& Qwen3-VL-8B
& 0\%
& 0\%
& 0\%
& \textbf{3\%}
& 5.87\% \\

&
& \modelname \mbox{(No Imagine)}
& 0\%
& 0\%
& 0\%
& \textbf{3\%}
& 6.2\% \\

&
& \modelname
& \textbf{6\%} \gain{+6\%}
& 3\% \gain{+3\%}
& 19\% \gain{+19\%}
& \textbf{3\%}
& \textbf{8.3\%} \gain{+2.1\%} \\

\bottomrule
\end{tabular}
\end{table*}

\begin{table}[t]
\centering
\renewcommand{\arraystretch}{0.9}
\small
\caption{
\textbf{Replacing imagined observations at inference.}
No Imagine is a separately trained action-only baseline; the four \modelname conditions share checkpoint, episodes, and decoding within each environment.
Blank replaces predictions with gray images, Stale with the latest real observation, and Oracle GT with ground-truth frames for the proposed actions.
}
\vspace{0.5em}
\label{tab:imagination_ablation}
\renewcommand{\arraystretch}{0.95}
\begin{tabular*}{\linewidth}{@{\extracolsep{\fill}}llcccc@{}}
\toprule
\textbf{Model} & \textbf{Condition} & \textbf{Sokoban} & \textbf{Pacman} & \textbf{Maze2D} & \textbf{PushT} \\
\midrule
No Imagine & --- & 2\% & 1\% & 23\% & 1\% \\
\midrule
\multirow{4}{*}{\modelname} & Own predictions & 40\% & 37\% & 40\% & 18\% \\
& Blank & 1\% & 0\% & 8\% & 1\% \\
& Stale & 0\% & 0\% & 8\% & 0\% \\
& Oracle GT & 43\% & 81\% & 69\% & 47\% \\
\bottomrule
\end{tabular*}
\end{table}

\myparagraph{Linear Rollout.}
A linear rollout alternates proposed actions and predicted observations:
\[
[o_t,g]
\rightarrow
\hat{a}_t
\rightarrow
\hat{o}_{t+1}
\rightarrow
\hat{a}_{t+1}
\rightarrow
\hat{o}_{t+2}
\rightarrow \cdots.
\]
Each imagined observation is fed back into the sequence before the next action is generated, allowing the model to continue planning without a new observation from the environment.

\myparagraph{Counterfactual Branching.}
A linear rollout commits to each action as it develops a plan.
Counterfactual branching instead checks local alternatives while maintaining a primary rollout.
At selected imagined states, the model tries a different action, predicts its immediate consequence, rejects the alternative if it is undesirable, and then continues along the main trajectory:
\[
[o_t,g]
\rightarrow
\hat{a}_t
\rightarrow
\hat{o}_{t+1}
\rightarrow
\begin{cases}
\hat{a}'_{t+1} \rightarrow \hat{o}'_{t+2} \rightarrow \text{reject},\\
\hat{a}_{t+1} \rightarrow \hat{o}_{t+2}
\rightarrow \cdots \rightarrow
\hat{o}_{j}
\rightarrow
\begin{cases}
\hat{a}'_j \rightarrow \hat{o}'_{j+1} \rightarrow \text{reject},\\
\hat{a}_j \rightarrow \hat{o}_{j+1} \rightarrow \cdots
\end{cases}
\end{cases}
\]
A single rollout may contain multiple such checks.
The main trajectory follows the current plan, while the counterfactual branches explore alternative actions that may lead to failure.

\myparagraph{Forward Thinking.}
While counterfactual branching explores local alternative actions at individual steps along a plan, forward thinking allows the agent to conduct long-horizontal imagination and compare different complete plans before deciding what to execute.
Each candidate is rolled forward over a longer horizon:
\[
[o_t,g]
\rightarrow
\begin{cases}
\hat{a}^{(1)}_t
\rightarrow
\hat{o}^{(1)}_{t+1}
\rightarrow
\hat{a}^{(1)}_{t+1}
\rightarrow
\hat{o}^{(1)}_{t+2}
\rightarrow \cdots,\\
\hat{a}^{(2)}_t
\rightarrow
\hat{o}^{(2)}_{t+1}
\rightarrow
\hat{a}^{(2)}_{t+1}
\rightarrow
\hat{o}^{(2)}_{t+2}
\rightarrow \cdots.
\end{cases}
\]
The model can therefore compare how different plans unfold before choosing which one to execute.
\subsection{Reinforcement Learning for Adaptive Imagination}
\label{sec:execution_rl}

SFT teaches the model how to generate and reason with imagination traces, but the demonstrations also specify which alternatives to explore and compare.
We use RL to optimize these choices directly from task success.
Starting from the SFT model, we apply GRPO~\citep{shao2024deepseekmath}.
For each problem, the model generates a complete planning trace and final executable actions without receiving additional observations from the real environment.
The selected actions are then executed, producing a binary task-success reward.
We obtain group-relative advantages by standardizing rewards within each group of sampled traces.
GRPO updates the language tokens in the planning trace; generated images remain in the multimodal context but receive no direct RL loss.
We use a group size of $8$ and batch size of $16$.
Additional optimization details are provided in Appendix~\ref{app:training}.

\section{Generalization and Planning with Imagination}
\label{sec:planning_evaluation}
We evaluate two capabilities enabled by agentic imagination.
First, we ask whether imagined observations can support multi-step planning
without frequent feedback from the environment, and whether this ability
generalizes beyond the configurations seen during training.
Second, we ask whether the agent can use imagined futures not only to extend
a rollout, but also to evaluate alternatives before acting.
We study these questions across four environments with qualitatively different
dynamics: Maze2D~\citep{wang2026visgym} for spatial navigation,
Sokoban~\citep{weber2017i2a} for irreversible object interactions,
Pacman~\citep{weber2017i2a} for reactive dynamics, and
PushT~\citep{florence2022ibc, chi2023diffusionpolicy} for continuous contact dynamics.
Figure~\ref{fig:transition_regimes} illustrates the environments and an example
imagined rollout.
Full environment details are provided in Appendix~\ref{app:task_desc}.

\subsection{\methodname Enables Easy-to-Hard Generalization}
\myparagraph{Setup.}
We first evaluate linear imagination rollouts.
For each environment, we train on simpler task configurations and evaluate on
harder or unseen configurations excluded from training
(Figure~\ref{fig:transition_regimes}(a)).
Let $K$ denote the number of actions planned and executed before the next
ground-truth observation is provided.
The fully reactive setting $K=1$ provides a reference with feedback after every action.
At $K=\infty$, no intermediate environment observation is provided during the rollout.
We denote the model trained with linear imagination rollouts as \modelname.
Our primary controlled baseline is \modelname \mbox{(No Imagine)}, which is separately fine-tuned
from the same BAGEL backbone on the same task data, but predicts action chunks
without intermediate visual states.
We also supervised fine-tune Qwen2.5-VL-7B~\citep{bai2025qwen25vl} and
Qwen3-VL-8B~\citep{bai2025qwen3vl} on the same task data as additional
action-only baselines.
Finally, we evaluate Gemini~3.1~Pro~\citep{deepmind2026gemini31pro} and
GPT-5.6~Sol~\citep{openai2026gpt56sol} as proprietary baselines.

\myparagraph{Imagination Transfers to Harder Settings and Improves Planning.}
Table~\ref{tab:combined_results_updated} shows that \methodname transfers from simpler training instances to harder and unseen settings across all four environments, especially when planning multiple actions without real feedback.
At $K=5$, \methodname improves over the action-only model by 36 points
in Pacman, 17 points in Maze2D, 38 points in Sokoban, and 17 points in PushT.
For example, after training on two-box Sokoban, \methodname reaches 40\% success
on four-box puzzles at $K=5$, compared with 2\% for the action-only model.
Performance eventually degrades for very long open-loop planning:
at $K=\infty$, both absolute performance and the advantage over action-only
planning shrink in several environments.

\myparagraph{\methodname Relies on the Content of Its Imagined Observations.}
Replacing imagined observations with blank or stale frames sharply reduces success across all four environments (Table~\ref{tab:imagination_ablation}).
Replacing predicted frames with ground-truth frames improves success by 29-44 percentage points in Pacman, Maze2D, and PushT, and by 3 points in Sokoban.
These results show that \methodname actually uses the content of its imagined observations for planning and benefits from more accurate predictions.

\subsection{Reasoning over Alternative Imagined Futures}
Linear rollouts use imagined observations to extend a single future.
We next test whether the same learned world model can support richer planning operations:
probing local counterfactual actions with branching and comparing long-horizon candidate futures with forward thinking.
\begin{figure*}[t]
\centering
\vspace{-2mm}
\includegraphics[width=\textwidth]{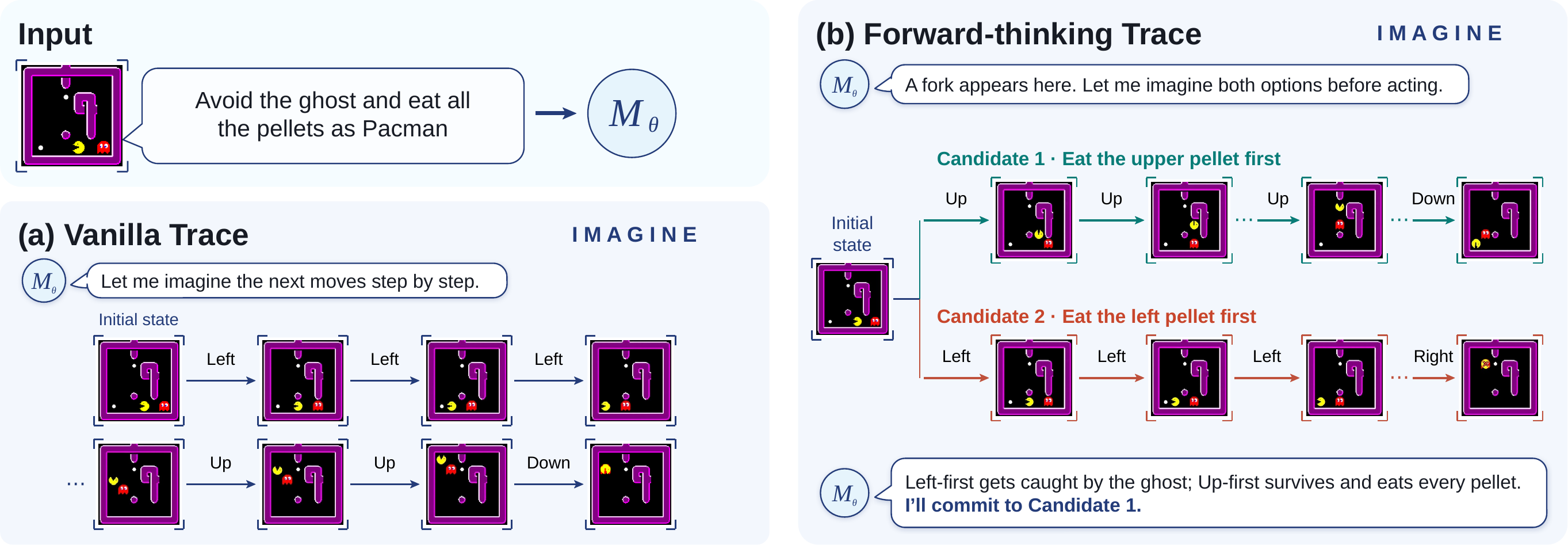}
\par
\vspace{-3mm}
\caption{\textbf{Forward thinking compares alternative futures before execution.}
A Pacman example (Table~\ref{tab:reasoning_results}(b)):
(a) \modelname is caught after executing a single rollout;
(b) \modelname \mbox{(Forward)} compares two imagined futures and selects Candidate~1, which succeeds.}
\label{fig:reasoning_demo}
\vspace{-1mm}
\end{figure*}
\myparagraph{Setup.}
All experiments use $K=\infty$, without intermediate observations from the real environment.
\modelname \mbox{(Branching)} and \modelname \mbox{(Forward)} use the same training instances as the linear-rollout model, differing only in the structure of their imagination traces.
The underlying training distribution remains unchanged.
In Sokoban, models trained on standard two-box puzzles are evaluated on two-, three-, and four-box \emph{Deadlock} and \emph{Structural Fork} sets, testing delayed consequences and critical action choices, respectively.
In Pacman, we evaluate route-order traps requiring long-horizon plan comparison on out-of-domain grid sizes.
Training traces and evaluation details are provided in Appendices~\ref{app:training} and~\ref{app:diagnostic_sets}.

\myparagraph{Counterfactual Branching Helps at Critical Decision Forks.}
Counterfactual branching allows the agent to explore alternative actions at each step along the main trajectory.
It is particularly helpful in the harder Structural Fork settings, where the agent encounters locally states that a seemingly reasonable action can lead to failure.
Compared with the linear-rollout model, branching improves three-box success from 34\% to 39\% and four-box success from 22\% to 32\% (Table~\ref{tab:reasoning_results}(a)).
These results show that \methodname can use imagined states to probe local alternatives that may lead to failure before actually committing to the next step.

\myparagraph{Forward Thinking Improves Long-Horizon Plan Selection.}
Forward thinking allows the agent to imagine and compare different futures before execution, helping it avoid failures that only become apparent several steps later.
On Sokoban Deadlock, forward thinking improves success from 22\% to 60\% on two-box puzzles and from 39\% to 48\% on three-box puzzles.
The effect is particularly consistent on Pacman route-order traps:
forward-thinking supervision improves success across every evaluated configuration, including from 22\% to 54\% on G8/F4 and from 5\% to 12\% on G10/F8 (Table~\ref{tab:reasoning_results}(b)).
Figure~\ref{fig:reasoning_demo} illustrates why long-horizon comparison matters.
With linear rollout, the model directly chooses the more tempting option: collecting the nearer food first, which appears safe because the ghost is still far away. However, when Pacman turns toward the remaining food, it can no longer escape the ghost and eventually dies.
With forward thinking, \methodname imagines alternative futures that collect the food items in different orders and discovers that collecting the farther food first leads to a successful outcome.
\begin{table*}[t]
\centering
\small
\setlength{\tabcolsep}{3pt}
\renewcommand{\arraystretch}{0.9}
\caption{\textbf{Success rates on Sokoban and Pacman reasoning tasks.}
(a) Sokoban Deadlock and Structural Fork configurations with 2, 3, and 4 boxes.
(b) Pacman route-order traps, where G denotes grid size, and F denotes the number of food items.
Bold indicates the best result in each column; changes in parentheses are absolute percentage-point differences relative to \modelname.}
\vspace{0.5em}
\label{tab:reasoning_results}
\label{tab:full_set_comparison}
\label{tab:pacman_forward_thinking}
\begin{tabularx}{\textwidth}{>{\raggedright\arraybackslash}p{11em} *{6}{>{\centering\arraybackslash}X}}
\toprule
\textbf{(a) Sokoban} & \multicolumn{3}{c}{\textbf{Deadlock}} & \multicolumn{3}{c}{\textbf{Structural Fork}} \\
\cmidrule(lr){2-4} \cmidrule(lr){5-7}
\textbf{Method} & \textbf{2-box} & \textbf{3-box} & \textbf{4-box} & \textbf{2-box} & \textbf{3-box} & \textbf{4-box} \\
\midrule
\rowcolor{black!7}
Gemini 3.1 Pro & \textbf{74\%} & 42\% & 19\% & 84\% & 34\% & 7\% \\
\rowcolor{black!7}
GPT-5.6 Sol & 44\% & 23\% & 17\% & 60\% & 20\% & 12\% \\
Qwen2.5-VL-7B & 1\% & 2\% & 0\% & 12\% & 2\% & 0\% \\
Qwen3-VL-8B & 6\% & 1\% & 1\% & 19\% & 2\% & 0\% \\
\modelname \mbox{(No Imagine)} & 12\% & 5\% & 1\% & 15\% & 2\% & 1\% \\
\modelname & 22\% & 39\% & 27\% & 87\% & 34\% & 22\% \\
\modelname \mbox{(Branching)} & 30\% \gain{+8\%} & 39\% \gain{+0\%} & \textbf{32\%} \gain{+5\%} & 84\% \drop{-3\%} & 39\% \gain{+5\%} & \textbf{32\%} \gain{+10\%} \\
\modelname \mbox{(Forward)} & 60\% \gain{+38\%} & \textbf{48\%} \gain{+9\%} & 24\% \drop{-3\%} & \textbf{93\%} \gain{+6\%} & \textbf{46\%} \gain{+12\%} & 22\% \gain{+0\%} \\
\end{tabularx}
\par\nointerlineskip
\begin{tabularx}{\textwidth}{>{\raggedright\arraybackslash}p{11em} *{5}{>{\centering\arraybackslash}X}}
\midrule
\textbf{(b) Pacman}
& \multicolumn{4}{c}{\textbf{G8}}
& \multicolumn{1}{c}{\textbf{G10}} \\
\cmidrule(lr){2-5} \cmidrule(lr){6-6}
\textbf{Method} & \textbf{F2} & \textbf{F4} & \textbf{F6} & \textbf{F8} & \textbf{F8} \\
\midrule

\rowcolor{black!7}
Gemini 3.1 Pro & 1\% & 1\% & 0\% & 1\% & 0\% \\
\rowcolor{black!7}
GPT-5.6 Sol & 16\% & 19\% & 6\% & 8\% & 0\% \\
Qwen2.5-VL-7B & 0\% & 0\% & 0\% & 0\% & 0\% \\
Qwen3-VL-8B & 0\% & 0\% & 0\% & 1\% & 0\% \\
\modelname \mbox{(No Imagine)} & 3\% & 0\% & 2\% & 0\% & 0\% \\
\modelname
& 7\% & 22\% & 16\% & 16\% & 5\% \\
\modelname \mbox{(Forward)}
& \textbf{40\%} \gain{+33\%}
& \textbf{54\%} \gain{+32\%}
& \textbf{41\%} \gain{+25\%}
& \textbf{37\%} \gain{+21\%}
& \textbf{12\%} \gain{+7\%} \\
\bottomrule
\end{tabularx}
\end{table*}

\section{Adaptive Reasoning through Agentic RL}
\label{sec:adaptive_self_simulation}

SFT teaches \methodname to construct and reason over imagined futures, but the training traces follow predefined heuristics.
Since \methodname combines language-space reasoning with visual prediction, we can apply agentic RL to improve how the agent uses its imagination for planning. 
Specifically, we ask whether task-level rewards can teach the agent when to think ahead, when to reconsider a plan, and where to explore counterfactual alternatives.

\begin{figure}[t]
    \centering
    \includegraphics[width=0.9\linewidth]{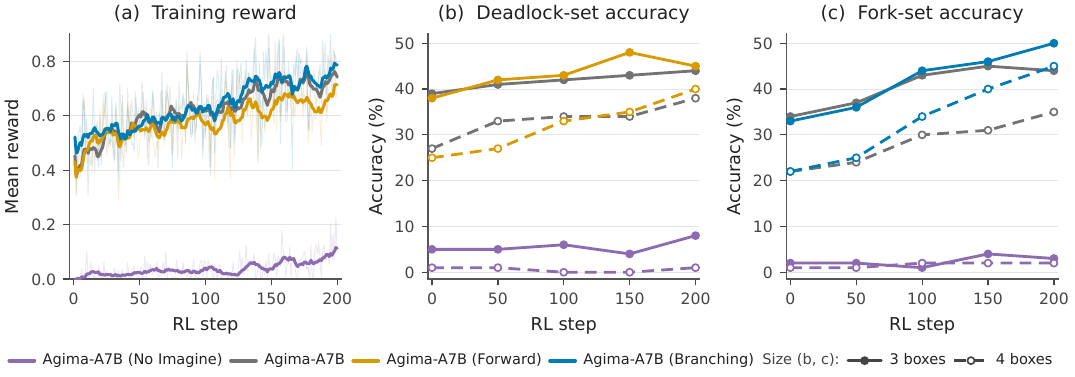}
    \vspace{-1em}
    \caption{\textbf{RL improves training reward and held-out planning performance.}
    (a) Training reward during RL.
    (b--c) Accuracy on the Deadlock and Structural Fork evaluation sets.
    Solid and dashed lines denote three-box and four-box tasks, respectively.}
    \label{fig:rl_training_curve}
\end{figure}

\myparagraph{Setup.}
We initialize RL from four SFT checkpoints trained on two-box Sokoban:
\modelname{} (No Imagine), \modelname, \modelname{} (Forward), and \modelname{} (Branching).
\modelname{} (No Imagine) provides a no-imagination baseline, generating actions without intermediate visual predictions.
For the latter two, we use mixed-format checkpoints trained on random mixtures of $\sim$50\% linear rollouts and $\sim$50\% of the corresponding reasoning traces.
This keeps their pre-RL performance close to the linear-rollout checkpoint while letting the model being able to generate multiple reasoning formats, leaving RL to learn when and how to use them.
Because ordinary two-box Sokoban is nearly saturated after SFT, we perform RL on ordinary three-box puzzles.
The agent receives only a binary task-success reward, with no additional expert demonstrations or trace-level supervision.
We train for 200 RL steps using GRPO, and evaluate on held-out three- and four-box Deadlock and Structural Fork sets.
Additional details are provided in Appendix~\ref{app:training_splits}.

\myparagraph{RL Improves Planning Beyond SFT.}
Figure~\ref{fig:rl_training_curve} shows that RL improves training reward for all four initializations.
Even \modelname{} (No Imagine) rises from near-zero reward to roughly 15\% by the end of training, showing that task reward can improve the action policy.
However, this gain transfers little to the held-out Deadlock and Structural Fork sets, where its accuracy remains near zero.
In contrast, all three imagination-based initializations improve substantially on both training reward and held-out planning performance, including the four-box settings never seen during RL.
The gain is particularly strong for \modelname{} (Branching): despite starting near the linear-rollout \modelname checkpoint, RL eventually pushes it beyond \modelname on Structural Fork.

\begin{figure}[!t]
    \centering
    \includegraphics[width=0.90\linewidth]{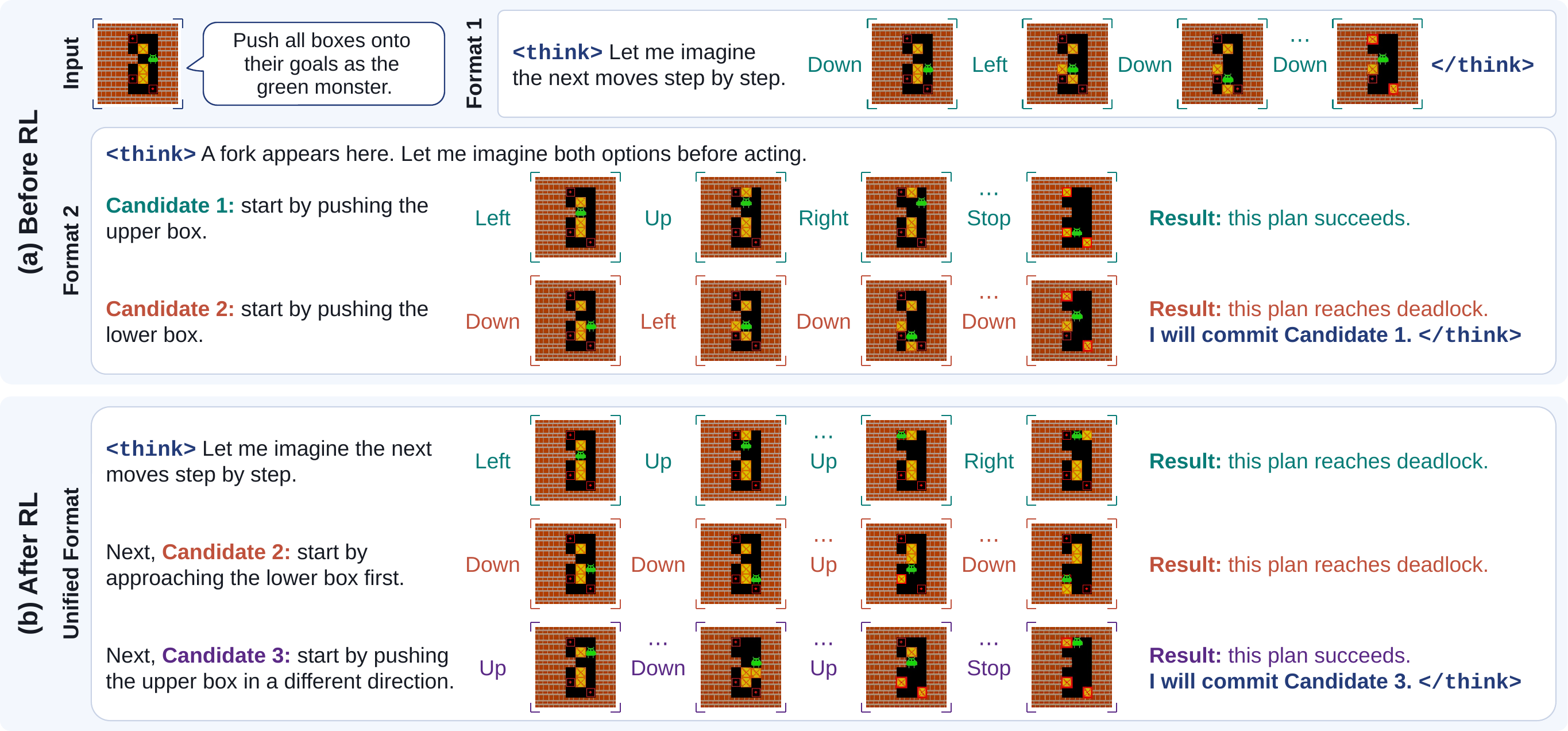}
    \par
    \vspace{-3mm}
\caption{\textbf{RL induces a unified, adaptive planning format.}
(a) SFT produces a linear rollout or a fixed two-candidate comparison.
(b) After RL, the model evaluates each imagined plan and generates alternatives as needed, here rejecting two failed plans before finding a successful third.}
    \vspace{-1mm}
    \label{fig:rl_qualitative}
\end{figure}

\begin{figure}[t]
    \centering
    \includegraphics[width=\linewidth]{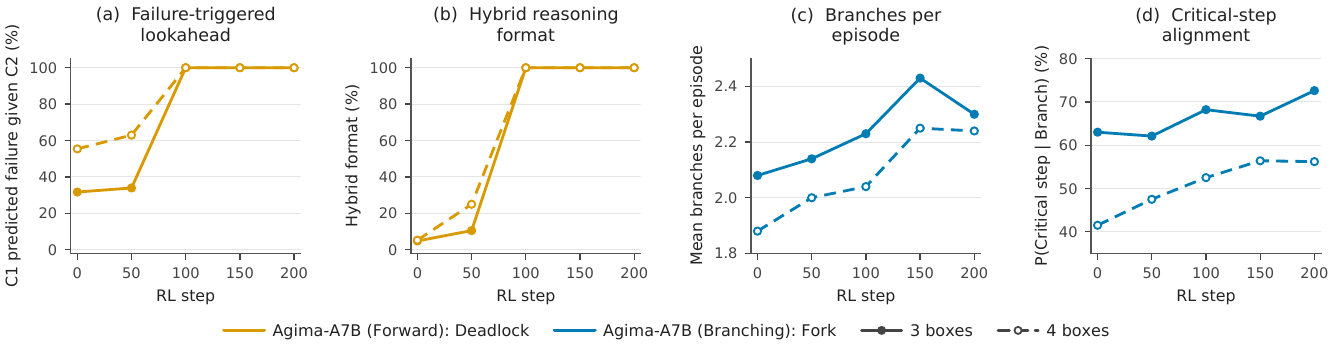}
    \vspace{-2.em}
    \caption{\textbf{RL induces adaptive use of imagination.}
    Forward thinking becomes increasingly failure-triggered, while branching explores more alternatives and shifts toward critical decision points.}
    \label{fig:example}
\end{figure}

\myparagraph{Forward Thinking Becomes Adaptive.}
RL teaches \methodname when to explore alternative plans in forward thinking.
During SFT, the training data only contains two predefined reasoning formats: (1) a linear rollout, or (2) a forward thinking trace that decides from the beginning to compare two candidates.
After 100 steps of RL training, a new format emerges that combines these two behaviors: the agent first rolls out a plan and evaluates its predicted outcome. If the plan is predicted to succeed, it ends the reasoning trace and proceeds with the plan; otherwise, it generates another candidate (Figure~\ref{fig:rl_qualitative}, Appendix~\ref{app:forward_gating}).
Meanwhile, the model also learns when additional exploration is necessary. When a second candidate is generated, the first plan fails in 100\% of cases, compared to 32\% before RL.
This suggests that the agent learns to reason over its predicted outcomes and only uses forward thinking when needed, rather than simply following the predefined formats seen during SFT.
The agent also learns to explore beyond the two-candidate format seen during SFT which is trained on two-box puzzles. As shown in Figure~\ref{fig:rl_qualitative}, in the three-box setting, it learns to explore a third candidate after the first two plans fail.

\myparagraph{Branching Becomes More Exploratory and Targeted.}
RL also changes the branching behavior of \methodname.
The model explores more alternatives once branching begins, with the average number of branches per trace increasing from 1.70 before RL to 1.95 after 150 steps of RL training.
We define \emph{Critical Branch Rate} as the fraction of branches opened at critical states, where at least one alternative action would make the puzzle unsolvable.
As shown in Figure~\ref{fig:example}, this rate increases after RL, indicating that counterfactual reasoning becomes more concentrated on decisions where the choice of action matters.
Meanwhile, the total number of imagined frames does not increase (Appendix~\ref{app:rollout_length}), suggesting that RL teaches the agent to use branching more effectively to avoid potential failures without increasing its overall imagination budget.
\FloatBarrier

\section{Related Work}

\myparagraph{Model-based planning.}
Model-based reinforcement learning learns a model of the environment dynamics and uses it to plan
or to train a policy \citep{sutton1991dyna, ha2018worldmodels, hafner2019planet, hafner2020dreamer,
schrittwieser2020muzero}.
I2A \citep{weber2017i2a} is particularly relevant here: it uses learned environment rollouts to
inform a model-free policy on Sokoban and MiniPacman.
These tasks share the irreversible decisions and long planning horizons of our setting.
Language models have also been combined with tree search, where they act as the policy, the world
model, or the evaluator \citep{hao2023rap, yao2023tot}, and recent video world models generate
high-fidelity future frames conditioned on actions \citep{bruce2024genie, deepmind2025genie3,
valevski2024gamengen, alonso2024diamond, yang2024unisim, agarwal2025cosmos, bar2025nwm}.
I2A leaves the choice of imagined actions to a rollout policy, while the tree-search methods use
an external search procedure.
In \methodname, action proposals and visual predictions come from the same model.
Each prediction stays in the context for subsequent reasoning, so choosing whether to extend a
rollout, try an alternative, or reject a plan becomes part of the generated trace.

\myparagraph{Reasoning with generated images.}
Several recent methods include images in the reasoning process of multimodal models.
Visual Sketchpad \citep{hu2024sketchpad} calls tools to draw auxiliary lines and boxes, MVoT
\citep{li2025mvot} generates images of intermediate states in spatial reasoning, and ThinkMorph
\citep{gu2025thinkmorph} alternates between text and image thoughts.
Other work keeps the visual thought in a latent space \citep{zhang2025latentsketchpad}, and
\citet{su2025thinkingwithimages} provide a survey.
In these methods, the image helps the model reach a final answer, and there is usually no single
correct image to generate.
UniWM \citep{dong2025uniwm} and WorldVLA \citep{cen2025worldvla} jointly predict actions and future
images for navigation and manipulation, but they follow a single predicted trajectory and do not use
the predicted images to compare alternative actions.
In our work, each generated image is the predicted result of a specific action, and the model learns
to carry out the planning steps itself, including counterfactual branching and forward thinking
(Section~\ref{sec:self_simulation}).

\myparagraph{Reinforcement learning for VLM agents.}
Reinforcement learning with outcome rewards has improved the reasoning of language models
\citep{shao2024deepseekmath, guo2025deepseekr1} and has been applied to multi-turn LLM agents
\citep{wang2025ragen, feng2025gigpo} and to VLM agents that act from visual observations
\citep{zhai2024rl4vlm}.
Two works are closest to ours.
VAGEN \citep{wang2025vagen} trains a VLM agent to describe the current state and predict the next
state in its textual reasoning, and rewards the accuracy of these predictions; Sokoban is among its
tasks.
Visual Planning \citep{xu2026visualplanning} represents a plan as a sequence of generated images and
rewards each generated state.
Neither method trains an agent to reason in language about its own visual predictions.
We apply RL with a binary task-success reward to an agent of this kind.
Only the text tokens are optimized.
These tokens carry the agent's planning decisions, such as whether to imagine another plan and where
to branch, while the generated images remain in the context as observations.

\Needspace{10\baselineskip}
\section{Conclusion}

We present \methodname, a framework that teaches agents to plan with predicted visual observations.
Across four environments, \methodname generalizes beyond training configurations and supports multi-step planning without intermediate feedback, counterfactual branching, and comparison of candidate futures.
With agentic RL, \methodname learns new behaviors, including a new reasoning format, conditioning forward thinking only on failure, and focusing branching on critical decisions.
Our results show that a single omni model can be trained as an imaginaive agent that can imagine possible futures and reason over them to guide its own actions.

\section*{Acknowledgment}
We thank Angjoo Kanazawa, Ranjay Krishna, Joey Gonzalez, Rulin Shao, Zixian Ma, Xiaochuang Han, and Baifeng Shi for feedback on our manuscript. 

\bibliography{iclr2026_conference}
\bibliographystyle{iclr2026_conference}
\setlength{\floatsep}{12pt plus 2pt minus 2pt}
\setlength{\textfloatsep}{20pt plus 2pt minus 4pt}
\clearpage
\appendix
\raggedbottom
\input{sections/appendix/contents}
\clearpage
\section{Environments and Evaluation Sets}
\label{app:task_desc}

The training and test configurations below are used for the easy-to-hard
evaluation in Table~\ref{tab:combined_results_updated}. All grid sizes include
the outer boundary walls.
Maze2D, Sokoban, and Pacman use four movement actions: up, down, left, and
right. PushT uses continuous $(x,y)$ pusher targets. Training traces end with
\texttt{STOP}. We report episode success rates for the three discrete tasks,
and both success rate and mean final coverage for PushT.

\myparagraph{Maze2D: Constrained Navigation.}\label{app:maze2d}
Maze2D is a grid navigation task from VisGym~\citep{wang2026visgym}. The input
image shows the agent, walls, and goal. The model must navigate to the goal
while predicting its position after each move; an episode succeeds when the
agent reaches the goal. We train on $11\times11$ mazes with $9\times9$
interiors and test on larger mazes.

\myparagraph{Sokoban: Irreversible Object Dynamics.}\label{app:sokoban}
Sokoban is a box-pushing puzzle~\citep{weber2017i2a}. The input image shows the
agent, boxes, walls, and targets. Movement into a box pushes it when the next
cell is free. A wrong push can permanently trap a box or create a deadlock,
so the model must predict how actions change the box configuration. Success
requires placing all boxes on targets. For Table~\ref{tab:combined_results_updated},
we train on $7\times7$ two-box puzzles and test on four-box puzzles.

\myparagraph{Pacman: Reactive Dynamics.}\label{app:pacman}
Pacman adds a moving opponent, as in prior work on MiniPacman~\citep{weber2017i2a}.
The input image shows Pacman, the red ghost, walls, and food pellets. After
each Pacman move, the ghost moves by greedily chasing Pacman. Success requires
collecting all food without being caught. We train on $9\times9$ grids with
8 food pellets. This training configuration is shared by the linear-rollout
and forward-thinking data. We test on $12\times12$ grids with 12 food pellets
and a closer initial ghost position.

\myparagraph{PushT: Continuous Contact Dynamics.}\label{app:pusht}
PushT is a continuous manipulation task~\citep{florence2022ibc, chi2023diffusionpolicy}.
The input image shows the pusher, a T-shaped object, and its target pose.
The model predicts pusher targets and the resulting object motion. Training
uses initial object positions nearer the center; evaluation uses unseen
positions farther from the center. Final coverage measures the fraction of
the target covered by the object at the end of execution. We report its mean
across episodes and the success rate, defined as the fraction of episodes
with final coverage greater than $0.80$.

\subsection{Planning Evaluation Sets}
\label{app:diagnostic_sets}
Table~\ref{tab:reasoning_results} evaluates counterfactual branching and
forward thinking on the Sokoban and Pacman sets below. These sets test
decisions whose consequences become apparent only after several actions,
separately from the easy-to-hard evaluation in
Table~\ref{tab:combined_results_updated}. We report task success rates with
$K=\infty$, without intermediate observations from the real environment.

\myparagraph{Sokoban Deadlock and Structural Fork.}
Each family contains 2-box, 3-box, and 4-box evaluation sets, with 100 episodes
per set. Deadlock instances contain locally legal pushes that can make the
puzzle unsolvable, with harmful consequences becoming apparent later in the
trajectory. Structural Fork instances offer multiple legal choices,
some of which lead to dead ends. The supervised checkpoints are trained on
standard 2-box puzzles; the 3-box and 4-box sets test transfer to larger
object configurations.

\myparagraph{Pacman route-order traps.}
We evaluate on out-of-domain grids of size $8\times8$ with 2, 4, 6, or 8
food items and $10\times10$ with 8 food items.
Both grid sizes differ from the $9\times9$ training grids.
In these instances, pursuing the nearer food first leaves no successful
continuation, while a farther-first route has a replay-verified solution
that clears the food, avoids the ghost, and ends with \texttt{STOP}.

We denote a configuration as G$x$/F$y$, where $x$ is the grid size and $y$
is the number of food items; for example, G8/F4 denotes an $8\times8$ grid
with four food items.

\clearpage
\section{Training Data and Optimization}
\label{app:training}
We first describe the supervised training traces and their statistics, then
give the model and optimization settings separately.

\subsection{Action Chunking and Linear Rollout Data}
\label{app:linear_training}
We segment expert trajectories into action chunks with
$K\in\{1,3,5,10,\infty\}$. Each example starts from an observation $o_t$ and
goal $g$. The action-only target contains the next $K$ expert actions; the
linear-imagination target interleaves each action with its successor image.
The images used as training targets are observations from the expert
trajectory. At inference time, the model generates these images as predicted
observations and uses them to plan subsequent actions.

For example, a $K=3$ training segment has the following schematic form:
\[
\begin{aligned}
\text{Input:}\quad & [o_t,g],\\
\text{Action-only target:}\quad & [a_t,a_{t+1},a_{t+2}],\\
\text{Linear-imagination target:}\quad &
[a_t,o_{t+1},a_{t+1},o_{t+2},a_{t+2},o_{t+3}].
\end{aligned}
\]
After these three actions are executed, the next chunk starts from the new
real observation. $K=1$ gives feedback after every action, while $K=\infty$
uses the full remaining plan without intermediate real observations.
The final segment ends with \texttt{STOP} when the expert task is complete.
Figure~\ref{fig:action_chunk_examples} shows a qualitative example of different action chunk sizes.

\subsection{Constructing Alternative-Future Traces}
\label{app:trace_construction}
Counterfactual-branching and forward-thinking traces use the same underlying
expert episodes as linear rollouts. We do not select additional training
instances specifically for failures or difficult forks; only the imagination
trace structure changes. The additional supervision teaches the model how
to explore and compare imagined futures.
Figures~\ref{fig:sokoban_train_examples}, \ref{fig:sokoban_forward_train}, and \ref{fig:pacman_train_examples} show examples of the three constructions below.

\myparagraph{Counterfactual branching.}
For Sokoban, expert trajectories are scanned for critical states where the
expert push keeps the level solvable and another legal push is certified
unsolvable by moving a box against a wall. At up to three earliest such states,
a trace inserts an alternative push, its successor image, a rejection, and a
return to the expert continuation. The executable sequence after
\texttt{</think>} remains the original expert solution. Episodes without a
certified harmful alternative remain single-plan rollouts.

\myparagraph{Forward thinking in Sokoban.}
Each trace compares two candidate box-pushing plans from the same initial
state. One follows the expert solution; the other starts with a different
first push and uses the best continuation found by search. Breadth-first
search (BFS) labels each candidate as solved, unfinished at the action limit,
or deadlocked. The target demonstrates
both futures and selects a successful plan, preferring the shorter plan when
both succeed. Candidate order is randomized.

\myparagraph{Forward thinking in Pacman.}
Each trace begins with a shared prefix of the expert trajectory, reaching
a decision state 2 to 10 moves into the episode. It then compares two
continuations from that state: the expert continuation and an alternative
that takes a different first move toward a nearer food item. The trace
returns to the same decision state between candidates, compares their
outcomes, and commits a complete plan. Candidate order is randomized.
Figure~\ref{fig:pacman_train_examples} shows an example in which the
alternative is caught by the ghost. The outcome proportions across all
training traces are given in Table~\ref{tab:train_data_reasoning}.

\subsection{Training Data Examples and Statistics}
\label{app:training_data_stats}
Table~\ref{tab:train_data_overview} summarizes the expert rollouts used to
construct action chunks. Table~\ref{tab:train_data_reasoning} reports the
alternative-future trace statistics. The Sokoban forward-thinking traces
used for training contain two candidates; their outcome proportions are
reported within that population. PushT expert trajectories are generated
with PPO~\citep{schulman2017ppo}. Training budgets are given separately below.

\begin{table}[H]
\centering
\small
\caption{
\textbf{Linear-rollout training data per task.}
Plan length counts the expert's actions excluding the final \texttt{STOP}
and is reported as median (mean) [range]. Grid sizes include the outer wall.
}
\label{tab:train_data_overview}
\setlength{\tabcolsep}{4pt}
\renewcommand{\arraystretch}{1.05}
\begin{tabularx}{\linewidth}{@{}l*{4}{>{\centering\arraybackslash}X}@{}}
\toprule
& \textbf{Maze2D} & \textbf{Sokoban} & \textbf{Pacman} & \textbf{PushT} \\
\midrule
Grid size & $11\times11$ & $7\times7$ & $9\times9$ & N/A \\
Training setting & maze navigation & 2 boxes & 8 food & diverse T start \\
Expert & shortest-path oracle & Sokoban solver & oracle planner & PPO policy \\
Action space & 4 moves & 4 moves & 4 moves & continuous $(x,y)$ \\
Action budget & 40 & 20 & 80 & 30 \\
Plan length & 9 (9.9) [5--33] & 10 (9.9) [2--19] & 23 (22.5) [9--27] & 20 (20.0) [4--29] \\
\bottomrule
\end{tabularx}
\end{table}

\begin{table}[htbp]
\centering
\small
\caption{
\textbf{Forward-Thinking and Counterfactual-Branching trace statistics.}
Forward-thinking outcome percentages are over two-candidate traces.
}
\label{tab:train_data_reasoning}
\definecolor{tracetableblue}{RGB}{240,245,251}
\setlength{\tabcolsep}{7pt}
\renewcommand{\arraystretch}{1.15}
\begin{tabularx}{\linewidth}{@{}Xr@{}}
\toprule
\rowcolor{tracetableblue}
\multicolumn{2}{l}{\textbf{Sokoban Forward Thinking}} \\
\midrule
Both solve & 79\% \\
At least one candidate deadlocks & 21\% \\
\midrule
\rowcolor{tracetableblue}
\multicolumn{2}{l}{\textbf{Sokoban Counterfactual Branching}} \\
\midrule
Branches per episode: 0 / 1 / 2 / 3 & 27.1 / 35.3 / 20.6 / 17.1\% \\
\midrule
\rowcolor{tracetableblue}
\multicolumn{2}{l}{\textbf{Pacman Forward Thinking}} \\
\midrule
Both candidates succeed & 96.4\% \\
A candidate dies & 3.6\% \\
\bottomrule
\end{tabularx}
\end{table}

\begingroup
\setlength{\intextsep}{3pt plus 1pt minus 1pt}
\setlength{\abovecaptionskip}{4pt}
\begin{figure}[H]
    \centering
    \includegraphics[page=1,trim=0 904bp 0 0bp,clip,width=\linewidth]{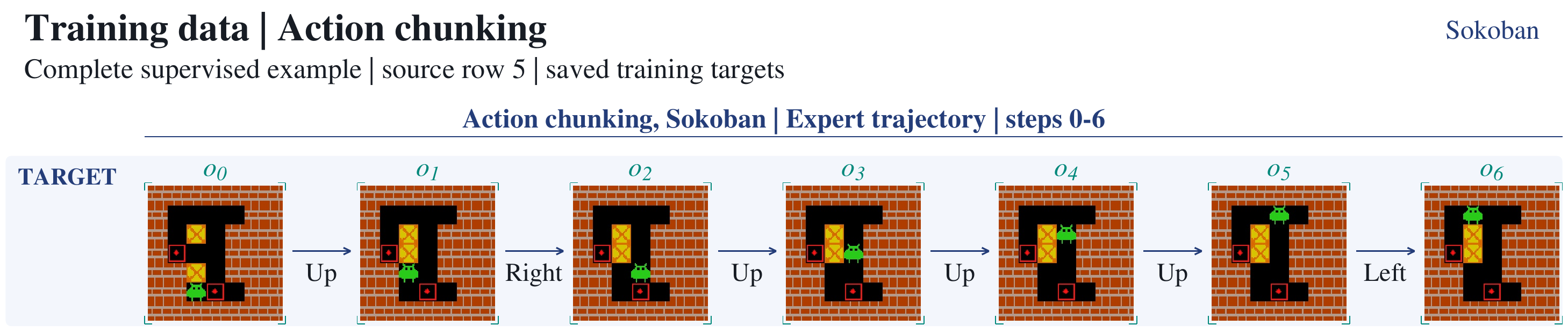}
\end{figure}

\begin{figure}[H]
    \centering
    \includegraphics[page=1,trim=0 696bp 0 296bp,clip,width=\linewidth]{figures/appendix_training_examples.pdf}
\end{figure}

\begin{figure}[H]
    \centering
    \includegraphics[page=1,trim=0 488bp 0 504bp,clip,width=\linewidth]{figures/appendix_training_examples.pdf}
\end{figure}

\begin{figure}[H]
    \centering
    \includegraphics[page=1,trim=0 0bp 0 712bp,clip,width=\linewidth]{figures/appendix_training_examples.pdf}
    \caption{\textbf{A complete expert trajectory and its action chunks.} The same 17-action training trajectory, including \texttt{STOP}, is segmented with $K\in\{1,3,5,10,\infty\}$. Numbers in the lower panel refer to the actions above. Images are saved environment observations used as supervision.}
    \label{fig:action_chunk_examples}
\end{figure}

\begin{figure}[H]
    \centering
    \includegraphics[page=2,trim=0 516bp 0 0bp,clip,width=\linewidth]{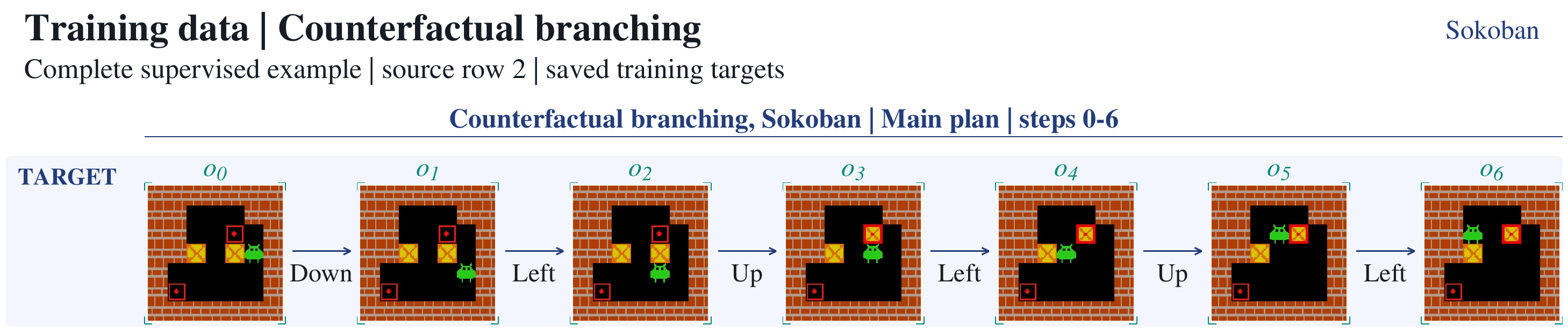}
\end{figure}

\begin{figure}[H]
    \centering
    \includegraphics[page=2,trim=0 308bp 0 296bp,clip,width=\linewidth]{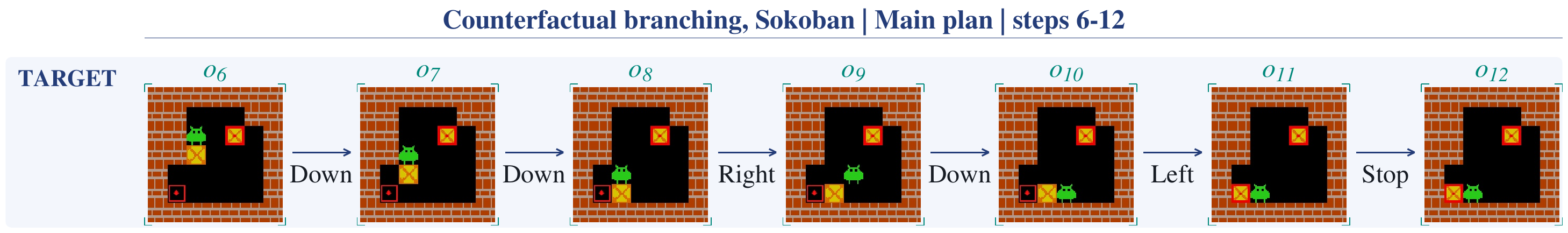}
\end{figure}

\begin{figure}[H]
    \centering
    \includegraphics[page=2,trim=0 0bp 0 504bp,clip,width=\linewidth]{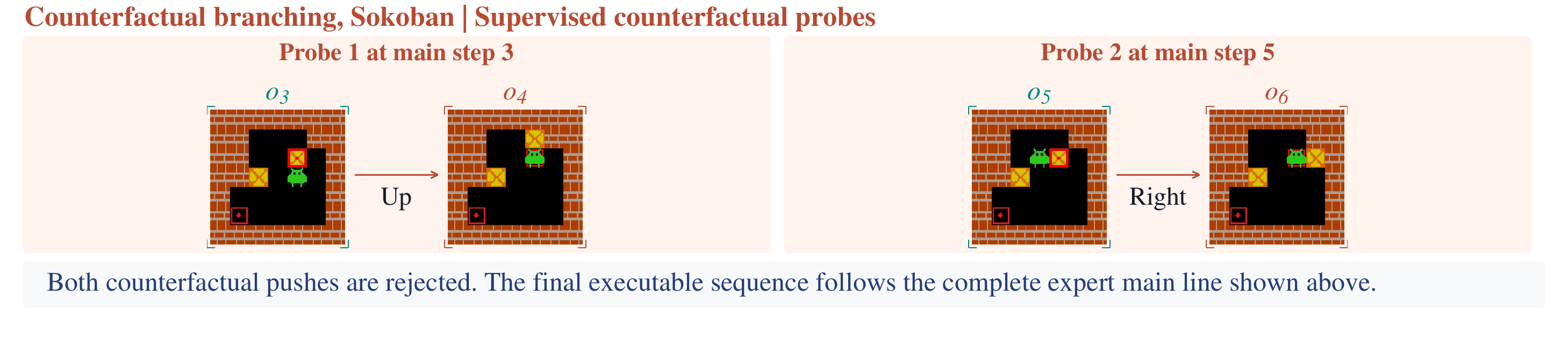}
    \caption{\textbf{Complete counterfactual-branching training example.} The main line contains 12 actions including \texttt{STOP}. The two alternative pushes, inserted after main-line actions 3 and 5, lead to certified deadlocks and are rejected.}
    \label{fig:sokoban_train_examples}
\end{figure}

\begin{figure}[H]
    \centering
    \includegraphics[page=3,trim=0 336bp 0 0bp,clip,width=\linewidth]{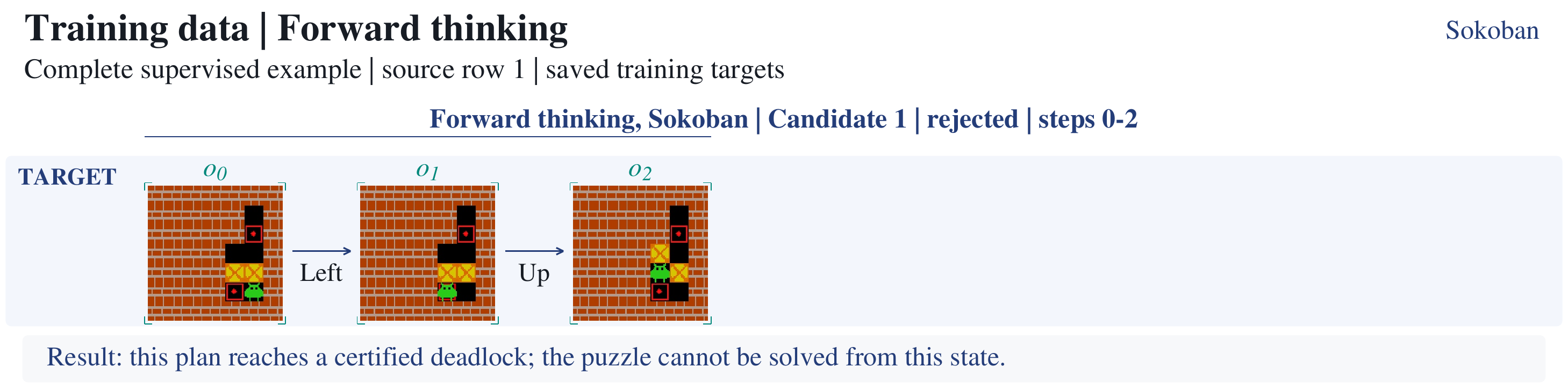}
\end{figure}

\begin{figure}[H]
    \centering
    \includegraphics[page=3,trim=0 0bp 0 346bp,clip,width=\linewidth]{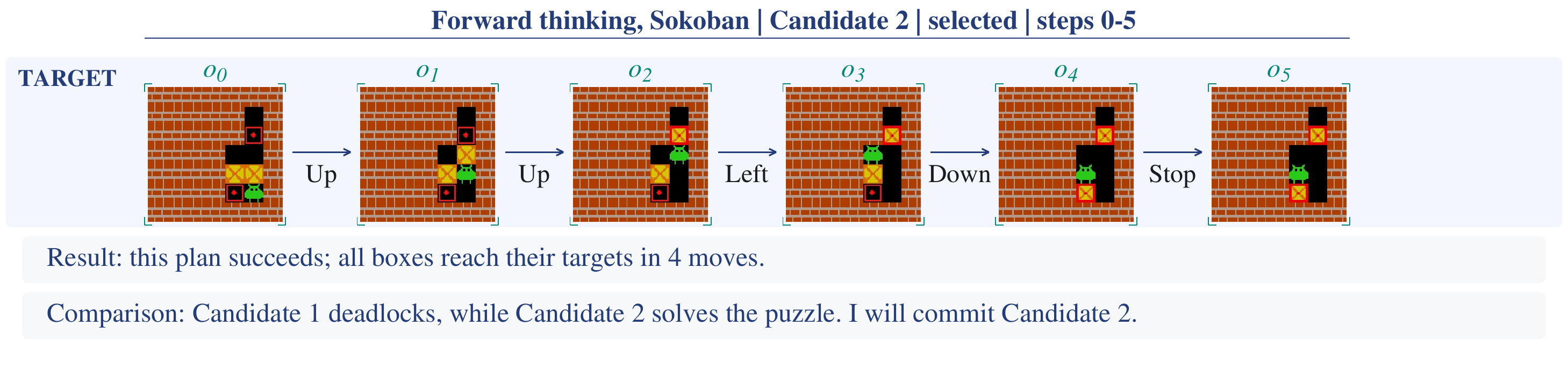}
    \caption{\textbf{Complete forward-thinking training example in Sokoban.} Candidate 1 deadlocks after two moves. Candidate 2 solves the puzzle in four moves and ends with \texttt{STOP}.}
    \label{fig:sokoban_forward_train}
\end{figure}

\begin{figure}[H]
    \centering
    \includegraphics[page=4,trim=0 1376bp 0 0bp,clip,width=\linewidth]{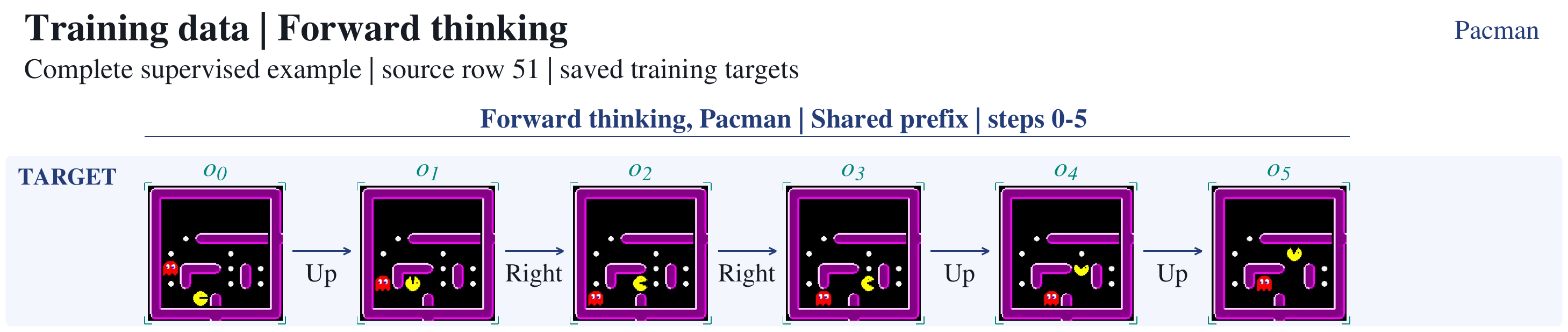}
\end{figure}

\begin{figure}[H]
    \centering
    \includegraphics[page=4,trim=0 1168bp 0 296bp,clip,width=\linewidth]{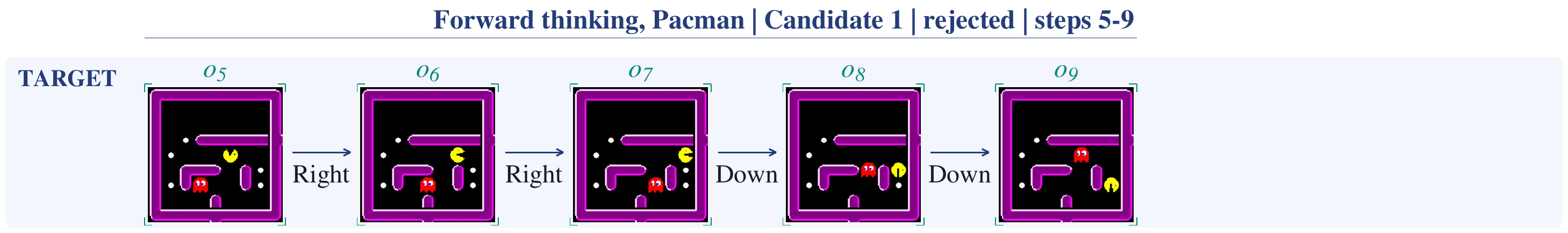}
\end{figure}

\begin{figure}[H]
    \centering
    \includegraphics[page=4,trim=0 910bp 0 504bp,clip,width=\linewidth]{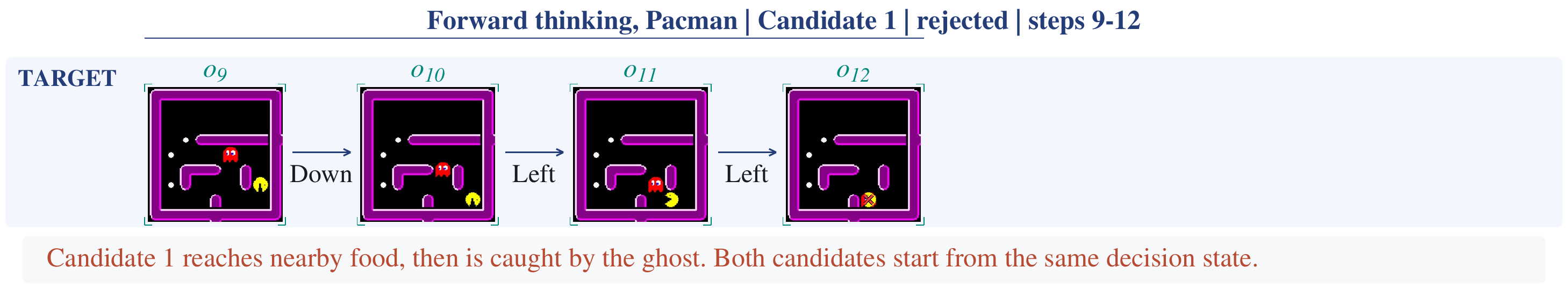}
\end{figure}

\begin{figure}[H]
    \centering
    \includegraphics[page=4,trim=0 702bp 0 762bp,clip,width=\linewidth]{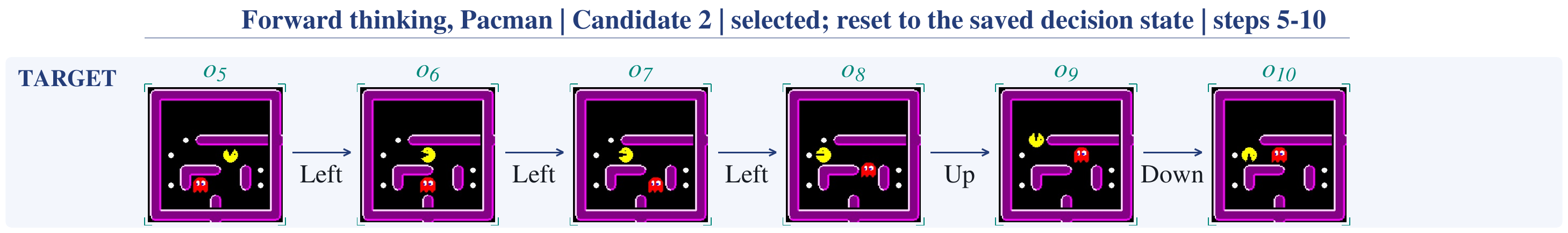}
\end{figure}

\begin{figure}[H]
    \centering
    \includegraphics[page=4,trim=0 494bp 0 970bp,clip,width=\linewidth]{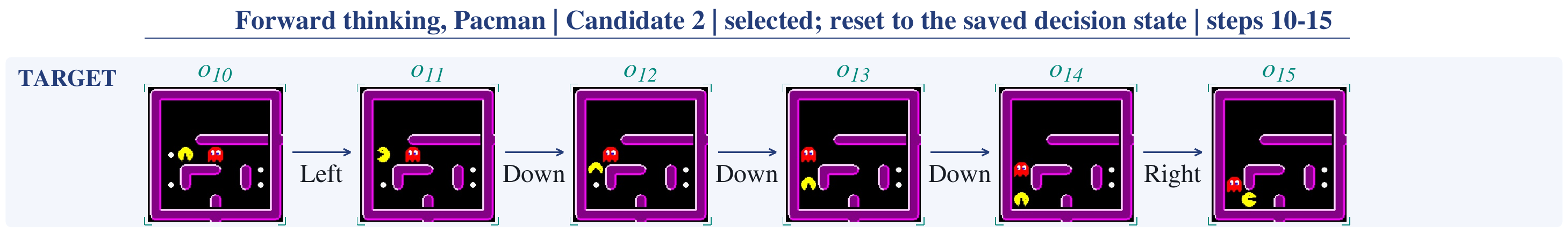}
\end{figure}

\begin{figure}[H]
    \centering
    \includegraphics[page=4,trim=0 286bp 0 1178bp,clip,width=\linewidth]{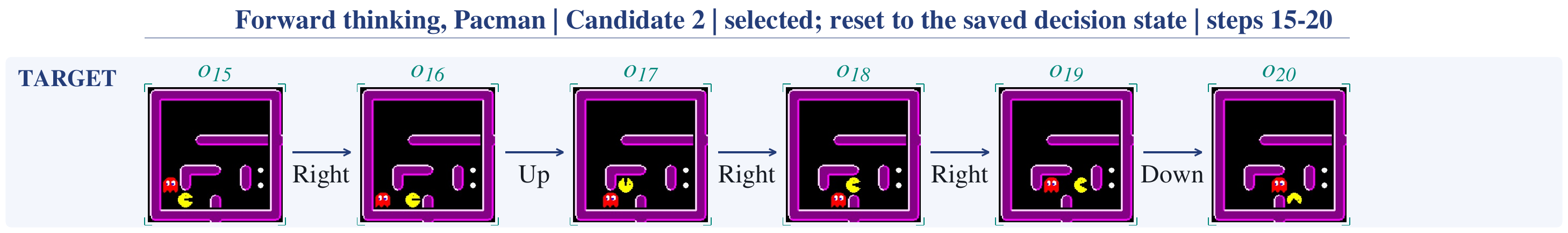}
\end{figure}

\begin{figure}[H]
    \centering
    \includegraphics[page=4,trim=0 0bp 0 1386bp,clip,width=\linewidth]{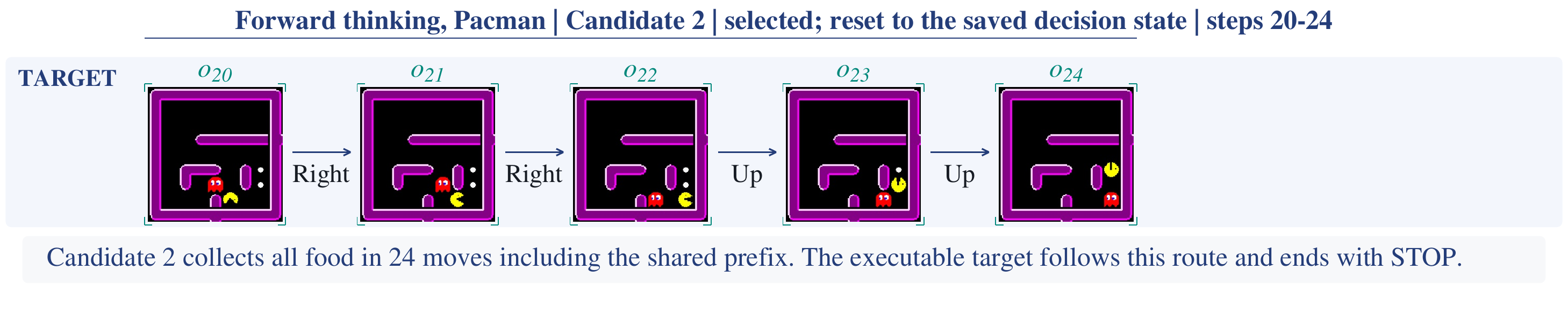}
    \caption{\textbf{Complete forward-thinking training example in Pacman.} After a shared five-move prefix, Candidate 1 is caught by the ghost after seven further moves. Candidate 2 collects all food after 19 further moves; its executable target ends with \texttt{STOP}. All saved training images are shown.}
    \label{fig:pacman_train_examples}
\end{figure}

\endgroup

\subsection{Model and Training Setup}
\label{app:training_setup}

\myparagraph{Backbone and SFT.}
We use BAGEL~\citep{deng2025bagel}, with 14B total parameters and 7B active
parameters. SFT uses teacher forcing, with next-token cross-entropy for text
outputs and the image-generation MSE objective described in
Section~\ref{sec:planning_formats}.

For PushT world-model training, we use a keep-last-VAE attention mask that restricts VAE conditioning to the most recent block while retaining all preceding text and ViT tokens.
At inference, we enforce the same policy by evicting the previous conditioning VAE block from the KV cache whenever a new image is added.

\begin{table}[H]
\centering
\small
\caption{\textbf{SFT settings.}}
\label{tab:sft_settings}
\setlength{\tabcolsep}{5pt}
\renewcommand{\arraystretch}{1.05}
\begin{tabularx}{\linewidth}{@{}lX@{}}
\toprule
\textbf{Setting} & \textbf{Value} \\
\midrule
Standard SFT steps & 5,000 \\
SFT batch size & 8 \\
Training-example presentations & 40,000 \\
Released Sokoban checkpoints & BAGEL-7B-MoT, exponential-moving-average (EMA) weights \\
Optimizer & AdamW ($\beta_1=0.9$, $\beta_2=0.95$) \\
Learning rate & $2\times10^{-5}$ \\
Learning-rate schedule & Cosine \\
Warmup steps & 100 \\
EMA decay & 0.993 \\
\bottomrule
\end{tabularx}
\end{table}

\myparagraph{Reinforcement learning.}
\label{app:rl_training}
\label{app:training_splits}
During RL, the model samples a reasoning trace containing textual decisions,
imagined visual states, and a final executable action sequence. It receives no
real environment observations during this deliberation. Only the final
sequence is executed, and the reward is binary task success.
We use GRPO~\citep{shao2024deepseekmath} with group size $N=8$, batch size 16,
and 200 training steps on ordinary $8\times8$ three-box Sokoban puzzles, without
additional expert demonstrations. The learning rate is $2\times10^{-6}$.
The Sokoban RL recipe uses sampling temperature $1.0$.
Evaluation uses held-out three- and
four-box Deadlock and Structural Fork sets.
The policy-gradient objective uses the language-token log probabilities.
Generated images are not directly supervised with an image reconstruction
loss during this stage.

\clearpage
\section{Analysis of Adaptive Imagination}
\label{app:rl_behavior}
We examine how RL changes the plans generated by the model. All results use
the four 3/4-box Deadlock and Structural Fork sets, with 100 episodes per set.
Pooled results therefore contain 400 episodes at each checkpoint. Pre-RL
denotes the SFT checkpoint; RL50, for example, denotes 50 RL training steps.

\subsection{Combining Linear Rollouts and Forward Thinking}\label{app:forward_gating}
SFT teaches two separate formats: a linear rollout that produces one plan,
and a fixed comparison that generates two candidate plans before choosing
one. RL combines them into a \emph{hybrid format}: the model begins with a
linear rollout, evaluates its predicted outcome, and adds another candidate
when it judges the first plan unsuccessful. Figure~\ref{fig:rl_qualitative}
shows this process continuing to a third candidate. The model can therefore
decide to compare plans after inspecting its first imagined outcome.
Related embodied agents use execution feedback to diagnose failures and revise plans.
Here, imagined outcomes trigger further planning before execution.
Table~\ref{tab:forward_behavior_detailed} separates the format change from
the decision to generate another candidate. From RL100 onward, almost every
comparison uses the hybrid format, and the model always judges the first
candidate unsuccessful before adding a second. We check that judgment by
executing the first candidate's actions from the original state. At RL200,
these actions fail to solve the puzzle in 95.5--100\% of the comparisons
across the four sets.

\begin{table}[H]
\centering
\small
\caption{
\textbf{Forward-Thinking formats and first-plan failures.}
C1 and C2 denote the first and second candidate plans.
Generates C2, Fixed comparison, and Hybrid use all 100 episodes per set as
their denominator. The last two rows use comparisons with readable candidate
actions and verdicts: Predicts C1 failure records the model's judgment;
Actual C1 failure records whether C1 fails when executed independently.
All values are percentages.
}
\label{tab:forward_behavior_detailed}
\setlength{\tabcolsep}{4pt}
\renewcommand{\arraystretch}{1.05}
\begin{tabular}{llccccc}
\toprule
\textbf{Set} & \textbf{Metric}
& \textbf{Pre-RL} & \textbf{RL50} & \textbf{RL100}
& \textbf{RL150} & \textbf{RL200} \\
\midrule

\multirow{5}{*}{3-box Deadlock}
& Generates C2      & 83\% & 57\% & 18\% & 20\% & 22\% \\
& Fixed comparison     & 79\% & 51\% & 0\%  & 0\%  & 0\%  \\
& Hybrid           & 4\%  & 6\%  & 18\% & 20\% & 22\% \\
& Predicts C1 failure       & 31.7\% & 33.9\% & 100\% & 100\% & 100\% \\
& Actual C1 failure       & 61.0\% & 64.3\% & 94.4\% & 100\% & 95.5\% \\
\midrule

\multirow{5}{*}{4-box Deadlock}
& Generates C2      & 76\% & 56\% & 34\% & 36\% & 37\% \\
& Fixed comparison     & 72\% & 42\% & 0\%  & 0\%  & 0\%  \\
& Hybrid           & 4\%  & 14\% & 34\% & 36\% & 37\% \\
& Predicts C1 failure       & 55.4\% & 63.0\% & 100\% & 100\% & 100\% \\
& Actual C1 failure       & 81.1\% & 83.3\% & 97.1\% & 97.2\% & 97.3\% \\
\midrule

\multirow{5}{*}{3-box Structural Fork}
& Generates C2      & 80\% & 59\% & 20\% & 29\% & 30\% \\
& Fixed comparison     & 75\% & 47\% & 0\%  & 0\%  & 0\%  \\
& Hybrid           & 5\%  & 12\% & 20\% & 29\% & 30\% \\
& Predicts C1 failure       & 53.5\% & 61.1\% & 100\% & 100\% & 100\% \\
& Actual C1 failure       & 73.2\% & 72.2\% & 100\% & 96.6\% & 96.7\% \\
\midrule

\multirow{5}{*}{4-box Structural Fork}
& Generates C2      & 77\% & 60\% & 31\% & 34\% & 34\% \\
& Fixed comparison     & 70\% & 42\% & 1\%  & 0\%  & 0\%  \\
& Hybrid           & 7\%  & 18\% & 30\% & 34\% & 34\% \\
& Predicts C1 failure       & 57.6\% & 60.0\% & 100\% & 100\% & 100\% \\
& Actual C1 failure       & 84.8\% & 83.6\% & 100\% & 100\% & 100\% \\
\bottomrule
\end{tabular}
\end{table}

\subsection{Counterfactual Branching}\label{app:branching_behavior}
RL mainly changes how many alternatives the model examines within a plan.
Across the four sets, the percentage of episodes with any branch changes
from 81.8\% before RL to 83.5\% at RL200. The mean number of branches per
episode increases from 1.70 to 1.89, reaching 1.95 at RL150.

Table~\ref{tab:branch_behavior_detailed} shows the results for each set.
On 3-box Structural Fork, the percentage of episodes with at least three
branches rises from 41\% before RL to 56\% at RL200; on 4-box Structural
Fork, it rises from 35\% to 54\%. Each alternative branch still contains
one imagined action. The critical-step analysis in Figure~\ref{fig:example}
measures whether these branches occur at decisions where a wrong action
could make the puzzle unsolvable.

\begin{table}[H]
\centering\small
\caption{\textbf{Number of counterfactual branches per episode.}
Mean counts branches across all 100 episodes in each row, including episodes
with no branch. Columns 0, 1, 2, and $\geq3$ give the percentage of episodes
with that many complete branches. Each branch contains one imagined action.}
\label{tab:branch_behavior_detailed}
\setlength{\tabcolsep}{5pt}
\renewcommand{\arraystretch}{1.05}
\begin{tabular}{llccccc}
\toprule
\textbf{Set} & \textbf{Checkpoint} & \textbf{Mean}
& \textbf{0} & \textbf{1} & \textbf{2} & $\mathbf{\geq3}$ \\
\midrule
\multirow{5}{*}{3-box Deadlock} & Pre-RL & 1.18 & 34\% & 31\% & 18\% & 17\% \\
 & RL50 & 1.13 & 33\% & 36\% & 16\% & 15\% \\
 & RL100 & 1.20 & 34\% & 30\% & 18\% & 18\% \\
 & RL150 & 1.25 & 35\% & 25\% & 20\% & 20\% \\
 & RL200 & 1.25 & 35\% & 25\% & 20\% & 20\% \\
\midrule
\multirow{5}{*}{4-box Deadlock} & Pre-RL & 1.67 & 22\% & 23\% & 21\% & 34\% \\
 & RL50 & 1.67 & 23\% & 20\% & 24\% & 33\% \\
 & RL100 & 1.73 & 17\% & 29\% & 18\% & 36\% \\
 & RL150 & 1.88 & 13\% & 24\% & 25\% & 38\% \\
 & RL200 & 1.76 & 19\% & 22\% & 23\% & 36\% \\
\midrule
\multirow{5}{*}{3-box Structural Fork} & Pre-RL & 2.08 & 7\% & 19\% & 33\% & 41\% \\
 & RL50 & 2.14 & 5\% & 23\% & 25\% & 47\% \\
 & RL100 & 2.23 & 4\% & 18\% & 29\% & 49\% \\
 & RL150 & 2.43 & 5\% & 11\% & 20\% & 64\% \\
 & RL200 & 2.30 & 6\% & 14\% & 24\% & 56\% \\
\midrule
\multirow{5}{*}{4-box Structural Fork} & Pre-RL & 1.88 & 10\% & 27\% & 28\% & 35\% \\
 & RL50 & 2.00 & 6\% & 26\% & 30\% & 38\% \\
 & RL100 & 2.04 & 7\% & 22\% & 31\% & 40\% \\
 & RL150 & 2.25 & 5\% & 19\% & 22\% & 54\% \\
 & RL200 & 2.24 & 6\% & 18\% & 22\% & 54\% \\
\bottomrule
\end{tabular}
\end{table}

\subsection{Task Success and Rollout Length}
\label{app:rollout_length}
Table~\ref{tab:rollout_length} reports task success and the mean numbers of
imagined and executed actions. From Pre-RL to RL200, forward thinking reduces
imagined actions from 27.82 to 22.18 per episode, while task success rises
from 30.3\% to 41.0\%. Branching reduces imagined actions from 23.62 to 21.59,
while task success rises from 30.0\% to 50.3\%. Thus, both models improve
without generating longer imagined trajectories. These are action counts;
we do not measure inference time or FLOPs here.

\begin{table}[H]
\centering\small
\caption{\textbf{Mean rollout lengths and pooled task success on the four 3/4-box sets.}
Each row pools 400 episodes. Task success is a percentage; Imagined and
Executed are mean action counts per episode.}
\label{tab:rollout_length}
\setlength{\tabcolsep}{7pt}
\begin{tabular}{llrrr}
\toprule
Format & Checkpoint & Success & Imagined & Executed\\
\midrule
Forward & Pre-RL & 30.3\% & 27.82 & 17.88 \\
 & RL50 & 35.0\% & 24.95 & 17.36 \\
 & RL100 & 37.5\% & 20.87 & 16.96 \\
 & RL150 & 38.5\% & 21.40 & 16.97 \\
 & RL200 & 41.0\% & 22.18 & 17.50 \\
\midrule
Branching & Pre-RL & 30.0\% & 23.62 & 18.63 \\
 & RL50 & 32.0\% & 22.60 & 18.15 \\
 & RL100 & 40.5\% & 21.88 & 17.95 \\
 & RL150 & 45.8\% & 23.16 & 18.87 \\
 & RL200 & 50.3\% & 21.59 & 18.24 \\
\bottomrule
\end{tabular}
\end{table}

\clearpage
\section{Qualitative Dynamics Predictions and Rollouts}
\label{app:dynamics_modeling}
\label{app:dynamics_modeling_qualitative}
The following examples compare predicted visual states with ground-truth
states along action sequences in each environment. These selected trajectories
provide qualitative evidence; they do not report aggregate prediction accuracy.
The examples show complete recorded trajectories,
from the initial observation to episode termination. Every action and saved
state is displayed; adjacent rows or pages repeat their shared endpoint.
Examples at different chunk sizes may start from different initial states.
Additionally, we show example traces of counterfactual branching and forward thinking.

\subsection{Maze2D Rollouts}
\label{app:qualitative_maze}
Task configurations are described in Appendix~\ref{app:maze2d}.
\begin{figure}[H]
    \centering
    \includegraphics[page=1,width=\linewidth,height=0.90\textheight,keepaspectratio]{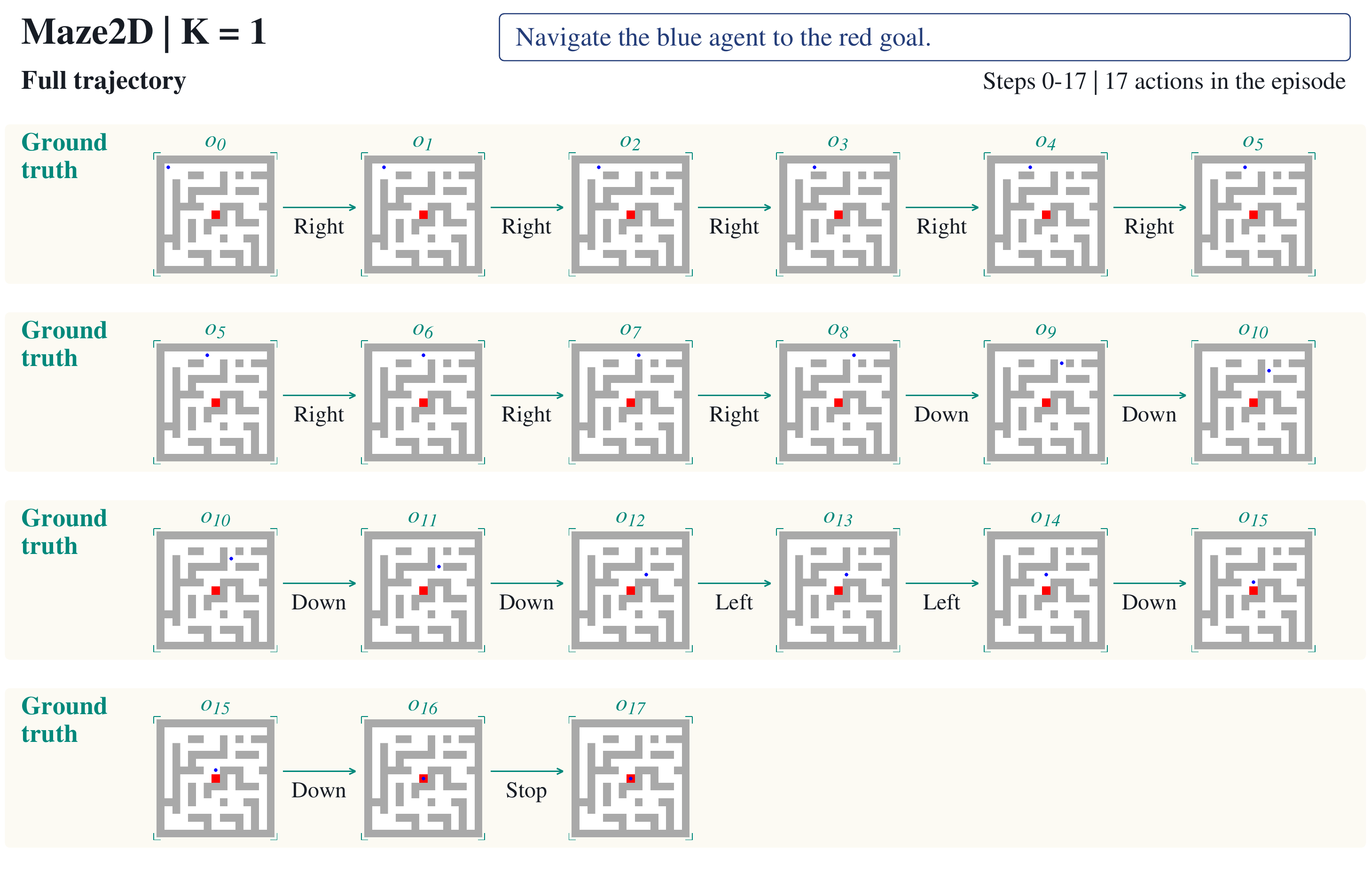}
    \caption{\textbf{Complete Maze2D trajectory at $K=1$.}
    Selected successful Ordinary Hard episode; all 17 actions and their states are shown.
    The reactive reference receives a real observation after every action.}
    \label{fig:maze2d_full_k1}
    \label{fig:maze2d_chunk_qualitative}
\end{figure}

\clearpage
\begin{figure}[H]
    \centering
    \includegraphics[page=2,width=\linewidth,height=0.90\textheight,keepaspectratio]{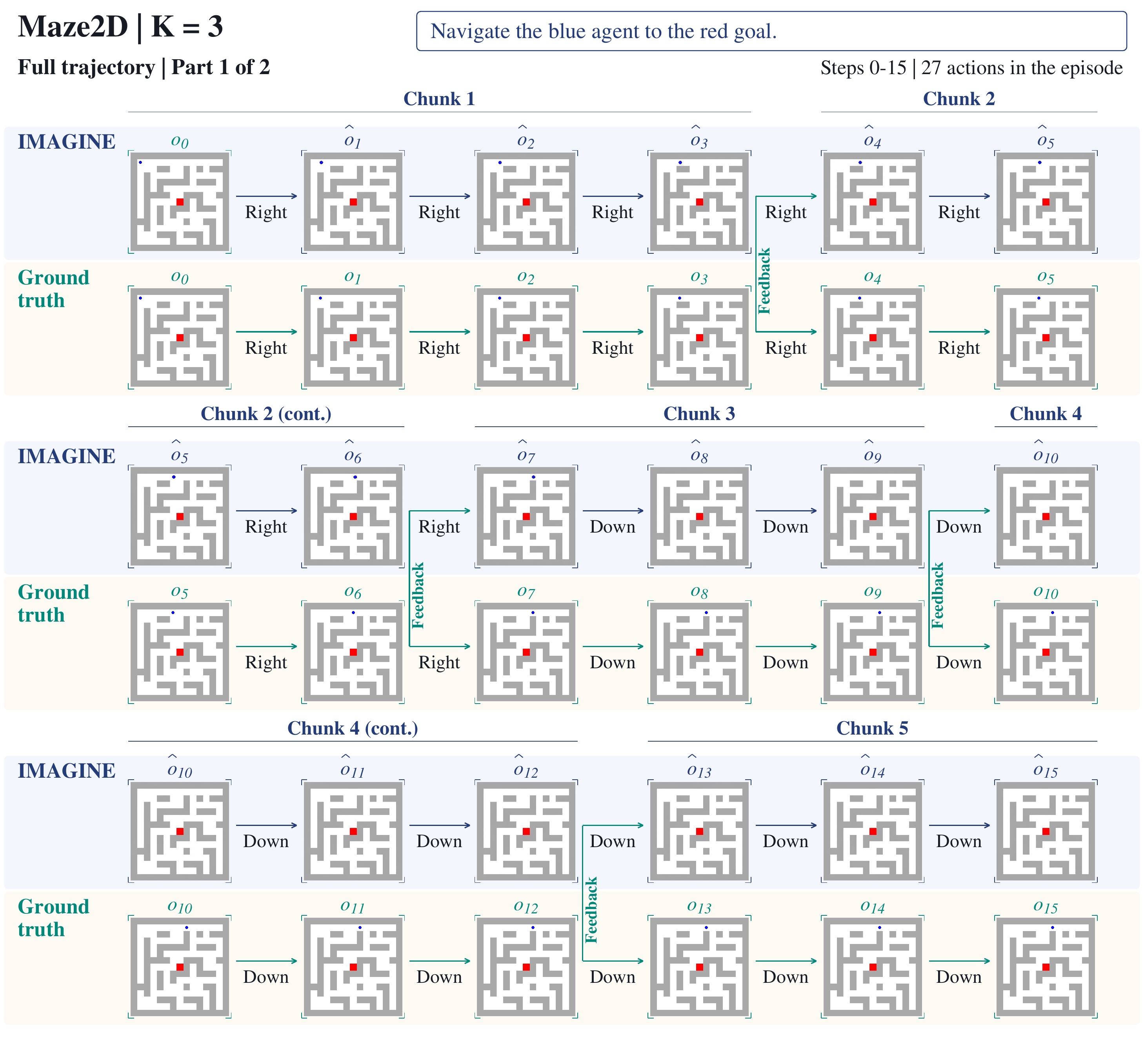}
    \caption{\textbf{Complete Maze2D trajectory at $K=3$.}
    Selected successful Ordinary Hard episode; this page shows steps 0--15 of 27.
    The trajectory continues on the next page.
    Blue rows show predictions; teal inputs and feedback arrows show the real
    observations supplied between chunks.}
    \label{fig:maze2d_full_k3}
\end{figure}

\clearpage
\begin{figure}[H]
    \centering
    \includegraphics[page=3,width=\linewidth,height=0.90\textheight,keepaspectratio]{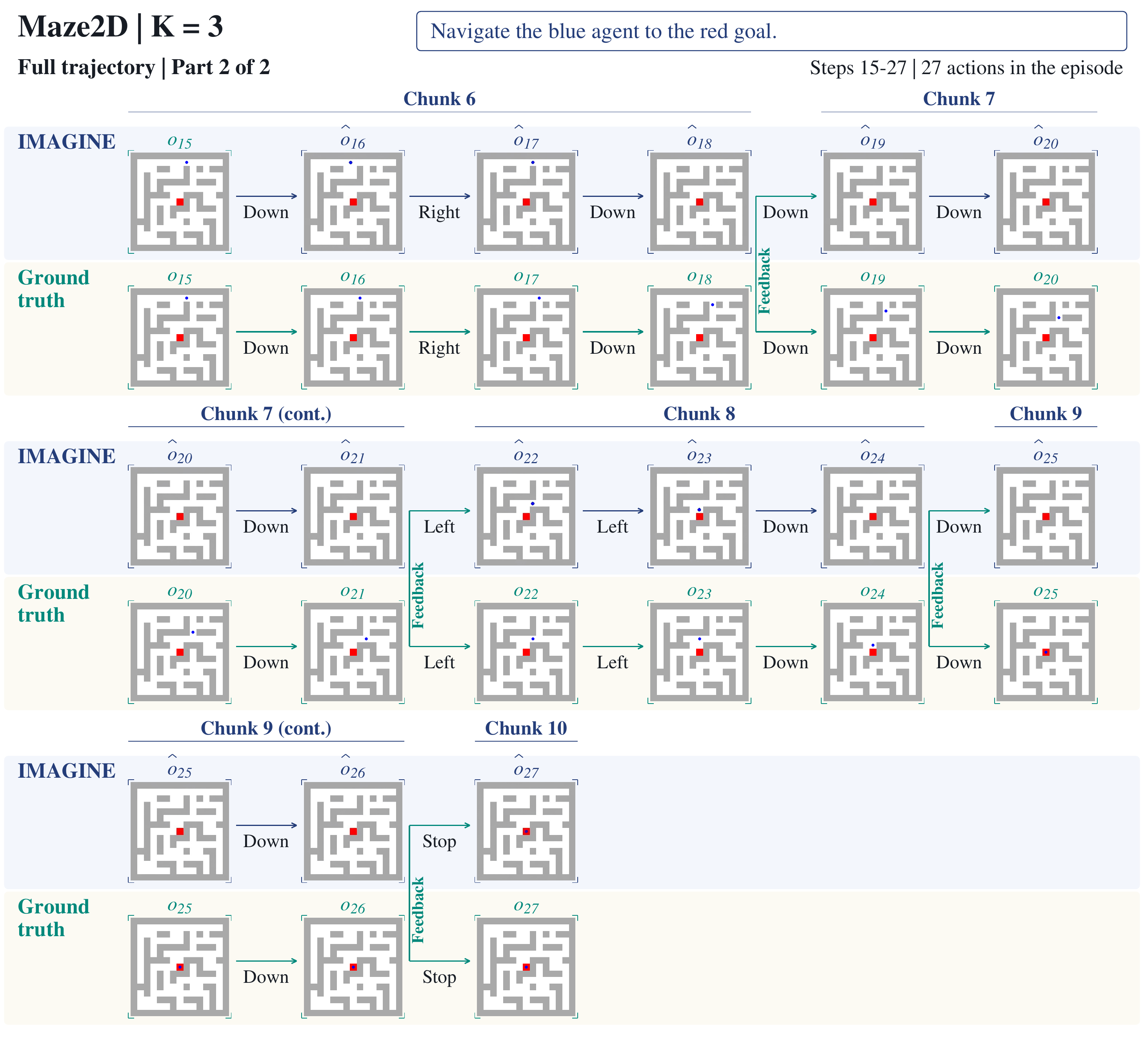}
    \par\vspace{0.5\baselineskip}\noindent\parbox{\linewidth}{Figure~\ref{fig:maze2d_full_k3} (continued).
    \textbf{Maze2D, $K=3$.} Steps 15--27, ending in success. The shared endpoint connects the two pages.}
\end{figure}

\clearpage
\begin{figure}[H]
    \centering
    \includegraphics[page=4,width=\linewidth,height=0.90\textheight,keepaspectratio]{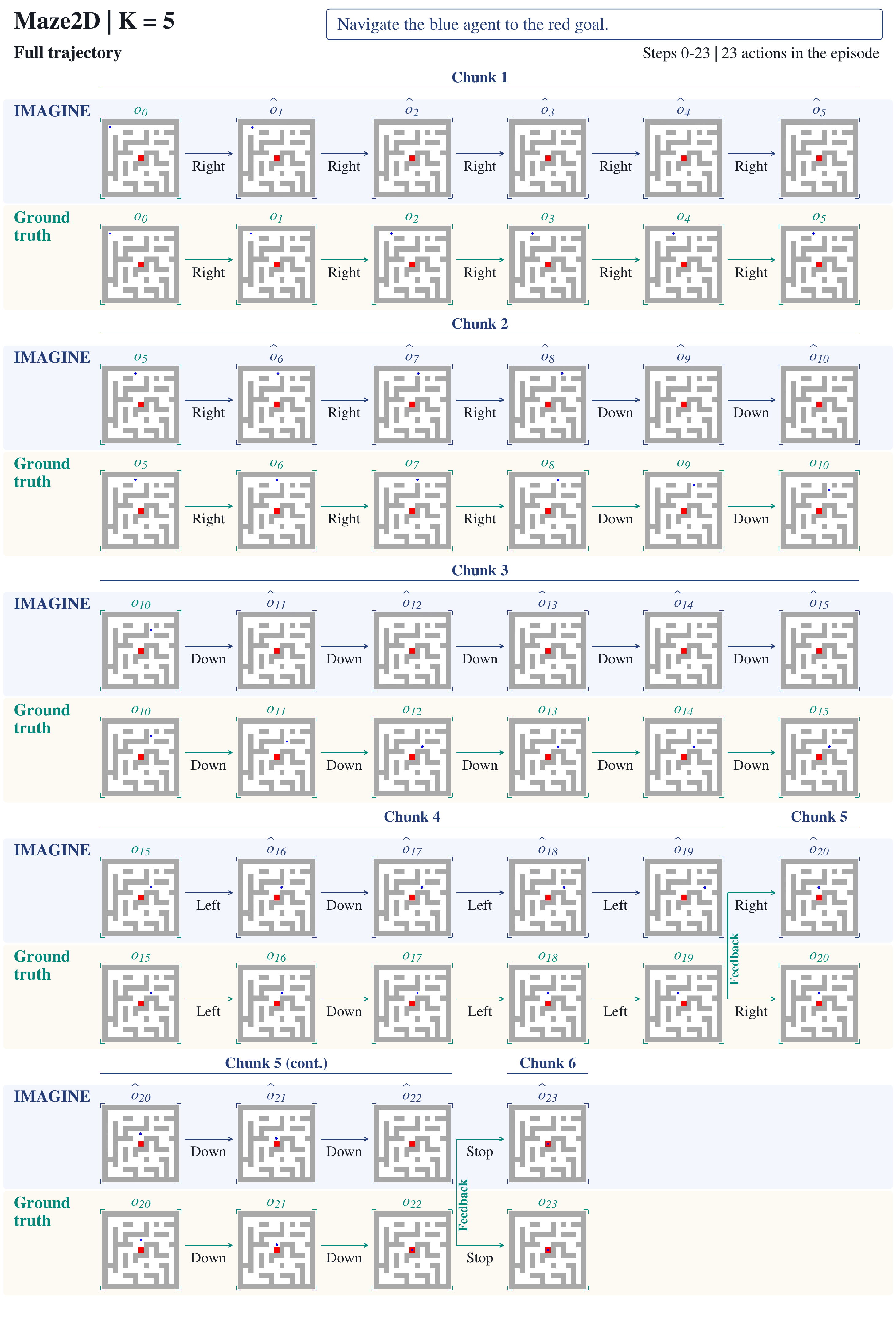}
    \caption{\textbf{Complete Maze2D trajectory at $K=5$.}
    Selected successful Ordinary Hard episode; all 23 actions and their states are shown.
    Blue rows show predictions; teal inputs and feedback arrows show the real
    observations supplied between chunks.}
    \label{fig:maze2d_full_k5}
\end{figure}

\clearpage
\begin{figure}[H]
    \centering
    \includegraphics[page=5,width=\linewidth,height=0.82\textheight,keepaspectratio]{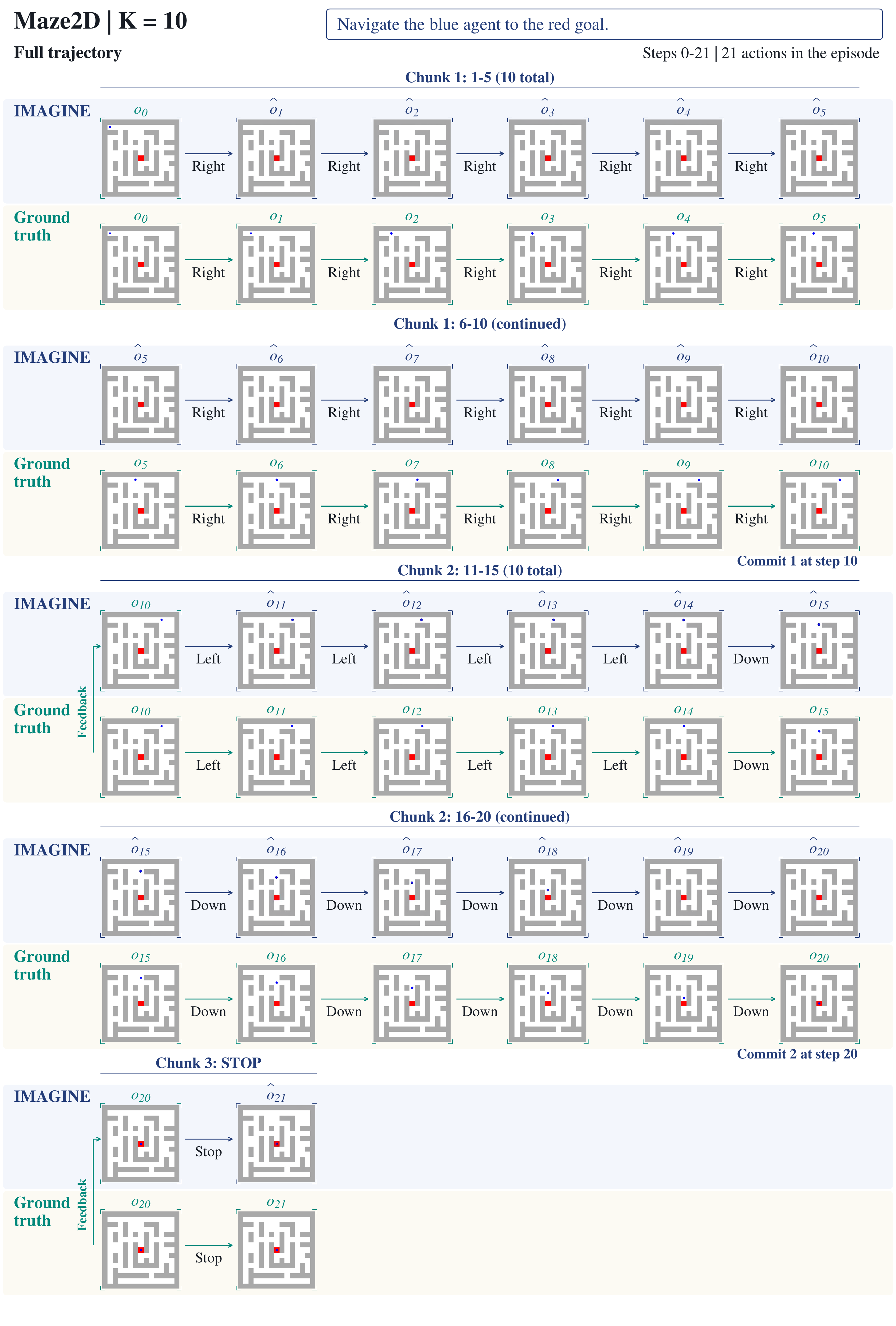}
    \caption{\textbf{Complete Maze2D trajectory at $K=10$.}
    Selected successful Ordinary Hard episode; all 21 actions and their states are shown.
    Blue rows show predictions; teal inputs and feedback arrows show the real
    observations supplied between chunks.}
    \label{fig:maze2d_full_k10}
\end{figure}

\clearpage
\begin{figure}[H]
    \centering
    \includegraphics[page=6,width=\linewidth,height=0.90\textheight,keepaspectratio]{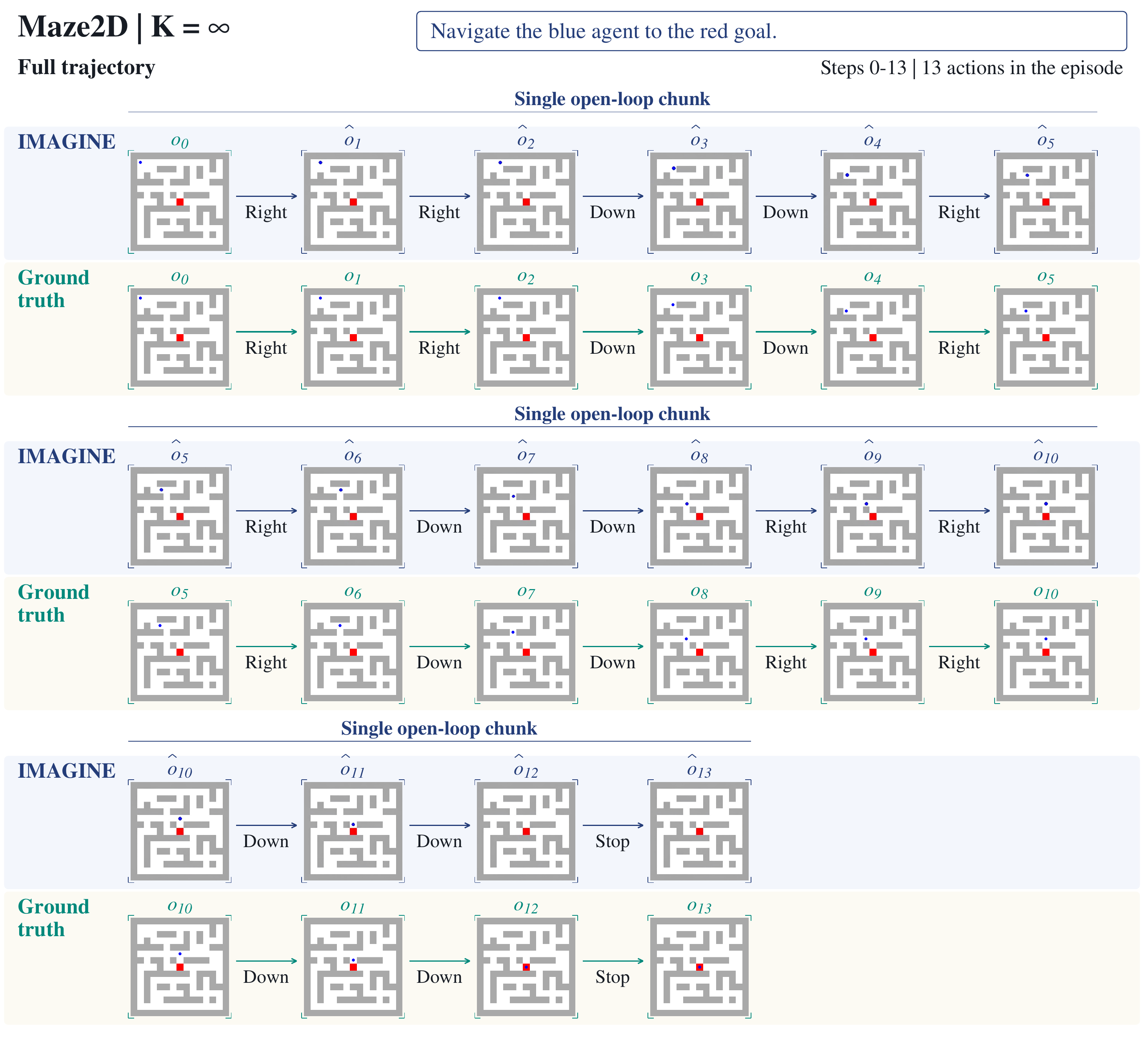}
    \caption{\textbf{Complete Maze2D trajectory at $K=\infty$.}
    Selected successful Ordinary Hard episode; all 13 actions and their states are shown.
    The model receives only the initial observation, with no intermediate
    feedback. Ground-truth states are shown for comparison.}
    \label{fig:maze2d_full_kinf}
    \label{fig:maze2d_easy_to_hard}
\end{figure}

\clearpage
\subsection{Sokoban Rollouts}
\label{app:qualitative_sokoban}
Task configurations are described in Appendix~\ref{app:sokoban}.
\begin{figure}[H]
    \centering
    \includegraphics[page=1,width=\linewidth,height=0.90\textheight,keepaspectratio]{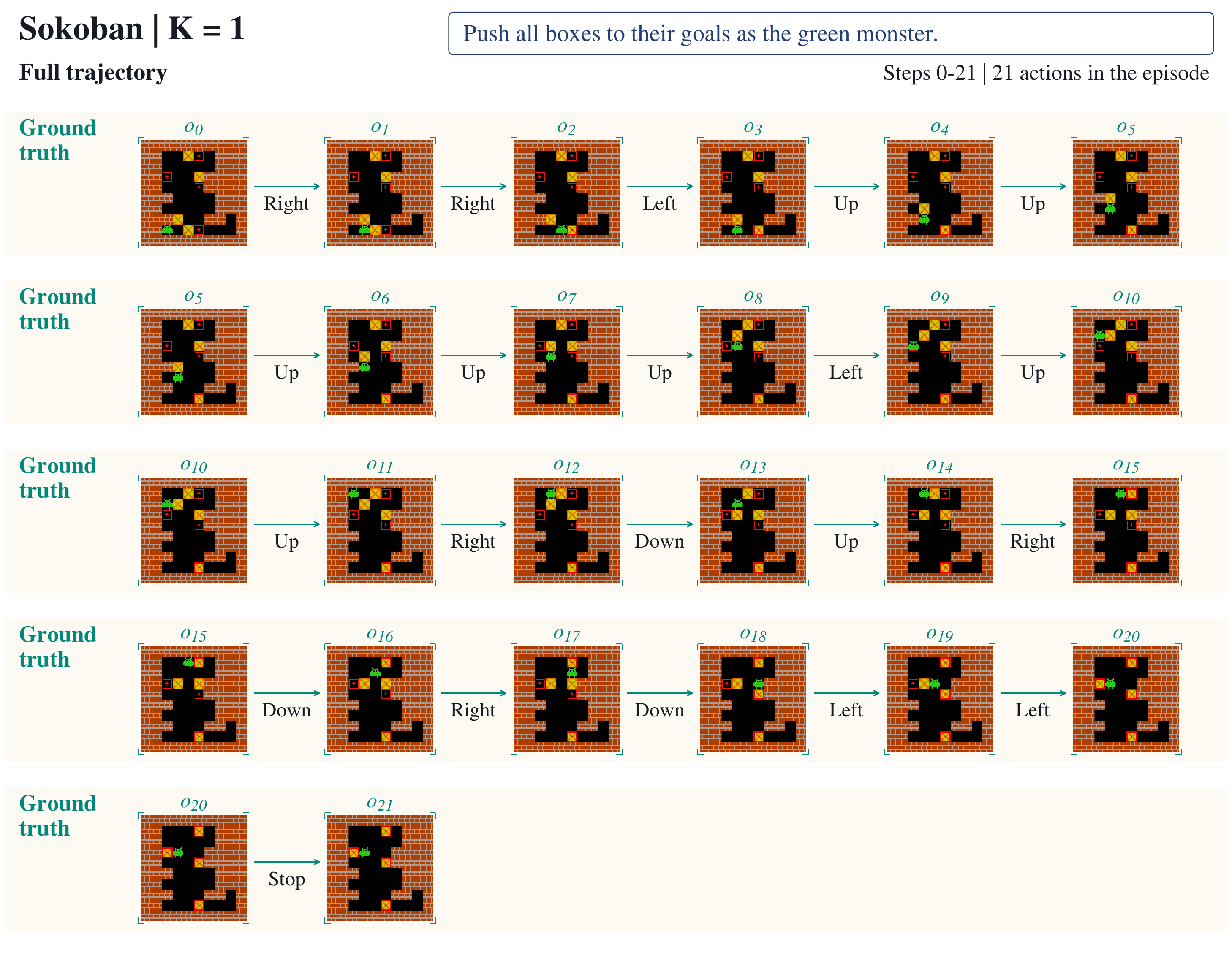}
    \caption{\textbf{Complete Sokoban trajectory at $K=1$.}
    Selected successful Ordinary Hard episode; all 21 actions and their states are shown.
    The reactive reference receives a real observation after every action.}
    \label{fig:sokoban_full_k1}
    \label{fig:sokoban_chunk_qualitative}
\end{figure}

\clearpage
\begin{figure}[H]
    \centering
    \includegraphics[page=2,width=\linewidth,height=0.90\textheight,keepaspectratio]{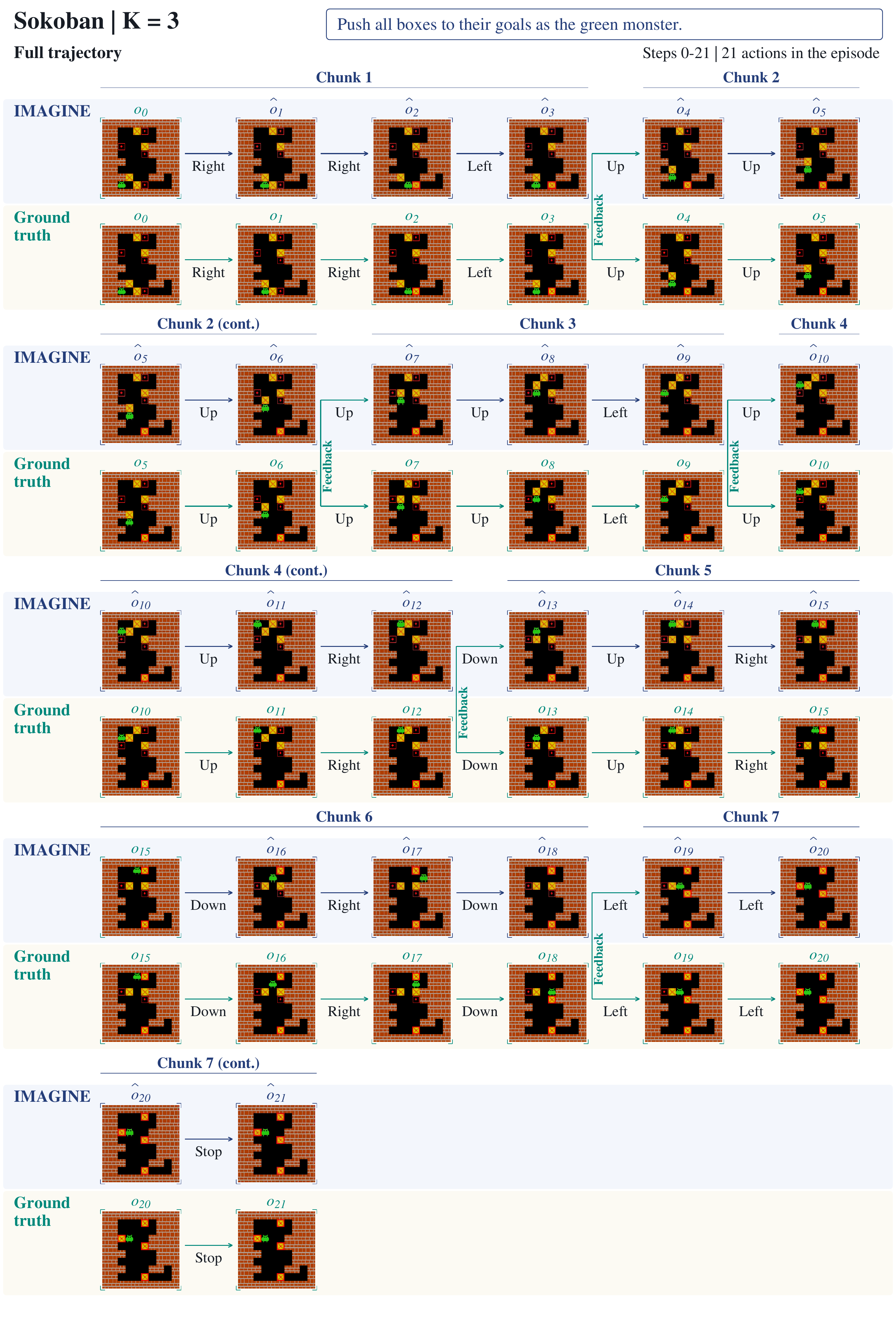}
    \caption{\textbf{Complete Sokoban trajectory at $K=3$.}
    Selected successful Ordinary Hard episode; all 21 actions and their states are shown.
    Blue rows show predictions; teal inputs and feedback arrows show the real
    observations supplied between chunks.}
    \label{fig:sokoban_full_k3}
\end{figure}

\clearpage
\begin{figure}[H]
    \centering
    \includegraphics[page=3,width=\linewidth,height=0.90\textheight,keepaspectratio]{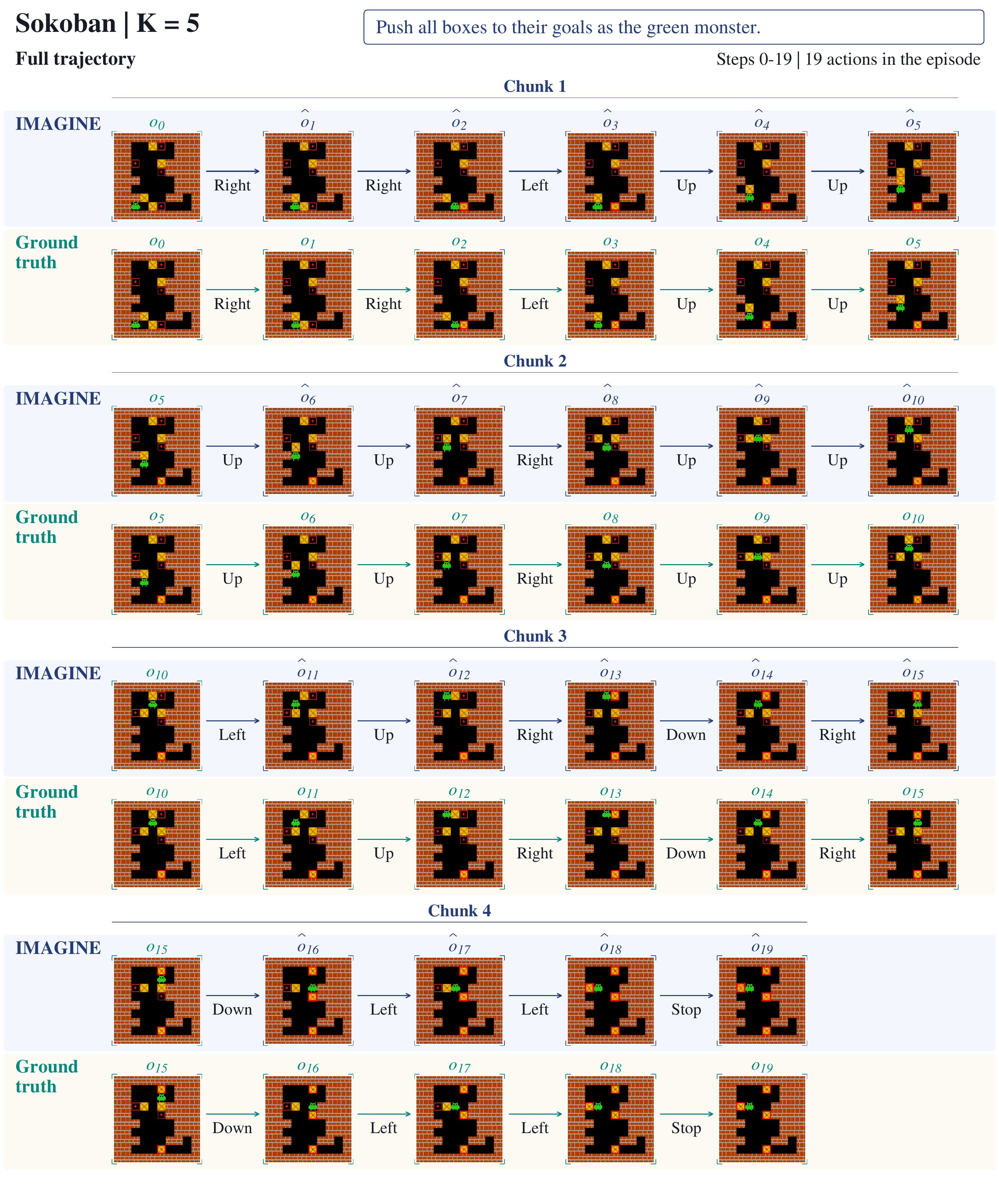}
    \caption{\textbf{Complete Sokoban trajectory at $K=5$.}
    Selected successful Ordinary Hard episode; all 19 actions and their states are shown.
    Blue rows show predictions; teal inputs and feedback arrows show the real
    observations supplied between chunks.}
    \label{fig:sokoban_full_k5}
\end{figure}

\clearpage
\begin{figure}[H]
    \centering
    \includegraphics[page=4,width=\linewidth,height=0.90\textheight,keepaspectratio]{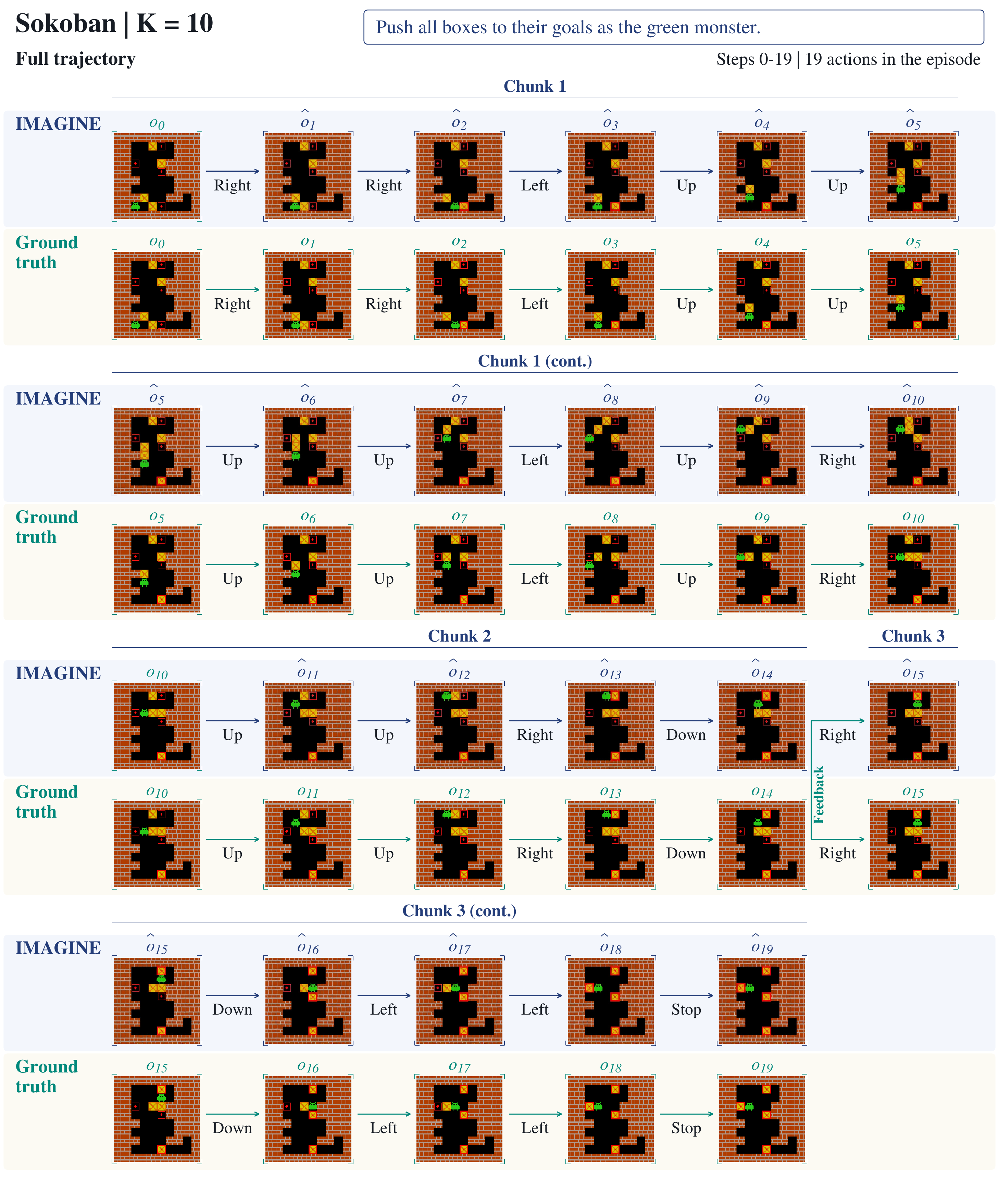}
    \caption{\textbf{Complete Sokoban trajectory at $K=10$.}
    Selected successful Ordinary Hard episode; all 19 actions and their states are shown.
    Blue rows show predictions; teal inputs and feedback arrows show the real
    observations supplied between chunks.}
    \label{fig:sokoban_full_k10}
\end{figure}

\clearpage
\begin{figure}[H]
    \centering
    \includegraphics[page=5,width=\linewidth,height=0.90\textheight,keepaspectratio]{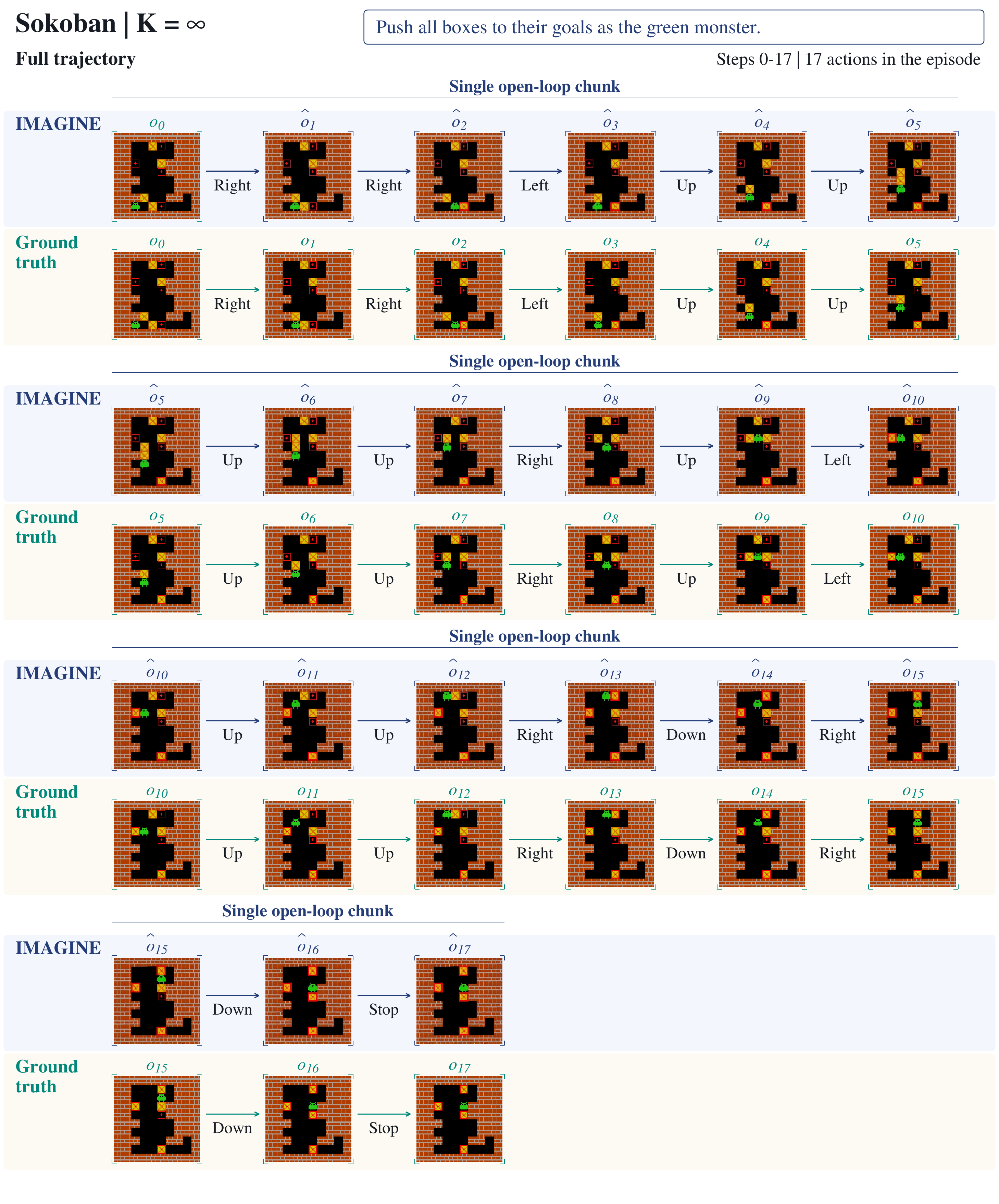}
    \caption{\textbf{Complete Sokoban trajectory at $K=\infty$.}
    Selected successful Ordinary Hard episode; all 17 actions and their states are shown.
    The model receives only the initial observation, with no intermediate
    feedback. Ground-truth states are shown for comparison.}
    \label{fig:sokoban_full_kinf}
    \label{fig:sokoban_easy_to_hard}
\end{figure}

\clearpage
\subsection{Pacman Rollouts}
\label{app:qualitative_pacman}
Task configurations are described in Appendix~\ref{app:pacman}.
\begin{figure}[H]
    \centering
    \includegraphics[page=1,width=\linewidth,height=0.90\textheight,keepaspectratio]{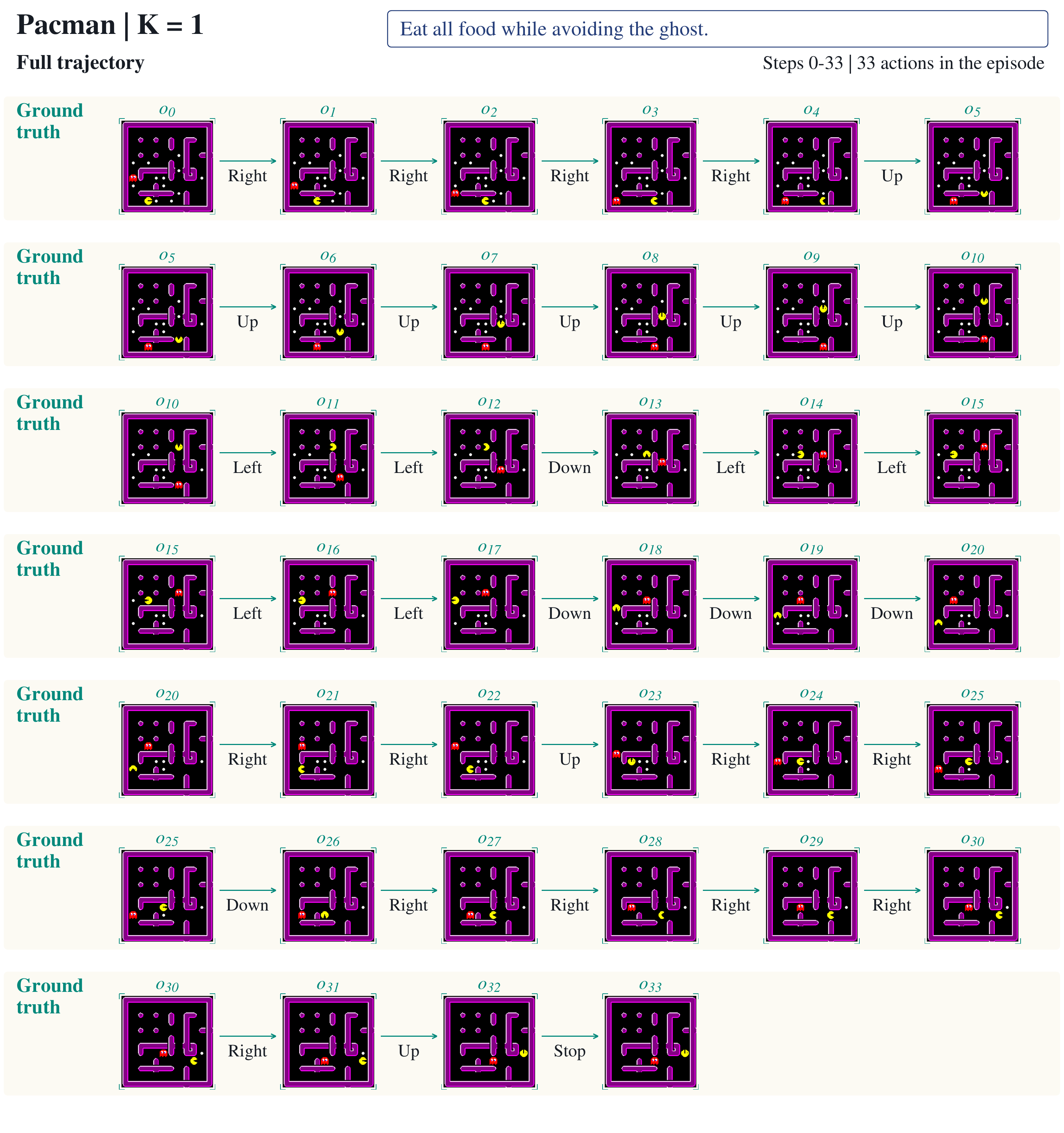}
    \caption{\textbf{Complete Pacman trajectory at $K=1$.}
    Selected successful Ordinary Hard episode; all 33 actions and their states are shown.
    The reactive reference receives a real observation after every action.}
    \label{fig:pacman_full_k1}
    \label{fig:pacman_chunk_qualitative}
\end{figure}

\clearpage
\begin{figure}[H]
    \centering
    \includegraphics[page=2,width=\linewidth,height=0.90\textheight,keepaspectratio]{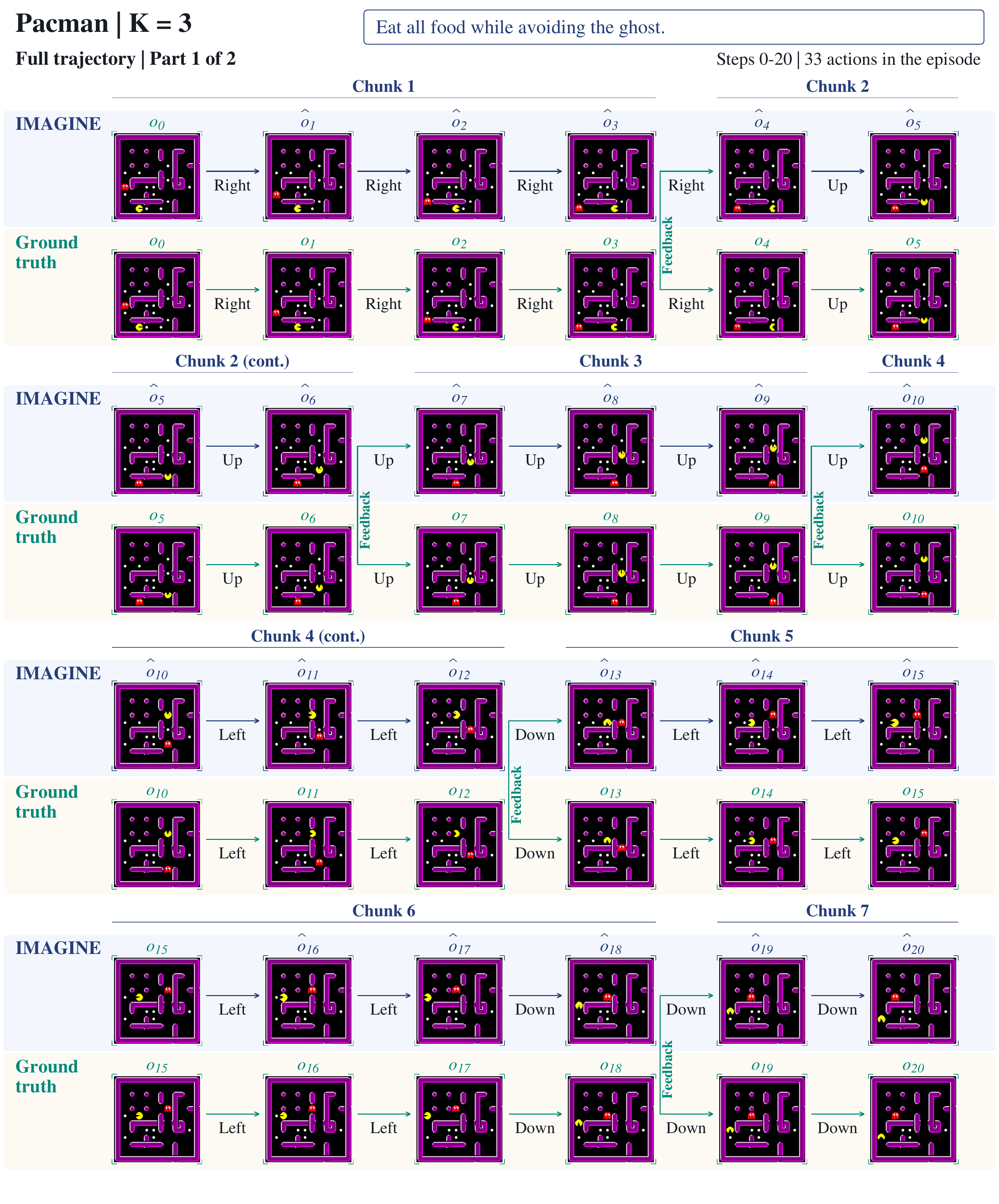}
    \caption{\textbf{Complete Pacman trajectory at $K=3$.}
    Selected successful Ordinary Hard episode; this page shows steps 0--20 of 33.
    The trajectory continues on the next page.
    Blue rows show predictions; teal inputs and feedback arrows show the real
    observations supplied between chunks.}
    \label{fig:pacman_full_k3}
\end{figure}

\clearpage
\begin{figure}[H]
    \centering
    \includegraphics[page=3,width=\linewidth,height=0.90\textheight,keepaspectratio]{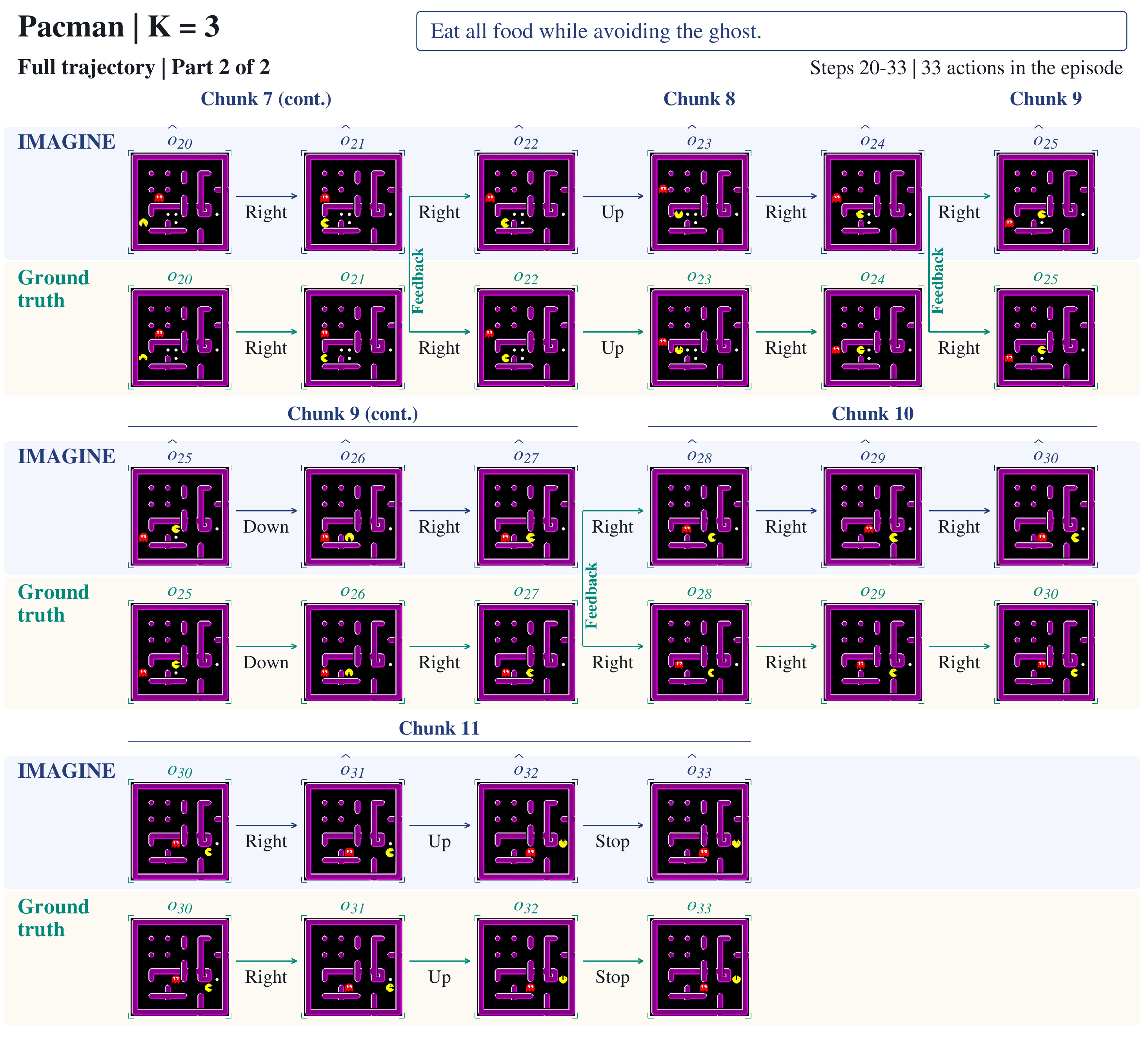}
    \par\vspace{0.5\baselineskip}\noindent\parbox{\linewidth}{Figure~\ref{fig:pacman_full_k3} (continued).
    \textbf{Pacman, $K=3$.} Steps 20--33, ending in success. The shared endpoint connects the two pages.}
\end{figure}

\clearpage
\begin{figure}[H]
    \centering
    \includegraphics[page=4,width=\linewidth,height=0.90\textheight,keepaspectratio]{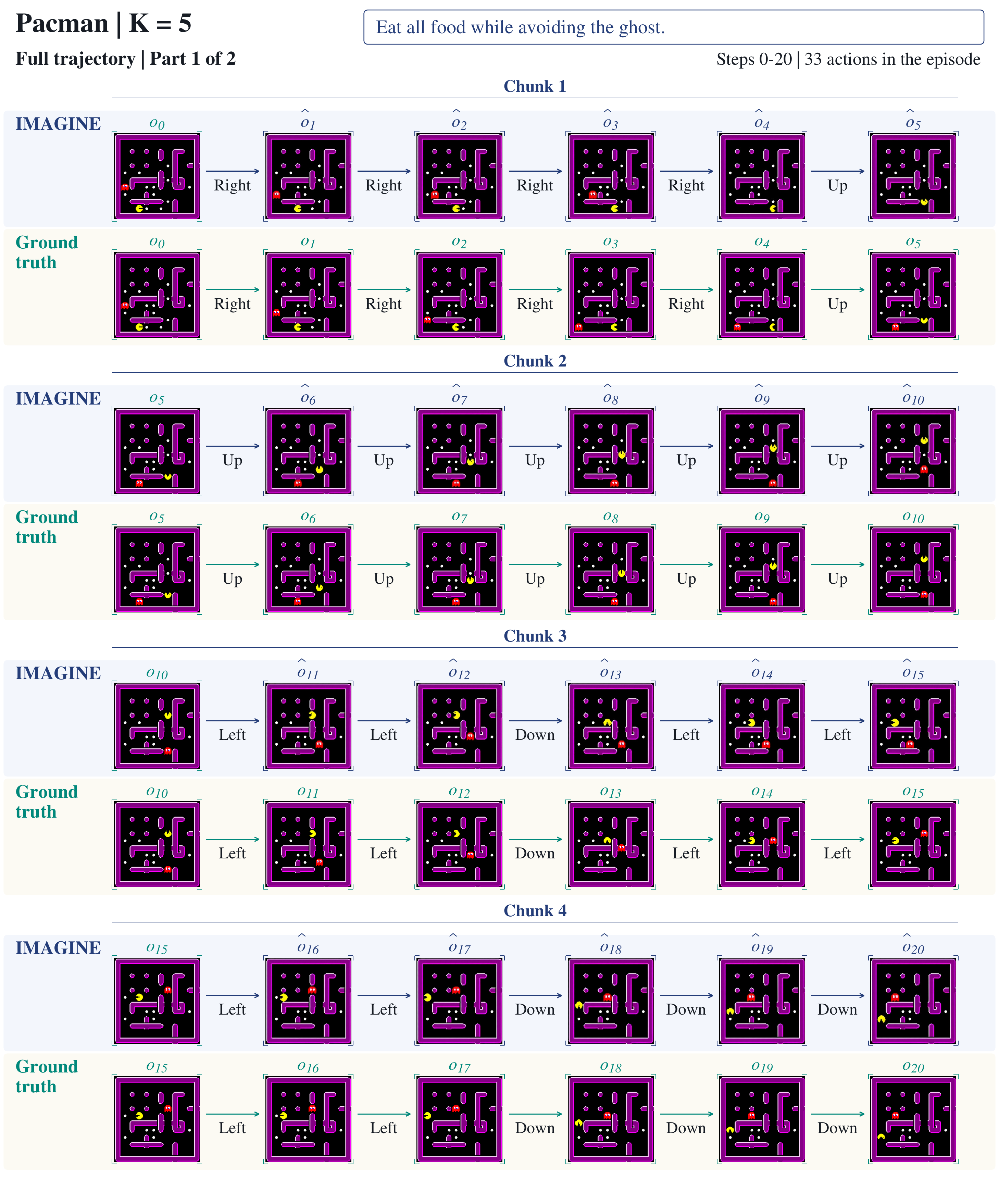}
    \caption{\textbf{Complete Pacman trajectory at $K=5$.}
    Selected successful Ordinary Hard episode; this page shows steps 0--20 of 33.
    The trajectory continues on the next page.
    Blue rows show predictions; teal inputs and feedback arrows show the real
    observations supplied between chunks.}
    \label{fig:pacman_full_k5}
\end{figure}

\clearpage
\begin{figure}[H]
    \centering
    \includegraphics[page=5,width=\linewidth,height=0.90\textheight,keepaspectratio]{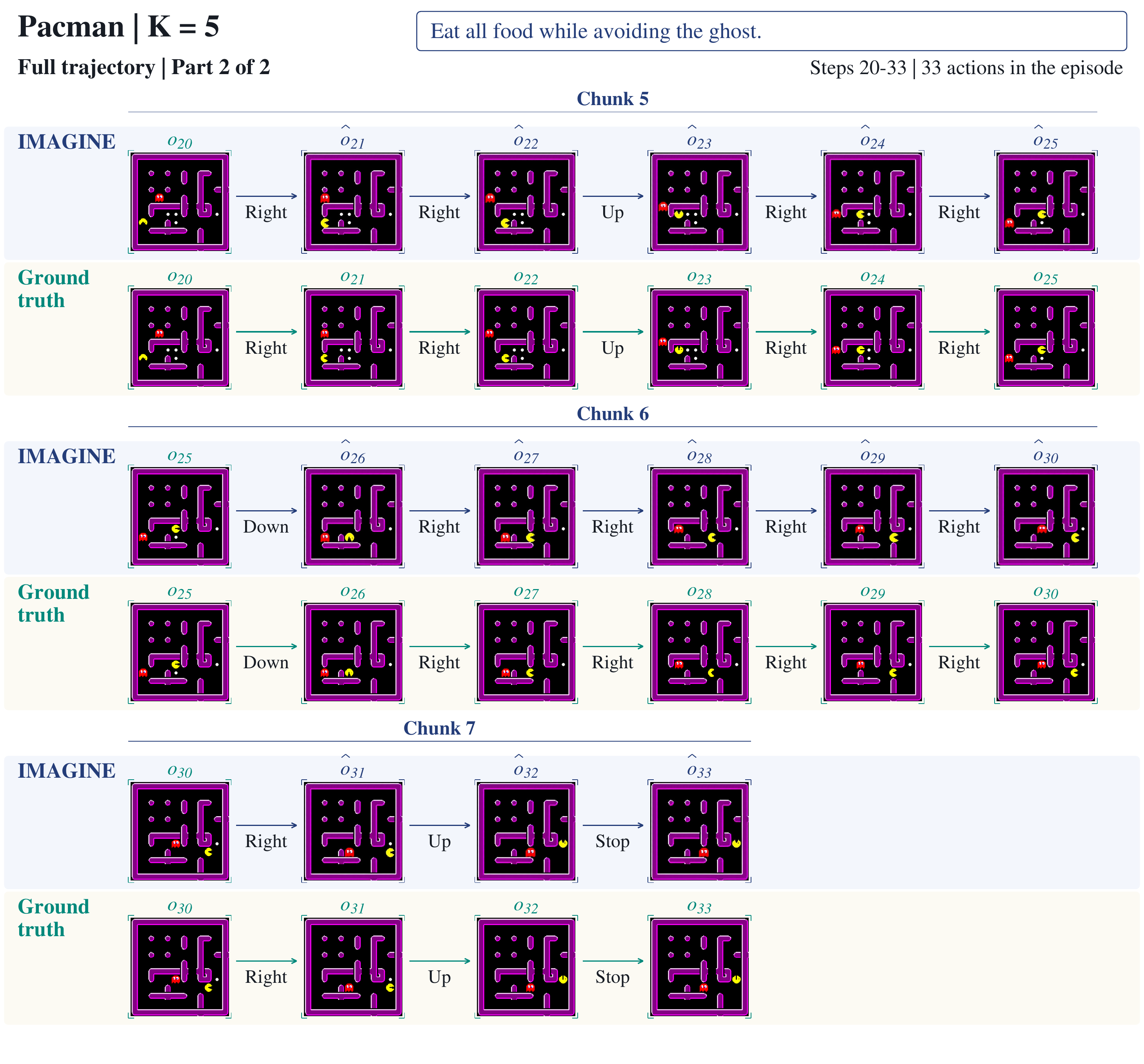}
    \par\vspace{0.5\baselineskip}\noindent\parbox{\linewidth}{Figure~\ref{fig:pacman_full_k5} (continued).
    \textbf{Pacman, $K=5$.} Steps 20--33, ending in success. The shared endpoint connects the two pages.}
\end{figure}

\clearpage
\begin{figure}[H]
    \centering
    \includegraphics[page=6,width=\linewidth,height=0.90\textheight,keepaspectratio]{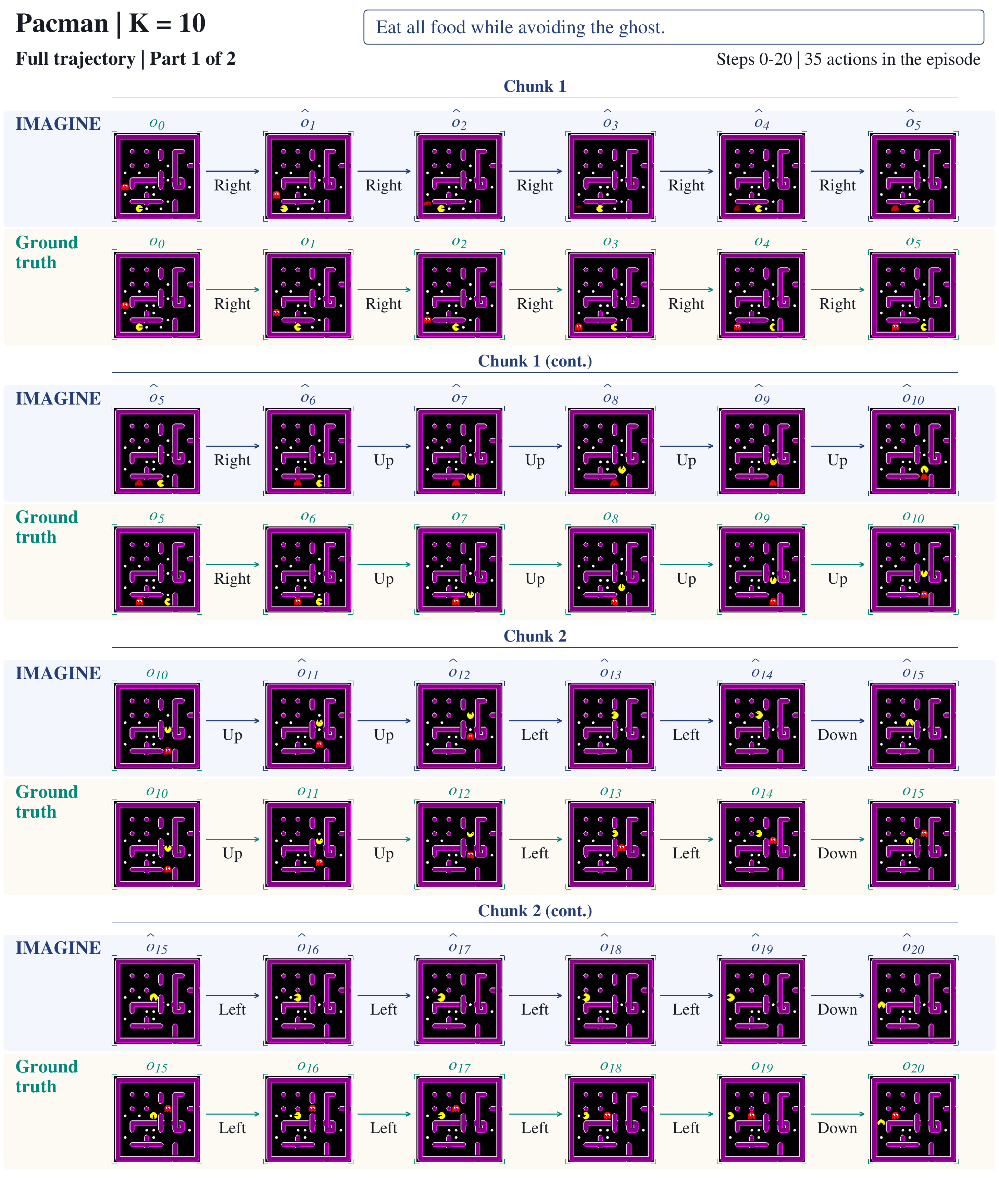}
    \caption{\textbf{Complete Pacman trajectory at $K=10$.}
    Selected successful Ordinary Hard episode; this page shows steps 0--20 of 35.
    The trajectory continues on the next page.
    Blue rows show predictions; teal inputs and feedback arrows show the real
    observations supplied between chunks.}
    \label{fig:pacman_full_k10}
\end{figure}

\clearpage
\begin{figure}[H]
    \centering
    \includegraphics[page=7,width=\linewidth,height=0.90\textheight,keepaspectratio]{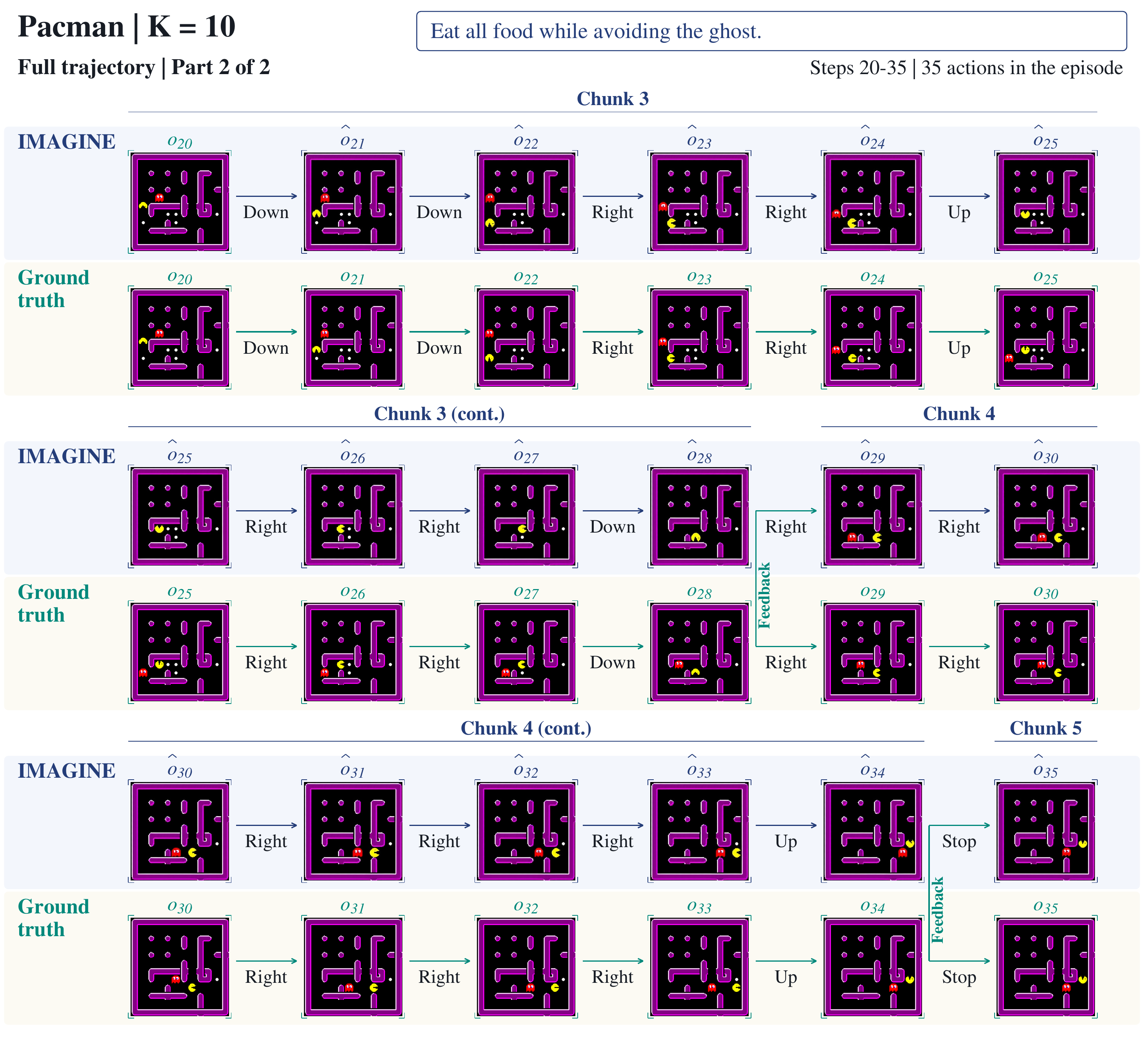}
    \par\vspace{0.5\baselineskip}\noindent\parbox{\linewidth}{Figure~\ref{fig:pacman_full_k10} (continued).
    \textbf{Pacman, $K=10$.} Steps 20--35, ending in success. The shared endpoint connects the two pages.}
\end{figure}

\clearpage
\begin{figure}[H]
    \centering
    \includegraphics[page=8,width=\linewidth,height=0.90\textheight,keepaspectratio]{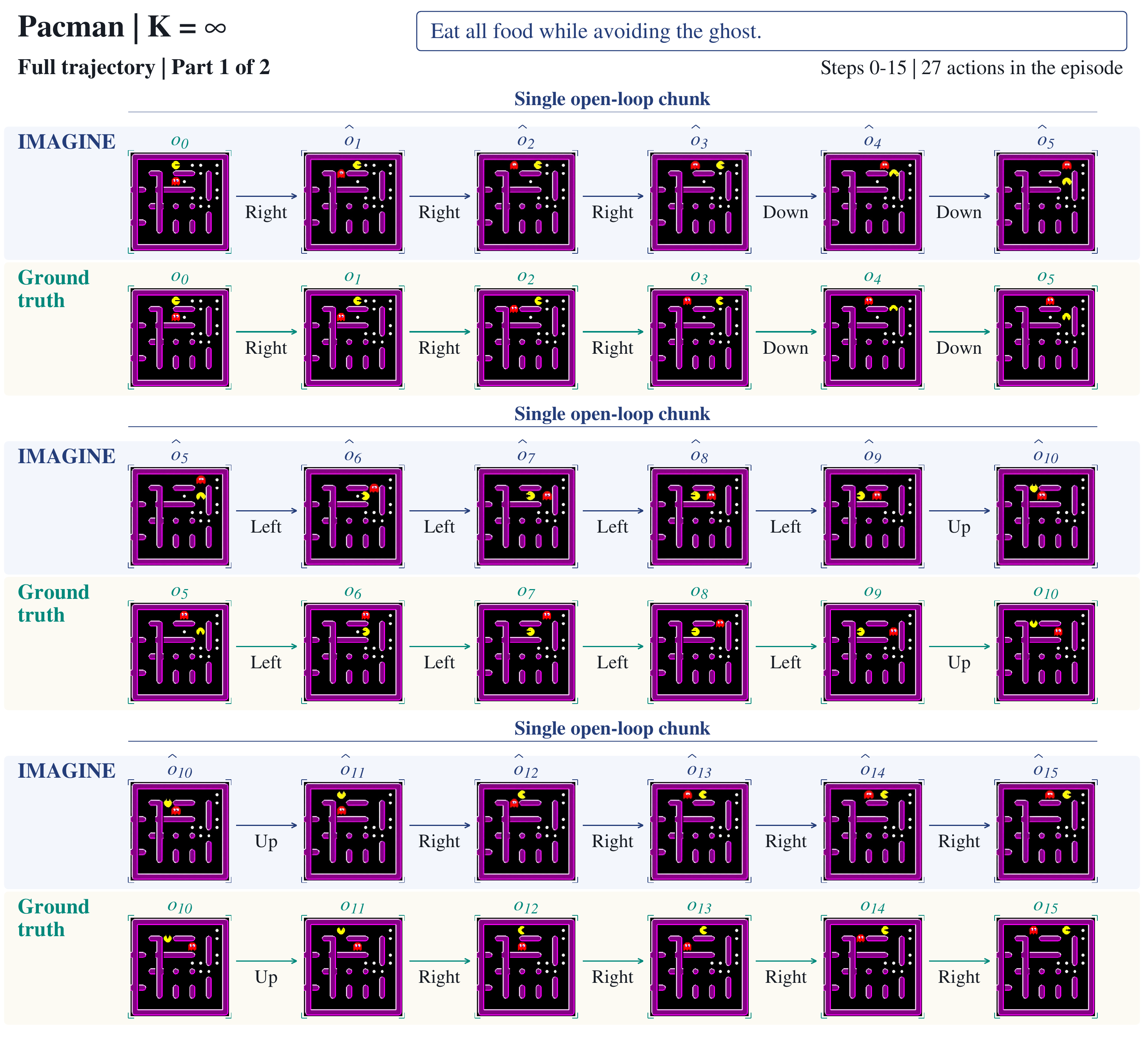}
    \caption{\textbf{Complete Pacman trajectory at $K=\infty$.}
    Selected successful Ordinary Hard episode; this page shows steps 0--15 of 27.
    The trajectory continues on the next page.
    The model receives only the initial observation, with no intermediate
    feedback. Ground-truth states are shown for comparison.}
    \label{fig:pacman_full_kinf}
    \label{fig:pacman_easy_to_hard}
\end{figure}

\clearpage
\begin{figure}[H]
    \centering
    \includegraphics[page=9,width=\linewidth,height=0.90\textheight,keepaspectratio]{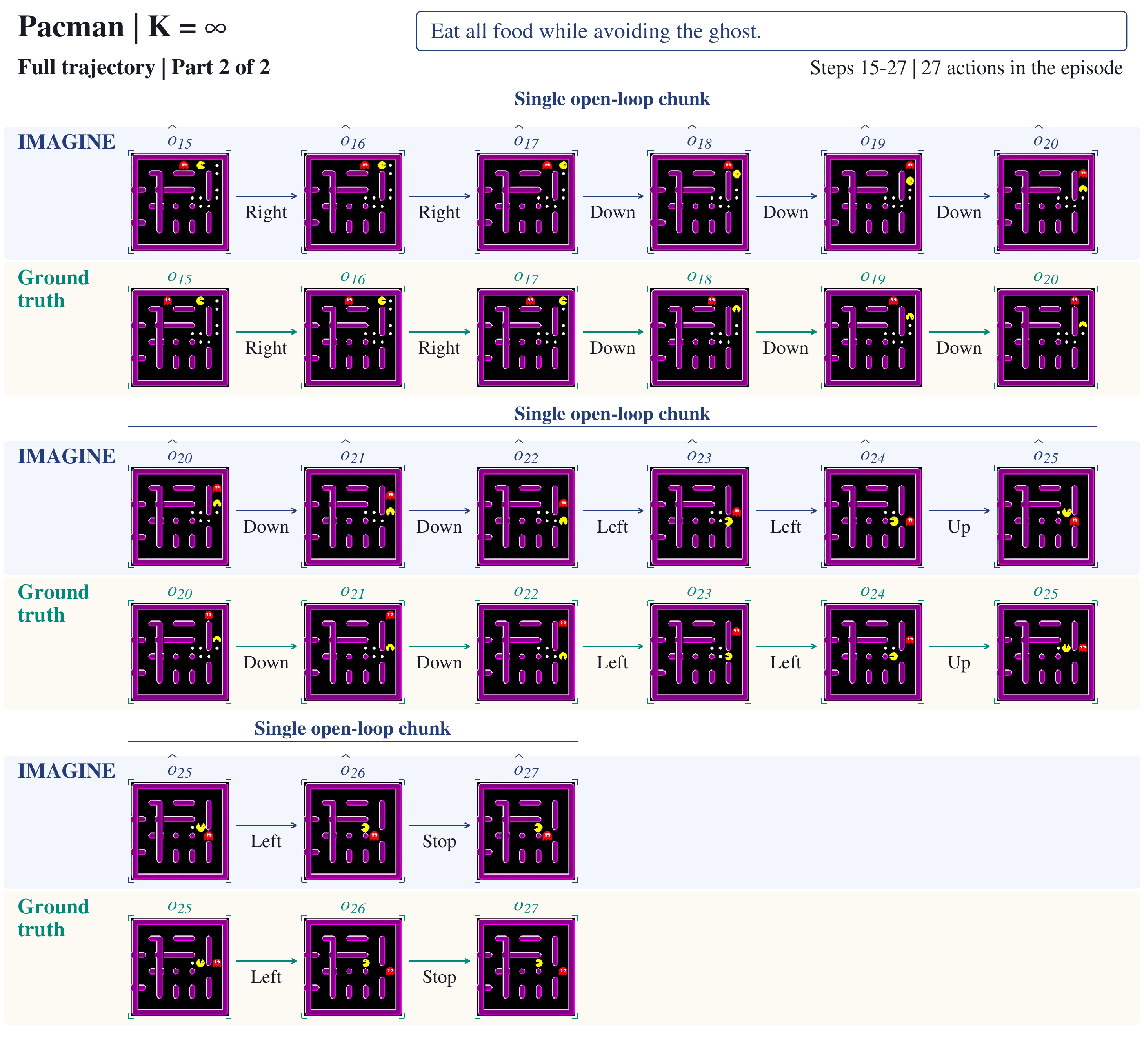}
    \par\vspace{0.5\baselineskip}\noindent\parbox{\linewidth}{Figure~\ref{fig:pacman_full_kinf} (continued).
    \textbf{Pacman, $K=\infty$.} Steps 15--27, ending in success.
    The shared endpoint connects the two pages; no new observation is supplied.}
\end{figure}

\clearpage
\subsection{PushT Rollouts}
\label{app:qualitative_pusht}
The starting-position shift is described in Appendix~\ref{app:pusht}.
\begin{figure}[H]
    \centering
    \includegraphics[page=1,width=\linewidth,height=0.90\textheight,keepaspectratio]{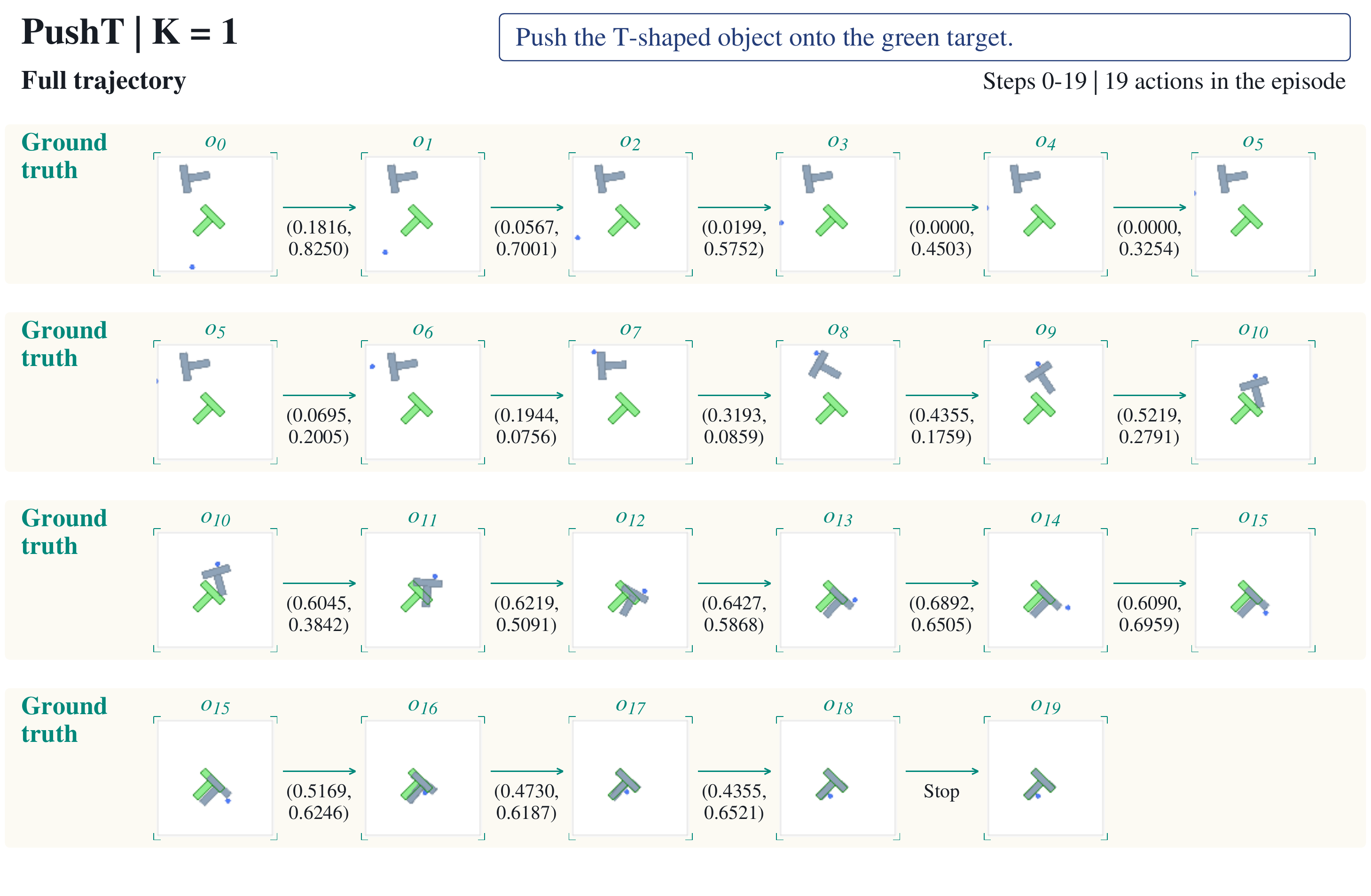}
    \caption{\textbf{Complete PushT trajectory at $K=1$.}
    Selected successful Far episode; all 19 actions and their states are shown.
    The reactive reference receives a real observation after every action. Actions give normalized target coordinates $(x,y)$.}
    \label{fig:pusht_full_k1}
    \label{fig:pusht_chunk_qualitative}
\end{figure}

\clearpage
\begin{figure}[H]
    \centering
    \includegraphics[page=2,width=\linewidth,height=0.90\textheight,keepaspectratio]{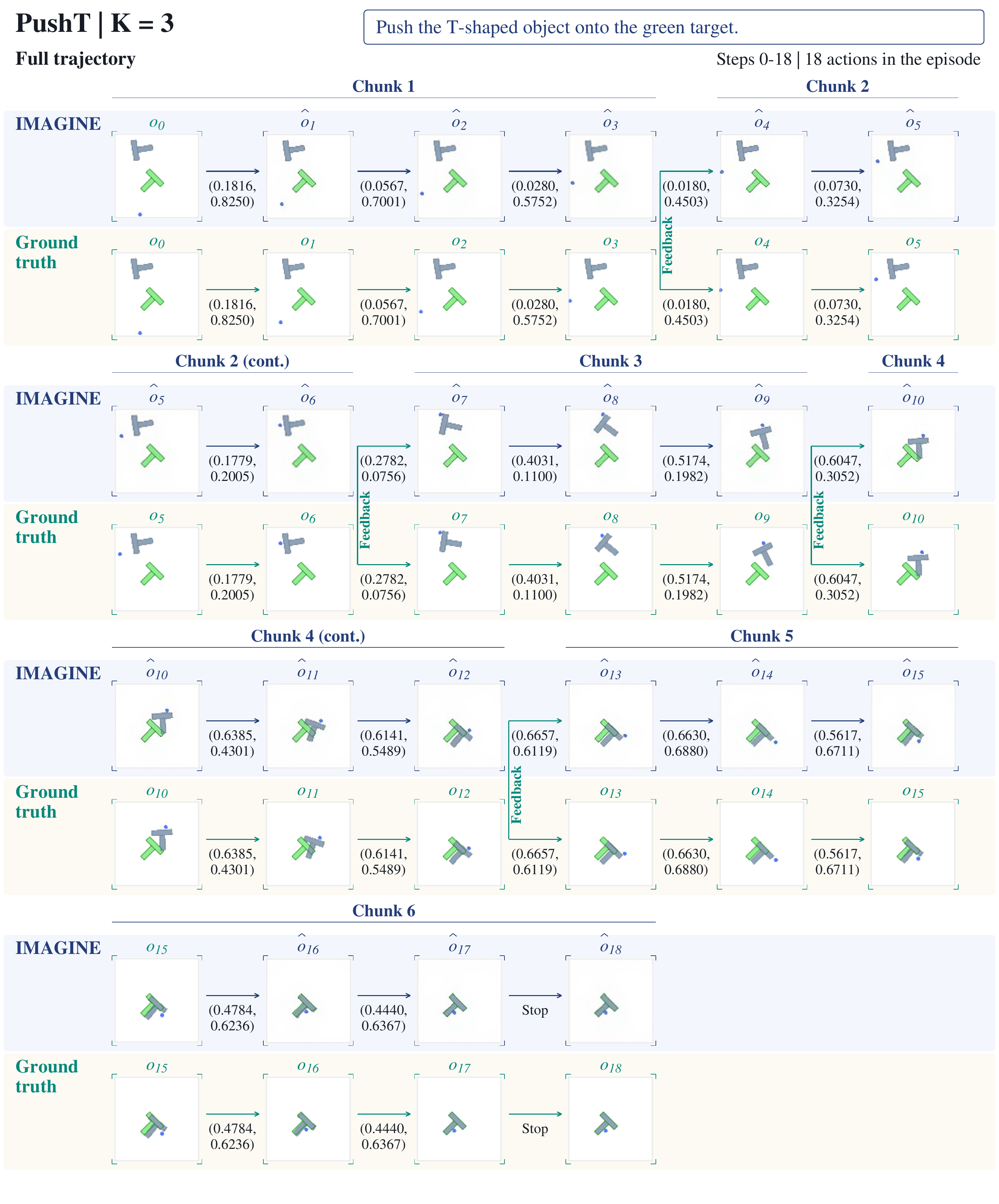}
    \caption{\textbf{Complete PushT trajectory at $K=3$.}
    Selected successful Far episode; all 18 actions and their states are shown.
    Blue rows show predictions; teal inputs and feedback arrows show the real
    observations supplied between chunks. Actions give normalized target coordinates $(x,y)$.}
    \label{fig:pusht_full_k3}
\end{figure}

\clearpage
\begin{figure}[H]
    \centering
    \includegraphics[page=3,width=\linewidth,height=0.90\textheight,keepaspectratio]{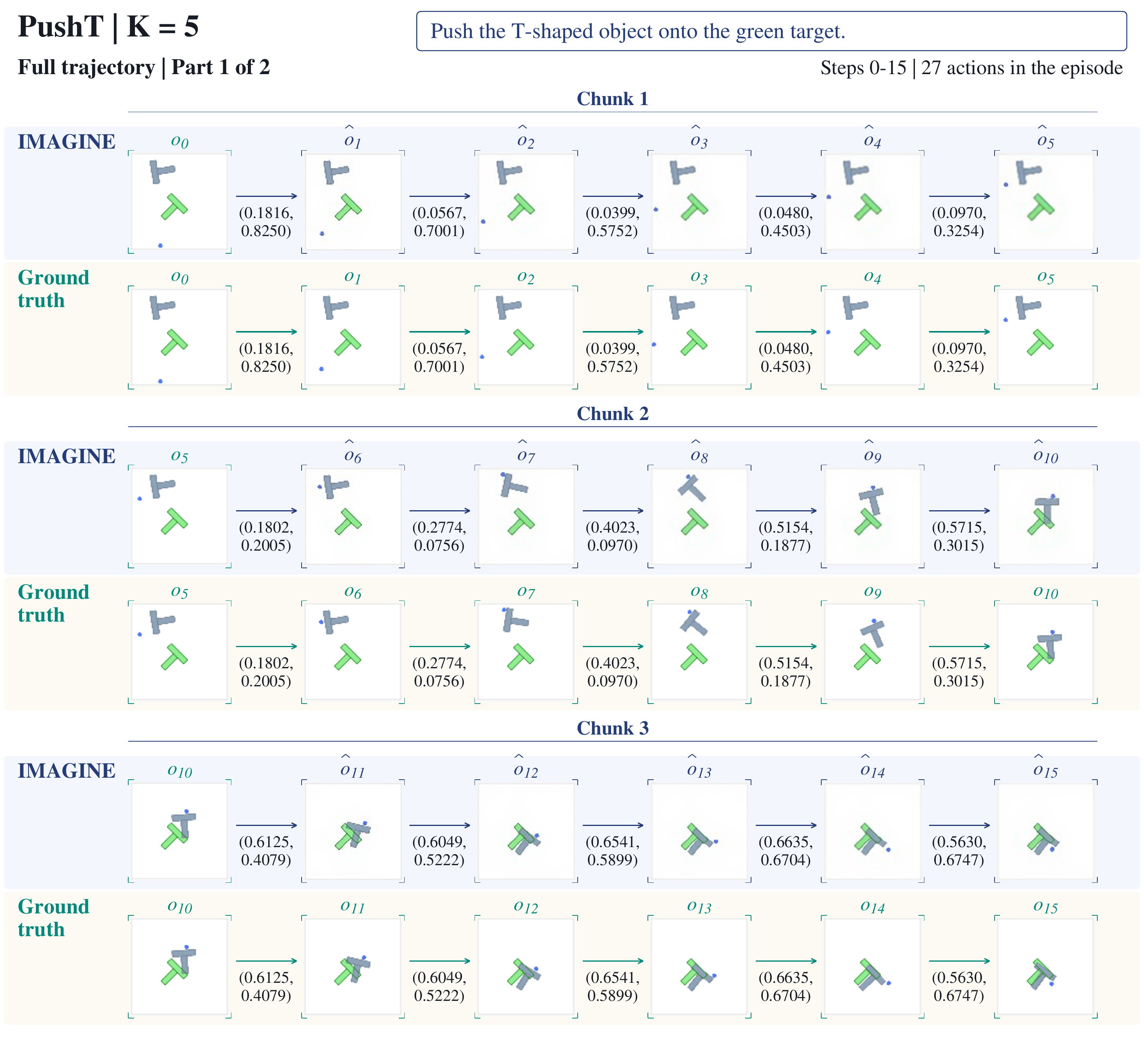}
    \caption{\textbf{Complete PushT trajectory at $K=5$.}
    Selected successful Far episode; this page shows steps 0--15 of 27.
    The trajectory continues on the next page.
    Blue rows show predictions; teal inputs and feedback arrows show the real
    observations supplied between chunks. Actions give normalized target coordinates $(x,y)$.}
    \label{fig:pusht_full_k5}
\end{figure}

\clearpage
\begin{figure}[H]
    \centering
    \includegraphics[page=4,width=\linewidth,height=0.90\textheight,keepaspectratio]{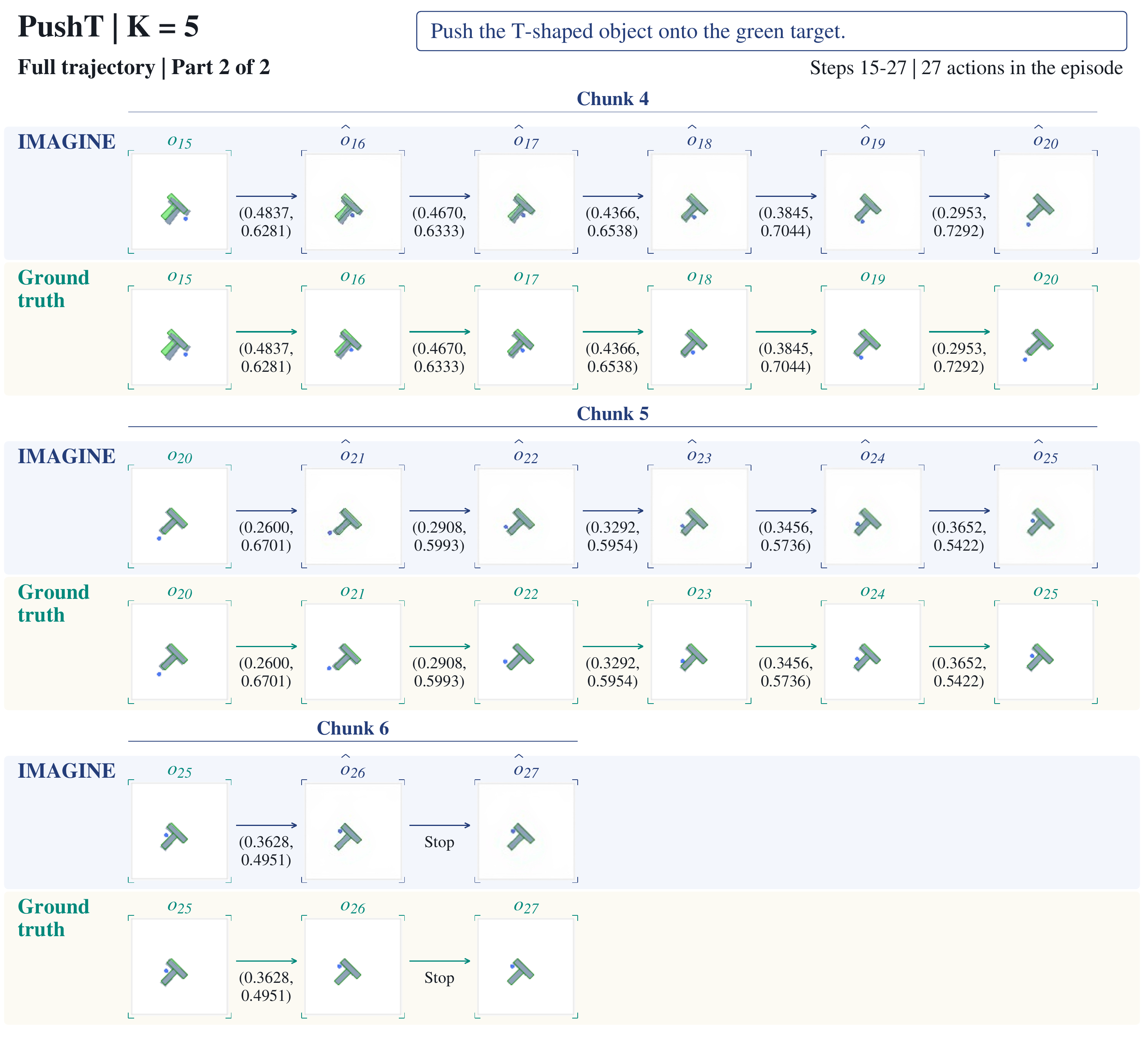}
    \par\vspace{0.5\baselineskip}\noindent\parbox{\linewidth}{Figure~\ref{fig:pusht_full_k5} (continued).
    \textbf{PushT, $K=5$.} Steps 15--27, ending in success. The shared endpoint connects the two pages.}
\end{figure}

\clearpage
\begin{figure}[H]
    \centering
    \includegraphics[page=5,width=\linewidth,height=0.90\textheight,keepaspectratio]{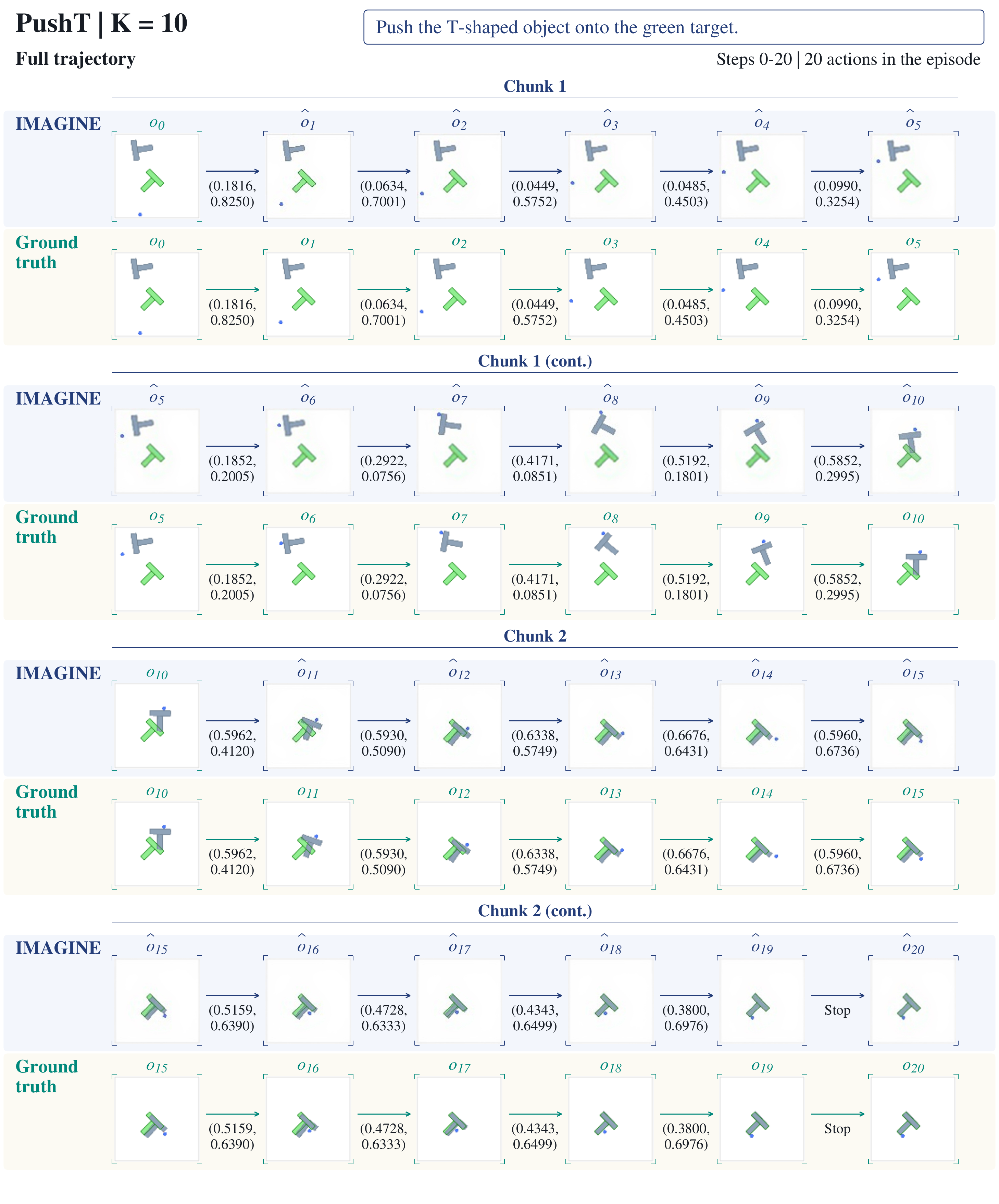}
    \caption{\textbf{Complete PushT trajectory at $K=10$.}
    Selected successful Far episode; all 20 actions and their states are shown.
    Blue rows show predictions; teal inputs and feedback arrows show the real
    observations supplied between chunks. Actions give normalized target coordinates $(x,y)$.}
    \label{fig:pusht_full_k10}
\end{figure}

\clearpage
\begin{figure}[H]
    \centering
    \includegraphics[page=6,width=\linewidth,height=0.90\textheight,keepaspectratio]{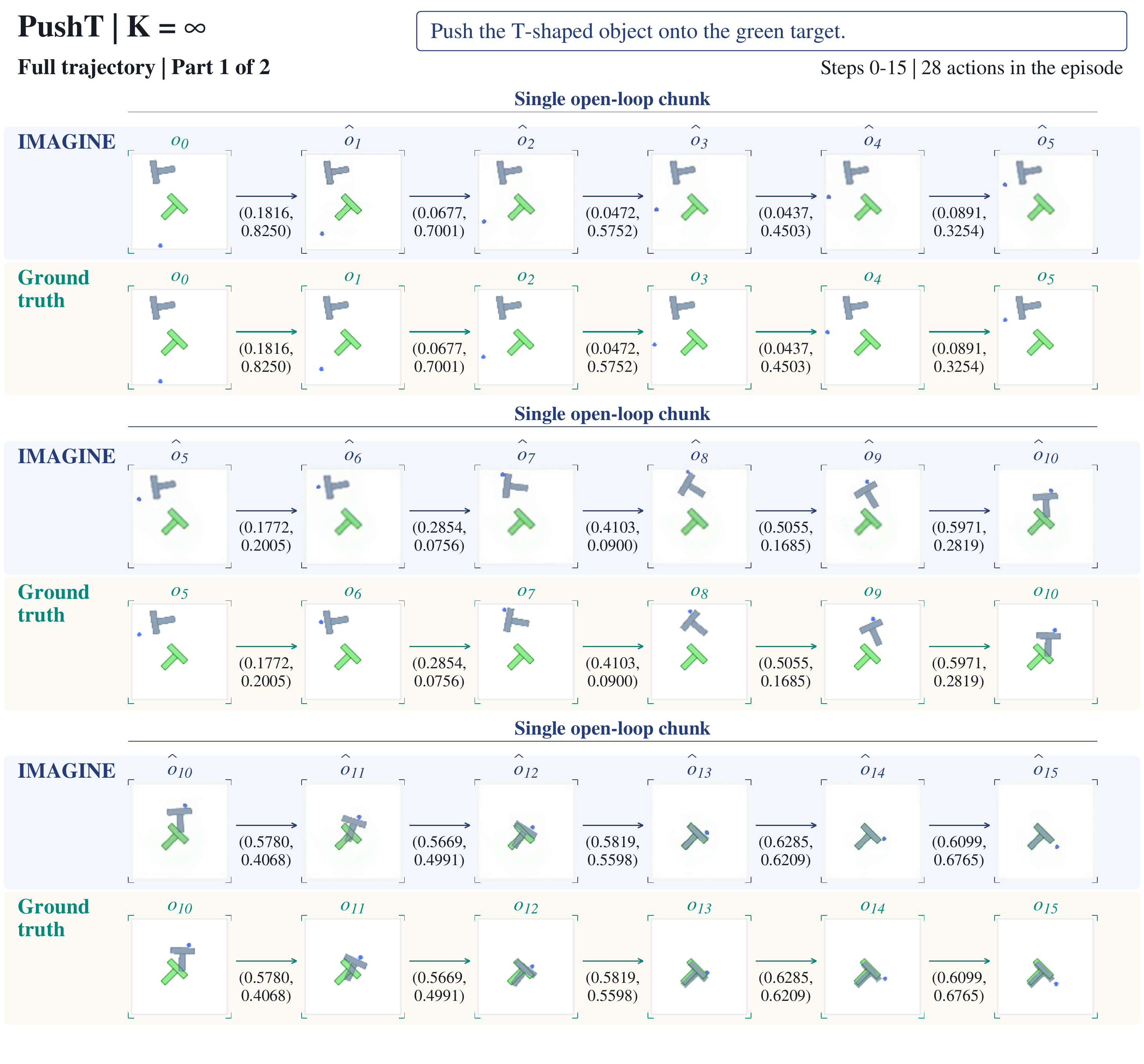}
    \caption{\textbf{Complete PushT trajectory at $K=\infty$.}
    Selected successful Far episode; this page shows steps 0--15 of 28.
    The trajectory continues on the next page.
    The model receives only the initial observation, with no intermediate
    feedback. Ground-truth states are shown for comparison.
    Actions give normalized target coordinates $(x,y)$.}
    \label{fig:pusht_full_kinf}
    \label{fig:pusht_easy_to_hard}
\end{figure}

\clearpage
\begin{figure}[H]
    \centering
    \includegraphics[page=7,width=\linewidth,height=0.90\textheight,keepaspectratio]{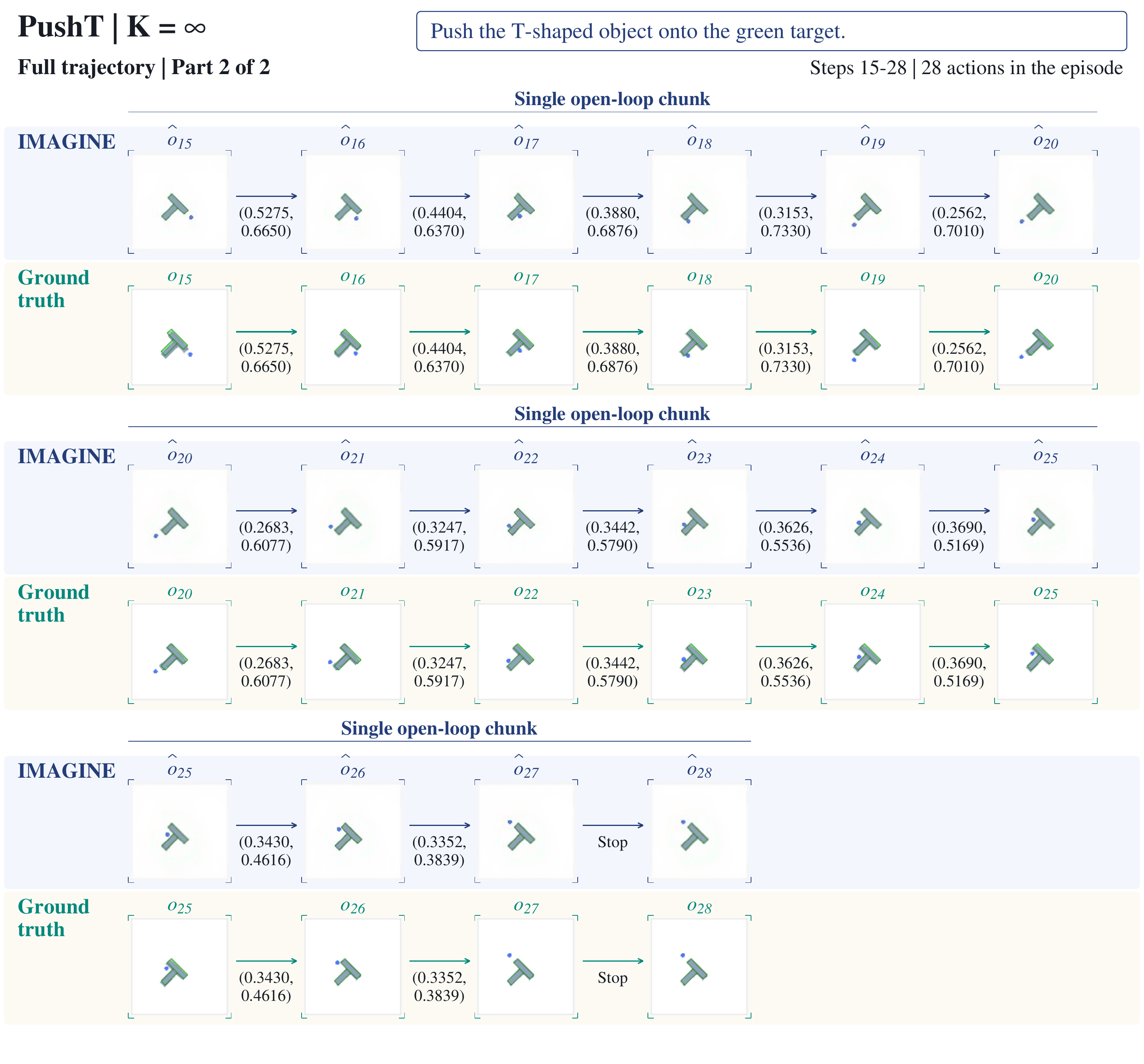}
    \par\vspace{0.5\baselineskip}\noindent\parbox{\linewidth}{Figure~\ref{fig:pusht_full_kinf} (continued).
    \textbf{PushT, $K=\infty$.} Steps 15--28, ending in success.
    The shared endpoint connects the two pages; no new observation is supplied.}
\end{figure}

\clearpage
\subsection{Rollouts for different reasoning formats}
\label{app:qualitative_reasoning_formats}
\begin{figure}[H]
    \centering
    \includegraphics[width=\linewidth,height=\textheight,keepaspectratio]{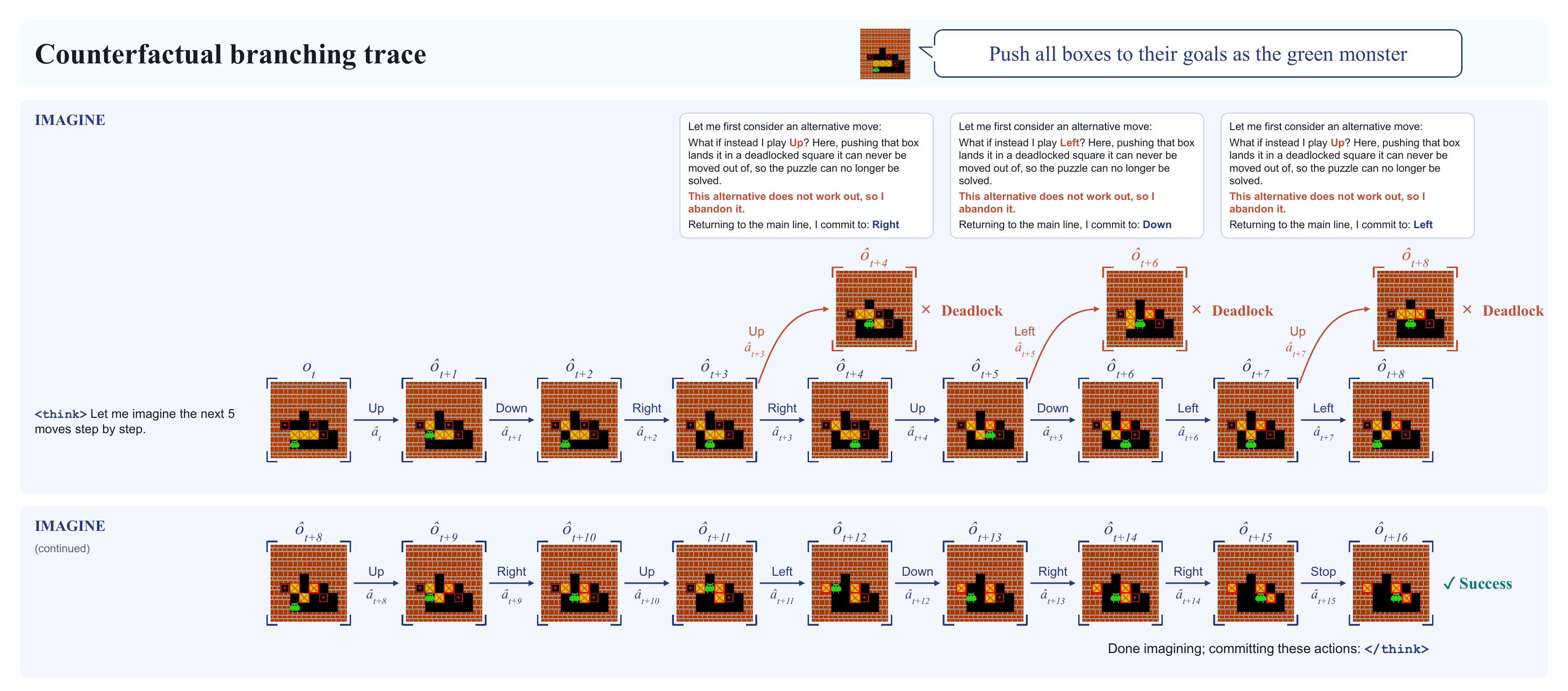}
    \caption{\textbf{Counterfactual Branching.}
    Example trace of counterfactual branching in Sokoban.}
\end{figure}

\begin{figure}[H]
    \centering
    \includegraphics[width=\linewidth,height=\textheight,keepaspectratio]{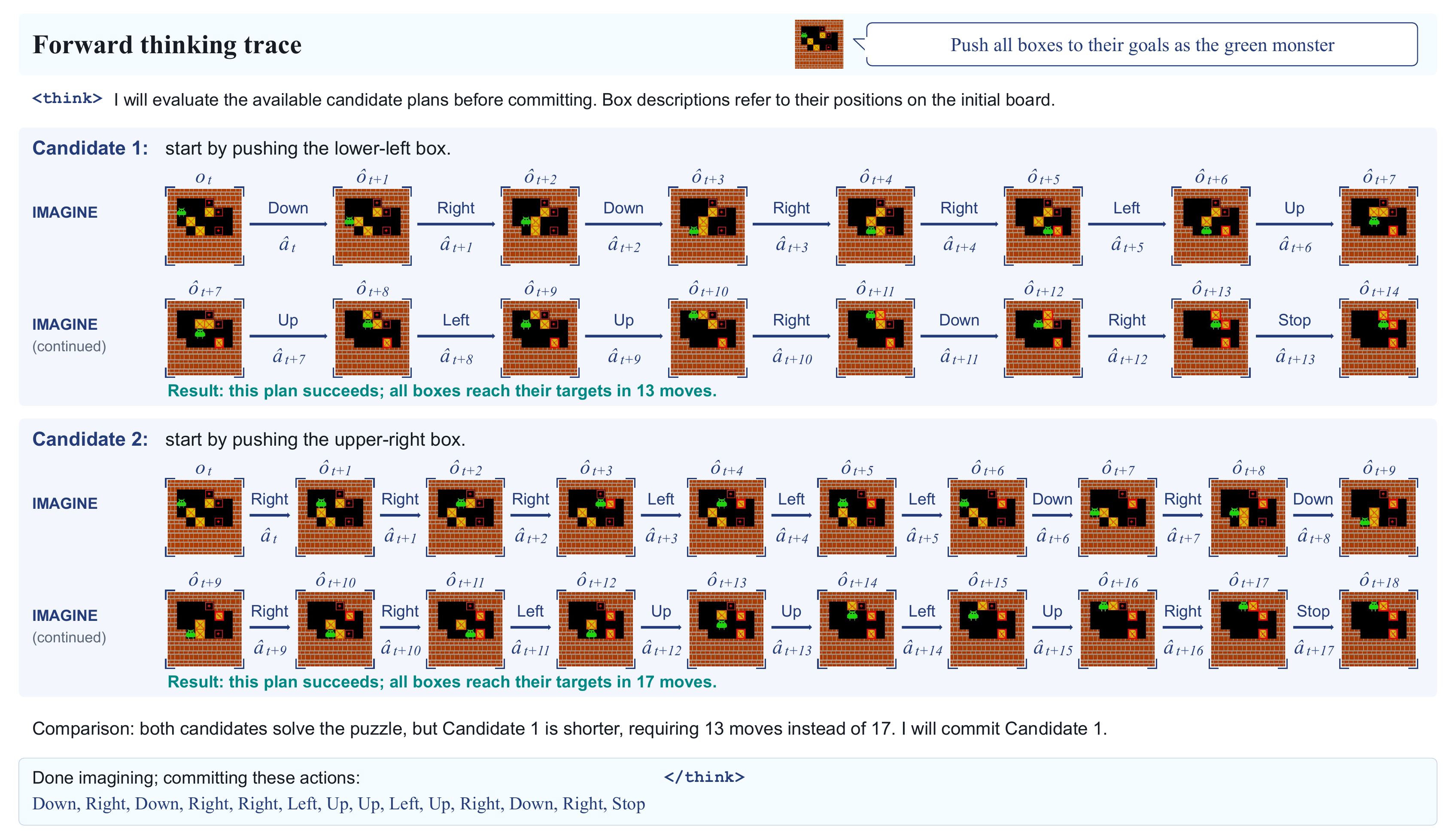}
    \caption{\textbf{Forward Thinking.}
    Example trace of Forward thinking in Sokoban.}
\end{figure}

\clearpage
\subsection{RL Trajectories across Checkpoints}
\label{app:rl_trajectories}
We show complete trajectories at RL50, RL100, RL150, and RL200 on the
three- and four-box Deadlock and Structural Fork sets. Each group uses one
fixed initial state across all four checkpoints. The Deadlock examples use
\methodname{} (Forward), and the Structural Fork examples use
\methodname{} (Branching).

Blue rows show model-generated states. Ground-truth rows show execution of
the committed plan; unexecuted candidates have no ground-truth row.
Orange panels show the alternatives rejected by the model and identify
where each branches from the main plan. Quoted assessments are the model's
own text. The recorded execution result is reported separately.
All saved generated frames, executed states, and actions are shown.

\begin{table}[H]
\centering
\small
\caption{\textbf{Outcomes of the illustrated RL episodes.}
Each row follows one selected initial state across four checkpoints.
Aggregate performance is reported in Figure~\ref{fig:rl_training_curve}.}
\label{tab:rl_illustrated_episodes}
\setlength{\tabcolsep}{5pt}
\begin{tabular}{@{}llrrrr@{}}
\toprule
Test set & Model & RL50 & RL100 & RL150 & RL200 \\
\midrule
3-box Deadlock & Forward & Fail & Success & Success & Success \\
4-box Deadlock & Forward & Fail & Fail & Success & Success \\
3-box Structural Fork & Branching & Fail & Success & Success & Success \\
4-box Structural Fork & Branching & Fail & Fail & Success & Success \\
\bottomrule
\end{tabular}
\end{table}

\begingroup
\setlength{\intextsep}{3pt plus 1pt minus 1pt}
\setlength{\abovecaptionskip}{4pt}
\begin{figure}[H]
    \centering
    \includegraphics[page=1,trim=0 1034bp 0 0bp,clip,width=\linewidth]{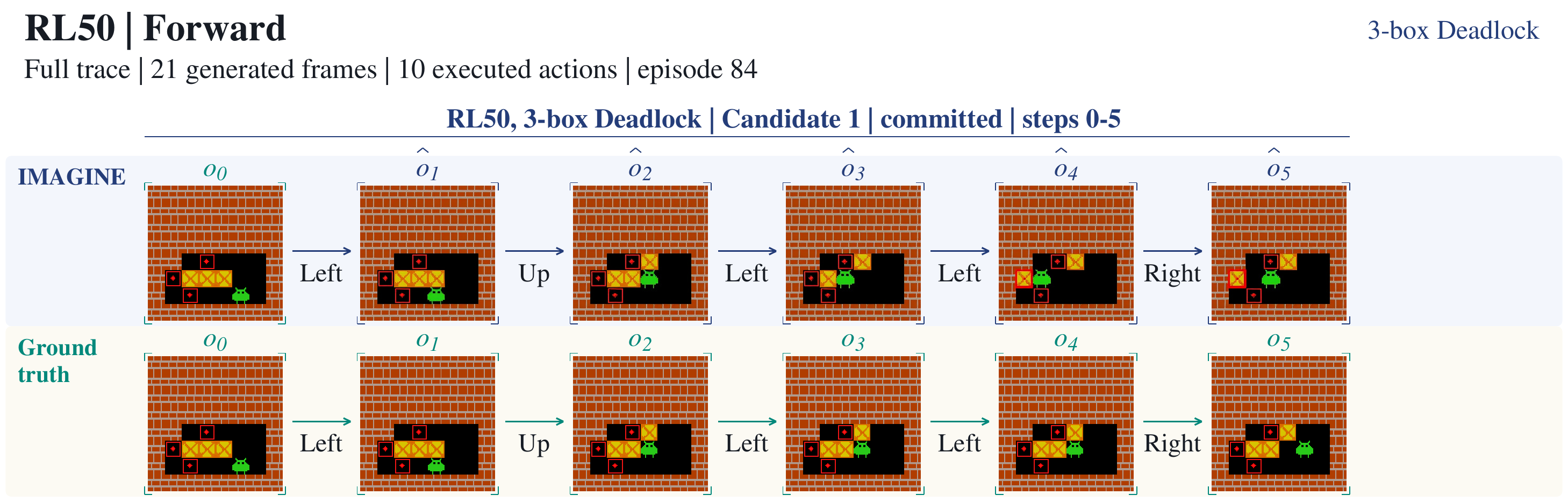}
\end{figure}

\begin{figure}[H]
    \centering
    \includegraphics[page=1,trim=0 624bp 0 448bp,clip,width=\linewidth]{figures/appendix_rl_3dead_traces.pdf}
\end{figure}

\begin{figure}[H]
    \centering
    \includegraphics[page=1,trim=0 416bp 0 858bp,clip,width=\linewidth]{figures/appendix_rl_3dead_traces.pdf}
\end{figure}

\begin{figure}[H]
    \centering
    \includegraphics[page=1,trim=0 0bp 0 1066bp,clip,width=\linewidth]{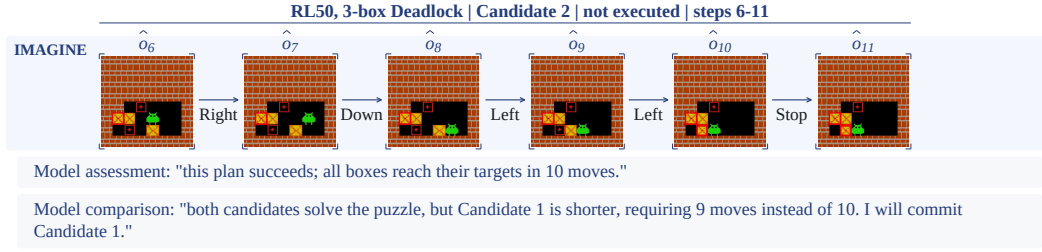}
    \caption{\textbf{Forward at RL50 on three-box Deadlock.} Complete trace for episode 84; recorded execution ends in failure.}
    \label{fig:rl_3dead_fwd3k_rl50}
\end{figure}

\begin{figure}[H]
    \centering
    \includegraphics[page=2,trim=0 438bp 0 0bp,clip,width=\linewidth]{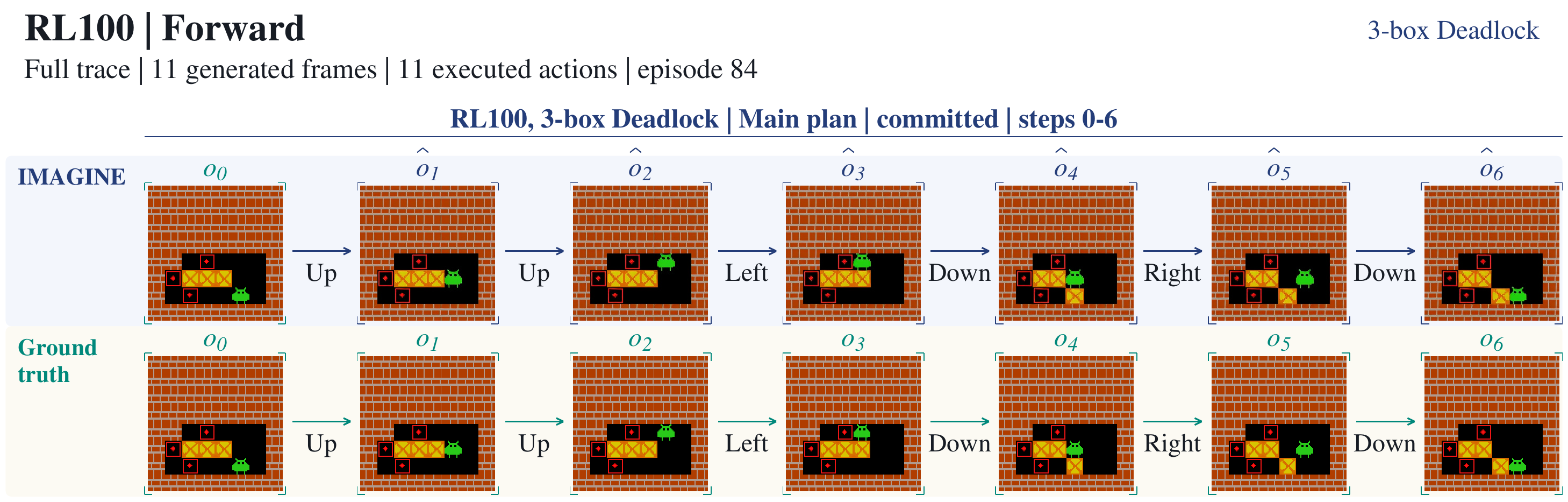}
\end{figure}

\begin{figure}[H]
    \centering
    \includegraphics[page=2,trim=0 0bp 0 448bp,clip,width=\linewidth]{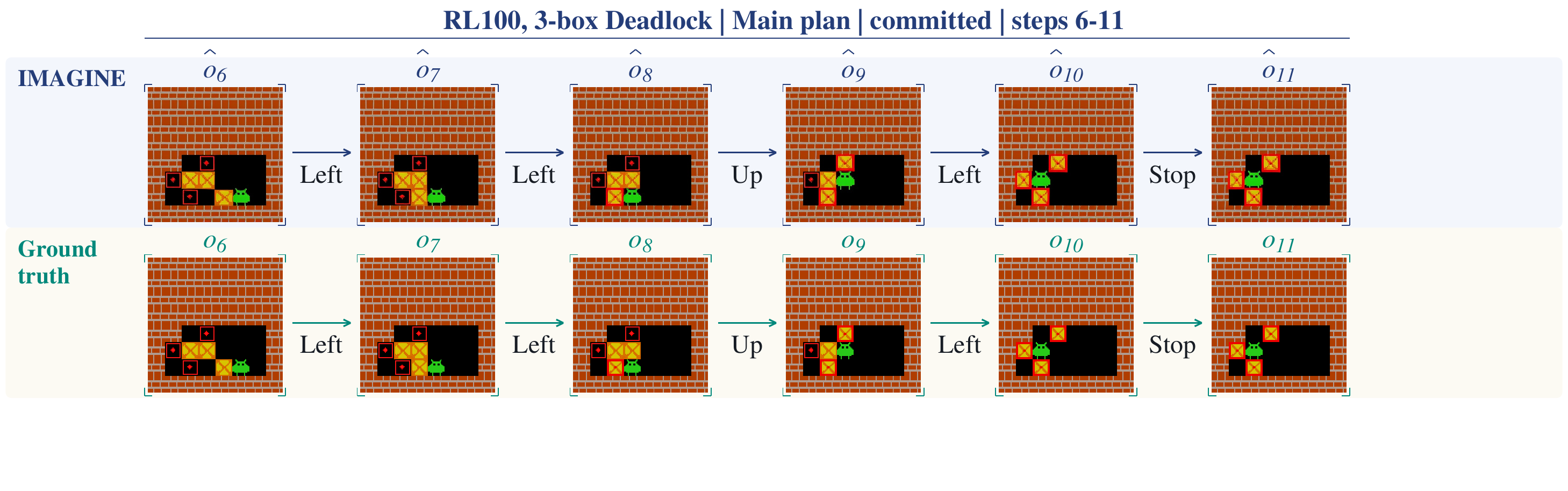}
    \caption{\textbf{Forward at RL100 on three-box Deadlock.} Complete trace for episode 84; recorded execution ends in success.}
    \label{fig:rl_3dead_fwd3k_rl100}
\end{figure}

\begin{figure}[H]
    \centering
    \includegraphics[page=3,trim=0 438bp 0 0bp,clip,width=\linewidth]{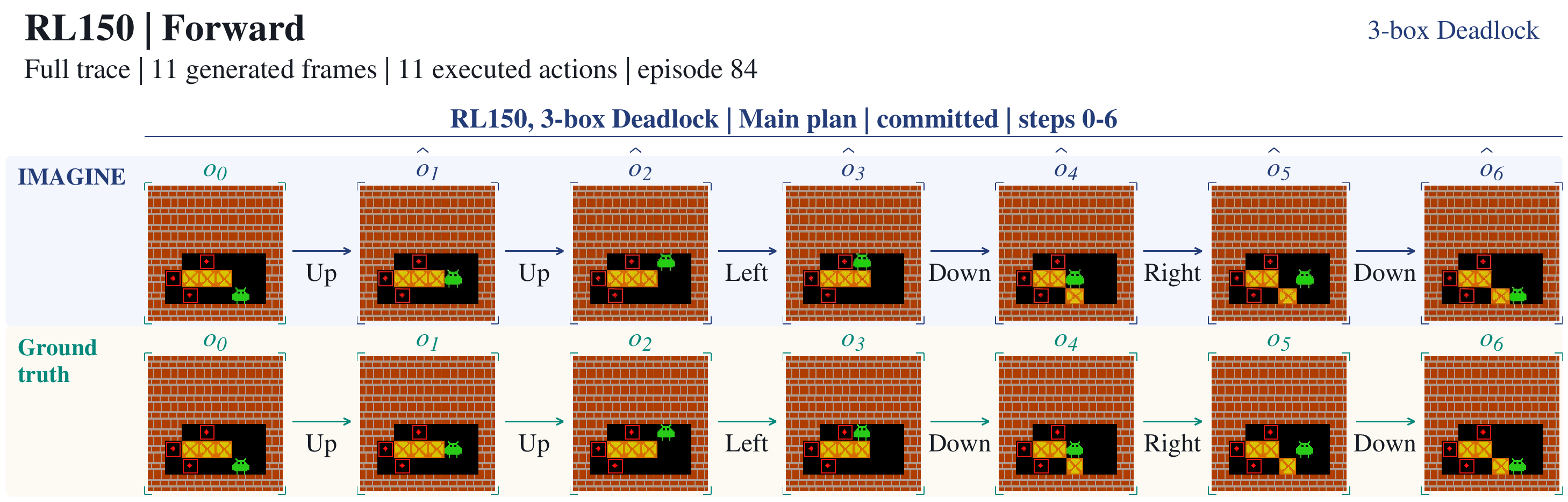}
\end{figure}

\begin{figure}[H]
    \centering
    \includegraphics[page=3,trim=0 0bp 0 448bp,clip,width=\linewidth]{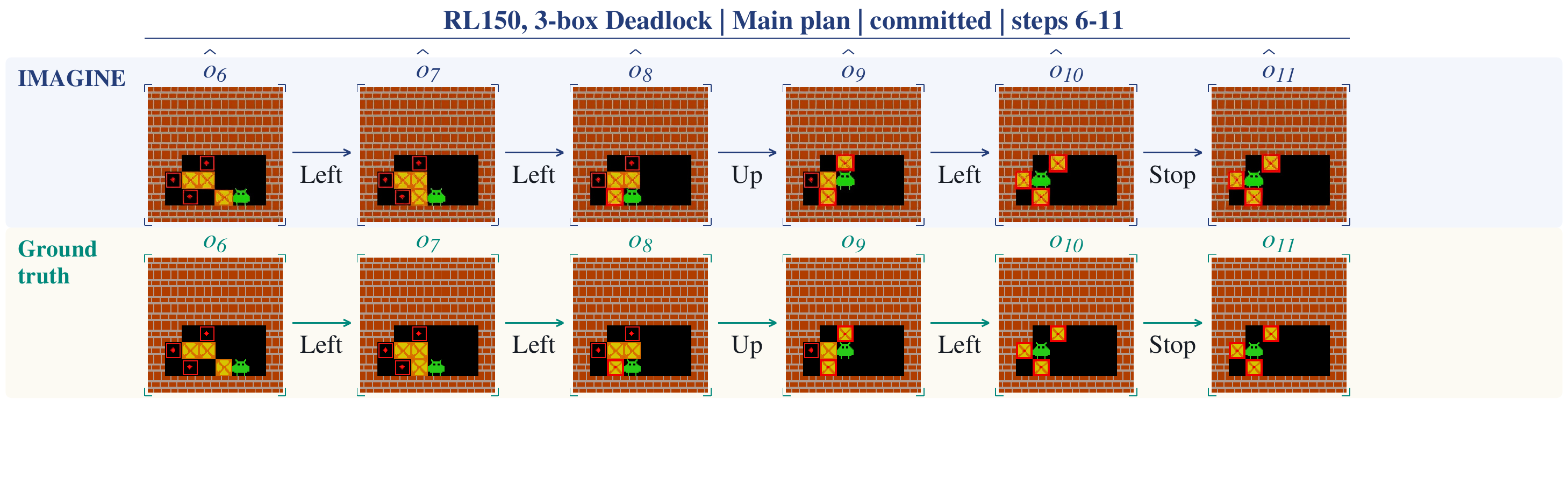}
    \caption{\textbf{Forward at RL150 on three-box Deadlock.} Complete trace for episode 84; recorded execution ends in success.}
    \label{fig:rl_3dead_fwd3k_rl150}
\end{figure}

\begin{figure}[H]
    \centering
    \includegraphics[page=4,trim=0 438bp 0 0bp,clip,width=\linewidth]{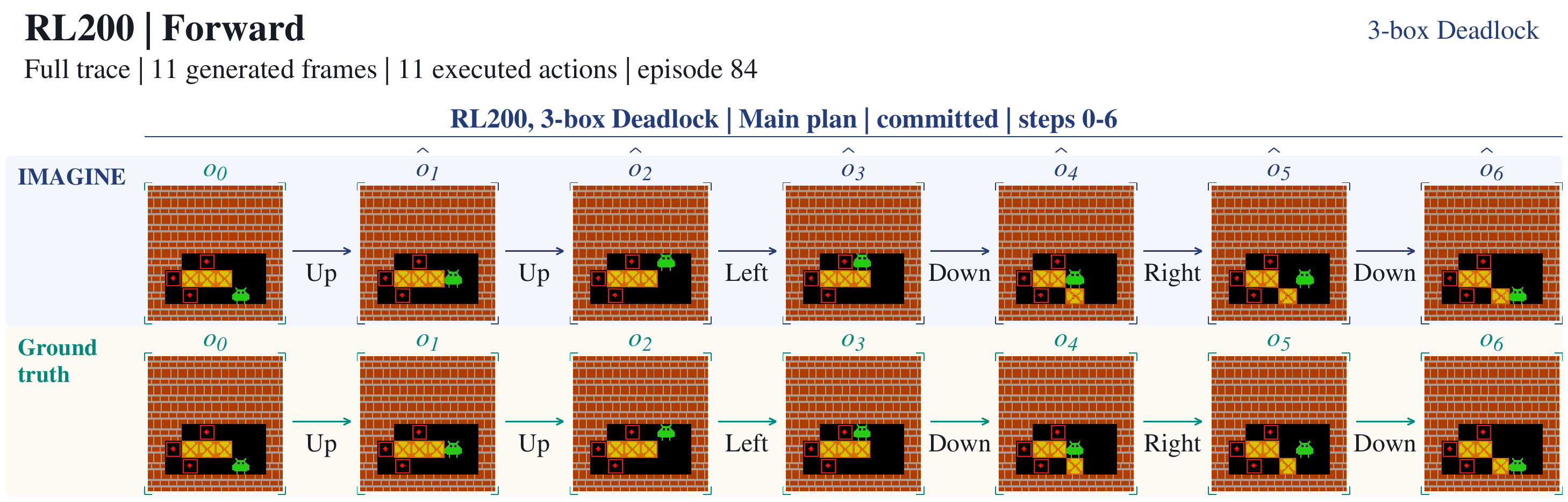}
\end{figure}

\begin{figure}[H]
    \centering
    \includegraphics[page=4,trim=0 0bp 0 448bp,clip,width=\linewidth]{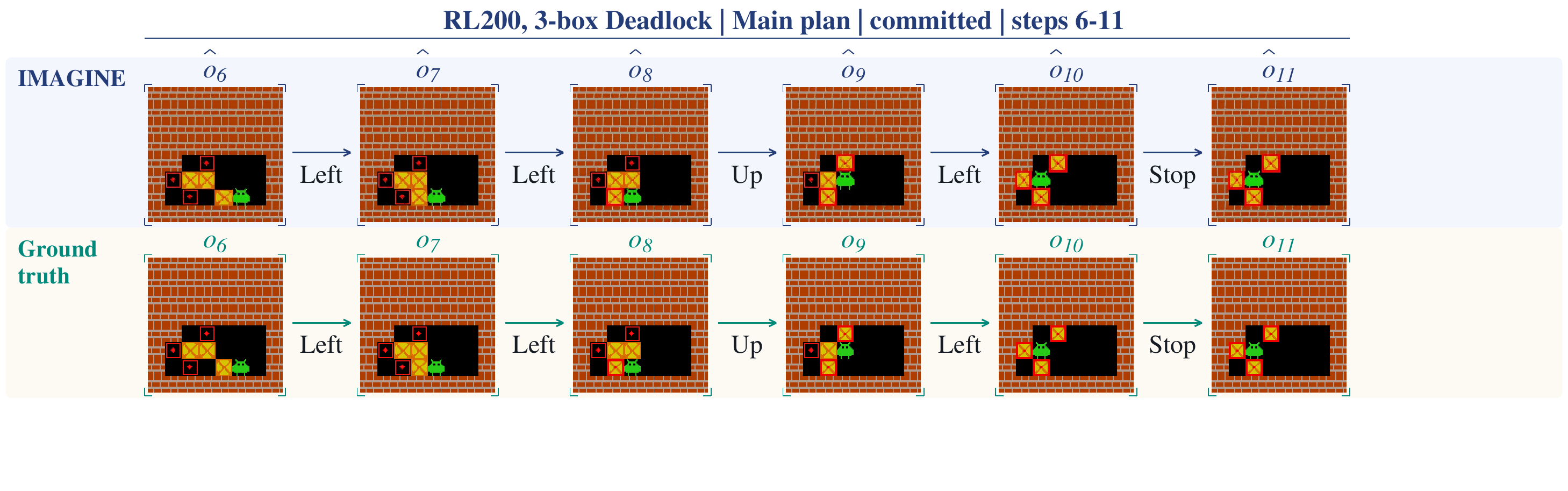}
    \caption{\textbf{Forward at RL200 on three-box Deadlock.} Complete trace for episode 84; recorded execution ends in success.}
    \label{fig:rl_3dead_fwd3k_rl200}
\end{figure}

\begin{figure}[H]
    \centering
    \includegraphics[page=1,trim=0 1810bp 0 0bp,clip,width=\linewidth]{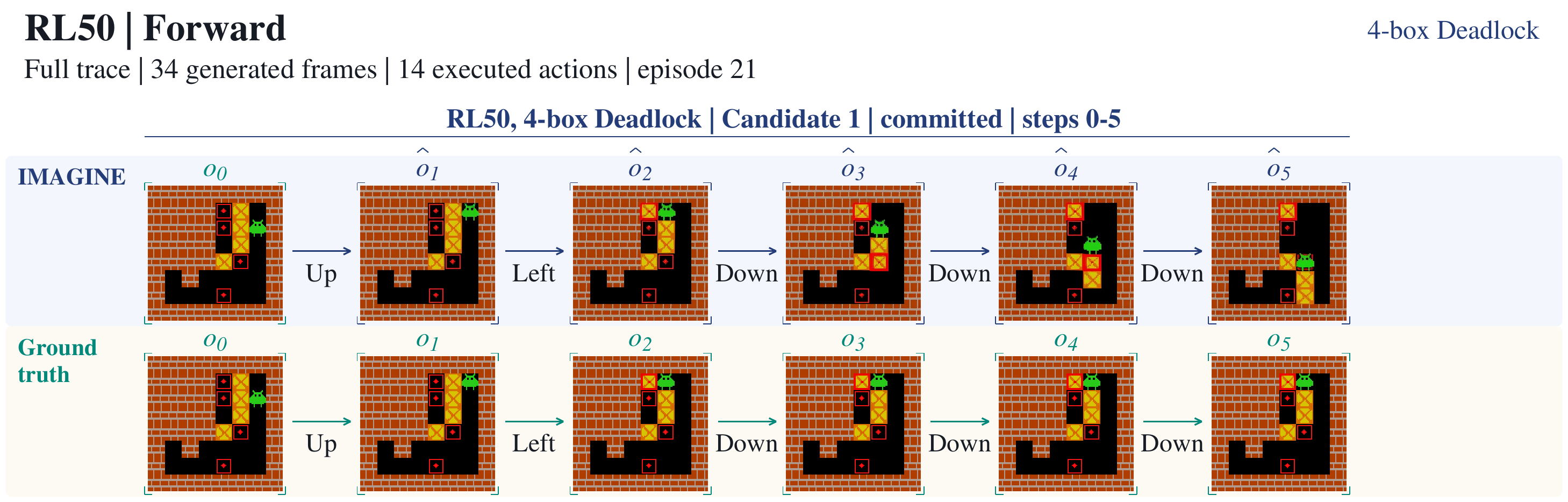}
\end{figure}

\begin{figure}[H]
    \centering
    \includegraphics[page=1,trim=0 1450bp 0 448bp,clip,width=\linewidth]{figures/appendix_rl_4dead_traces.pdf}
\end{figure}

\begin{figure}[H]
    \centering
    \includegraphics[page=1,trim=0 1040bp 0 808bp,clip,width=\linewidth]{figures/appendix_rl_4dead_traces.pdf}
\end{figure}

\begin{figure}[H]
    \centering
    \includegraphics[page=1,trim=0 832bp 0 1218bp,clip,width=\linewidth]{figures/appendix_rl_4dead_traces.pdf}
\end{figure}

\begin{figure}[H]
    \centering
    \includegraphics[page=1,trim=0 624bp 0 1426bp,clip,width=\linewidth]{figures/appendix_rl_4dead_traces.pdf}
\end{figure}

\begin{figure}[H]
    \centering
    \includegraphics[page=1,trim=0 416bp 0 1634bp,clip,width=\linewidth]{figures/appendix_rl_4dead_traces.pdf}
\end{figure}

\begin{figure}[H]
    \centering
    \includegraphics[page=1,trim=0 0bp 0 1842bp,clip,width=\linewidth]{figures/appendix_rl_4dead_traces.pdf}
    \caption{\textbf{Forward at RL50 on four-box Deadlock.} Complete trace for episode 21; recorded execution ends in failure.}
    \label{fig:rl_4dead_fwd3k_rl50}
\end{figure}

\begin{figure}[H]
    \centering
    \includegraphics[page=2,trim=0 1962bp 0 0bp,clip,width=\linewidth]{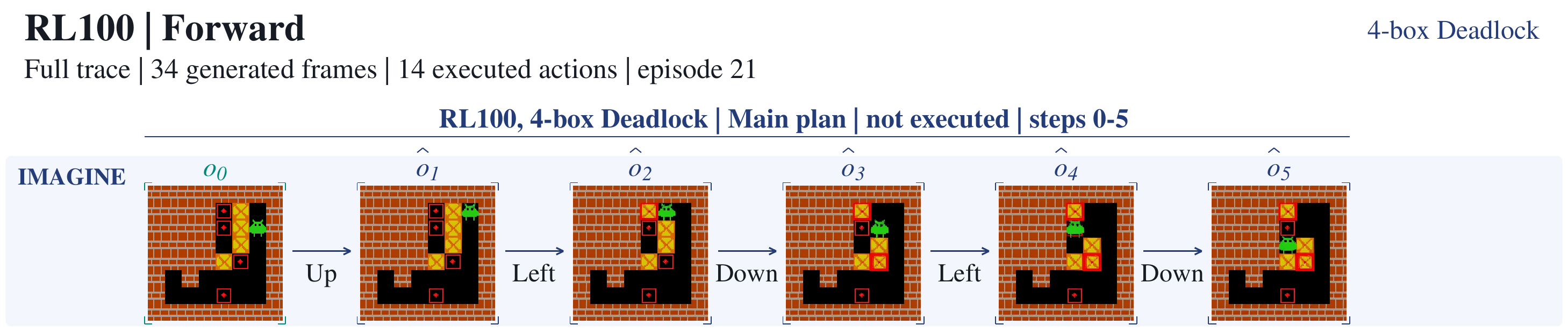}
\end{figure}

\begin{figure}[H]
    \centering
    \includegraphics[page=2,trim=0 1754bp 0 296bp,clip,width=\linewidth]{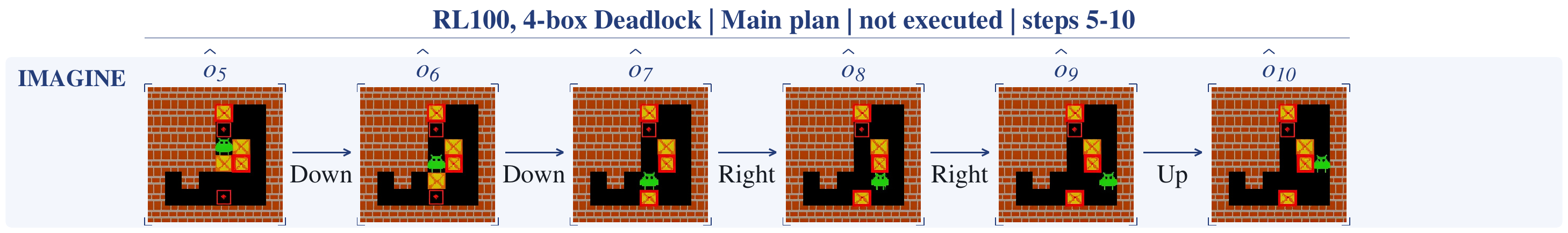}
\end{figure}

\begin{figure}[H]
    \centering
    \includegraphics[page=2,trim=0 1546bp 0 504bp,clip,width=\linewidth]{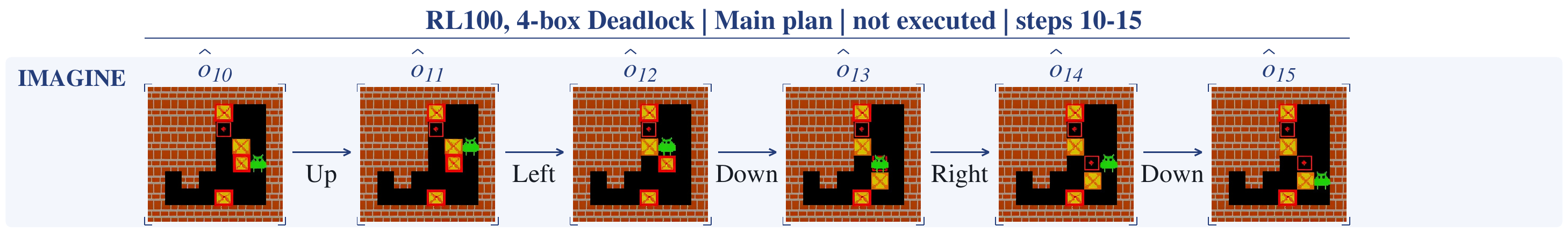}
\end{figure}

\begin{figure}[H]
    \centering
    \includegraphics[page=2,trim=0 1288bp 0 712bp,clip,width=\linewidth]{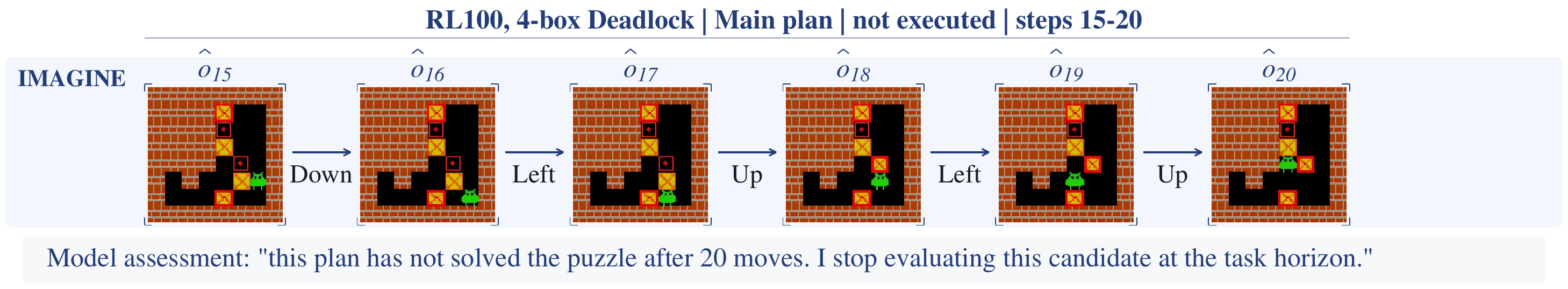}
\end{figure}

\begin{figure}[H]
    \centering
    \includegraphics[page=2,trim=0 928bp 0 970bp,clip,width=\linewidth]{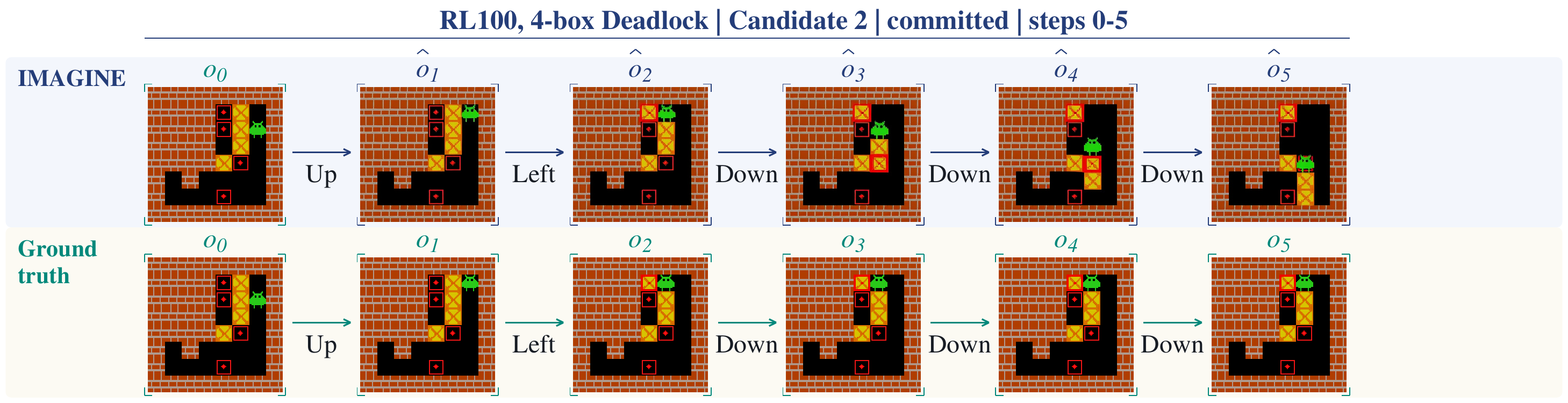}
\end{figure}

\begin{figure}[H]
    \centering
    \includegraphics[page=2,trim=0 568bp 0 1330bp,clip,width=\linewidth]{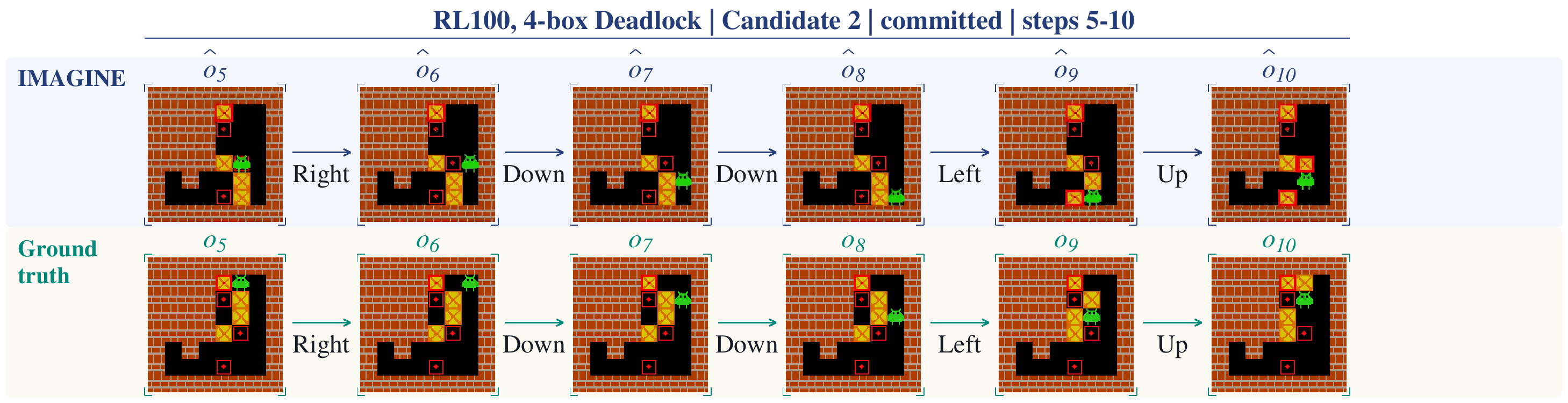}
\end{figure}

\begin{figure}[H]
    \centering
    \includegraphics[page=2,trim=0 0bp 0 1690bp,clip,width=\linewidth]{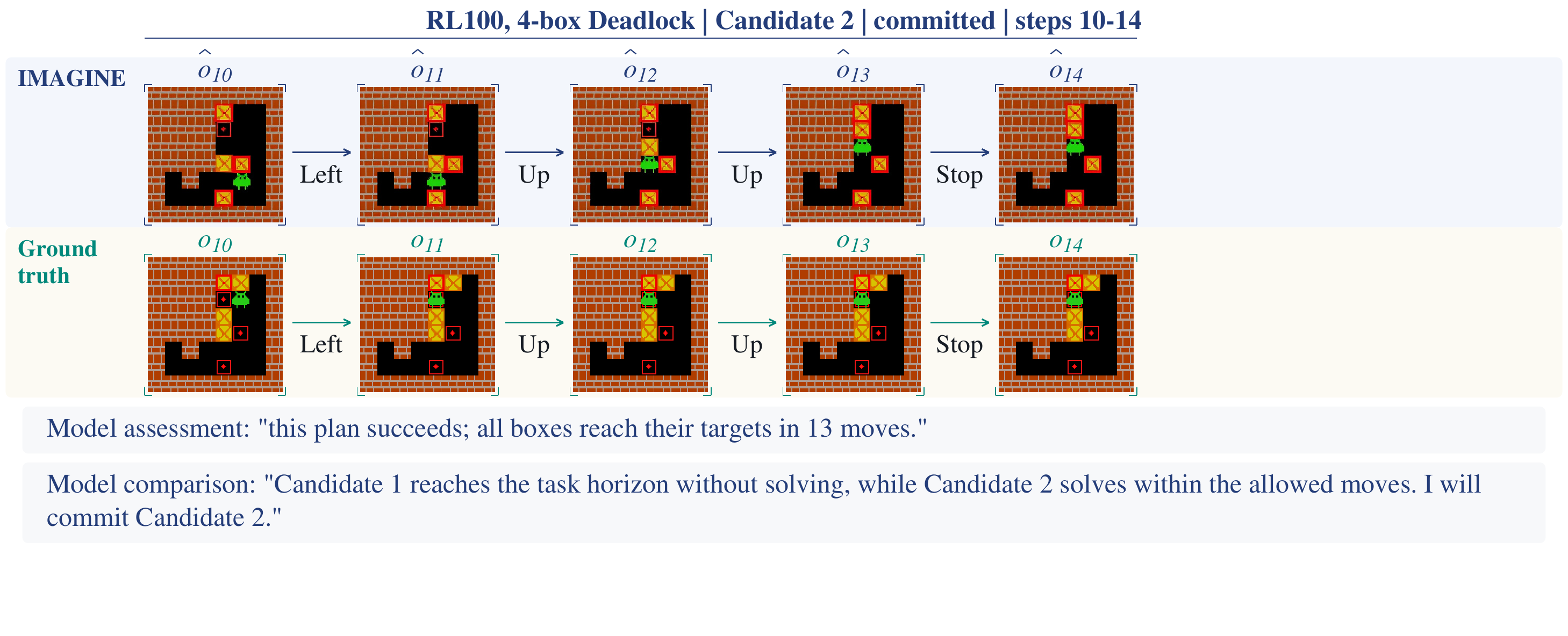}
    \caption{\textbf{Forward at RL100 on four-box Deadlock.} Complete trace for episode 21; recorded execution ends in failure.}
    \label{fig:rl_4dead_fwd3k_rl100}
\end{figure}

\begin{figure}[H]
    \centering
    \includegraphics[page=3,trim=0 438bp 0 0bp,clip,width=\linewidth]{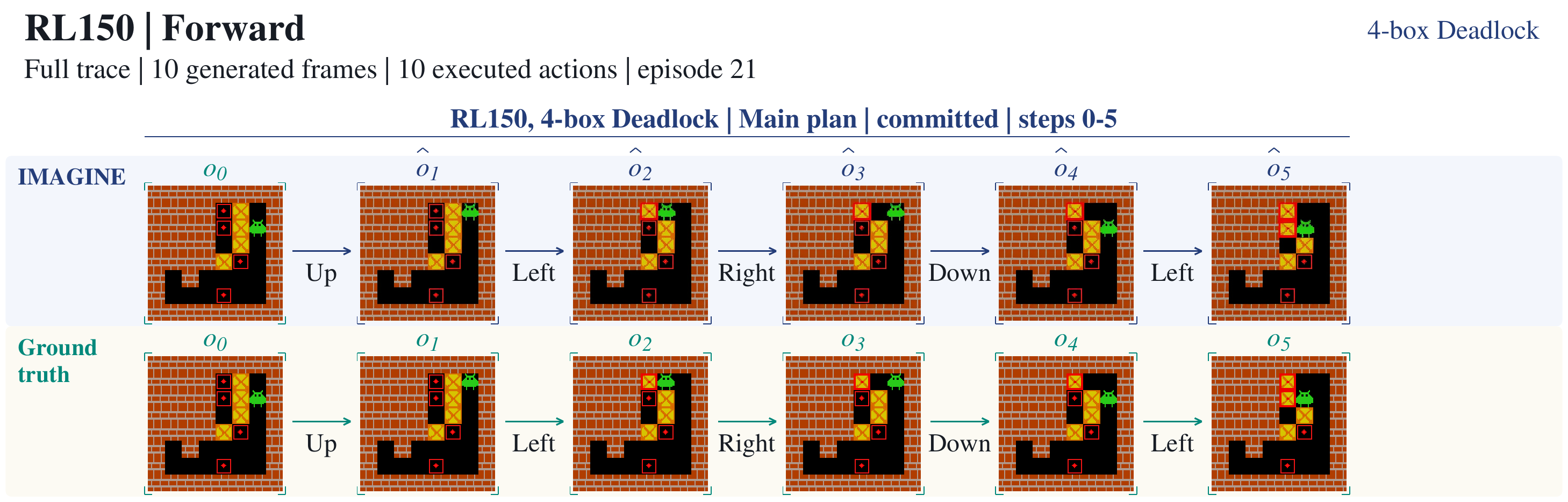}
\end{figure}

\begin{figure}[H]
    \centering
    \includegraphics[page=3,trim=0 0bp 0 448bp,clip,width=\linewidth]{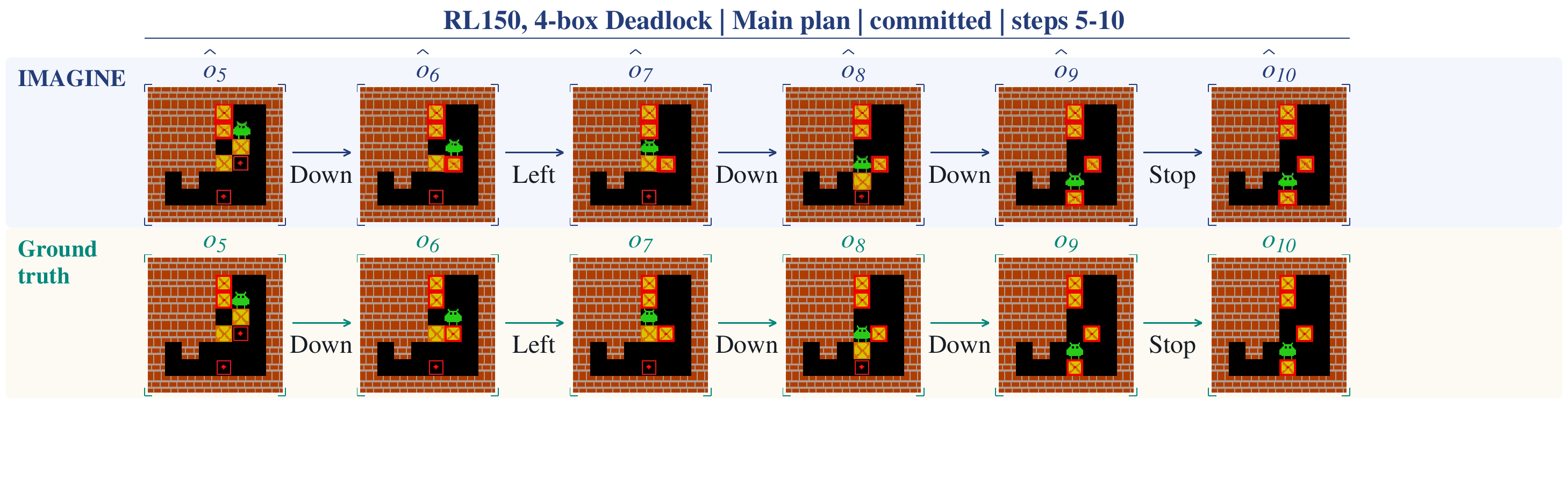}
    \caption{\textbf{Forward at RL150 on four-box Deadlock.} Complete trace for episode 21; recorded execution ends in success.}
    \label{fig:rl_4dead_fwd3k_rl150}
\end{figure}

\begin{figure}[H]
    \centering
    \includegraphics[page=4,trim=0 438bp 0 0bp,clip,width=\linewidth]{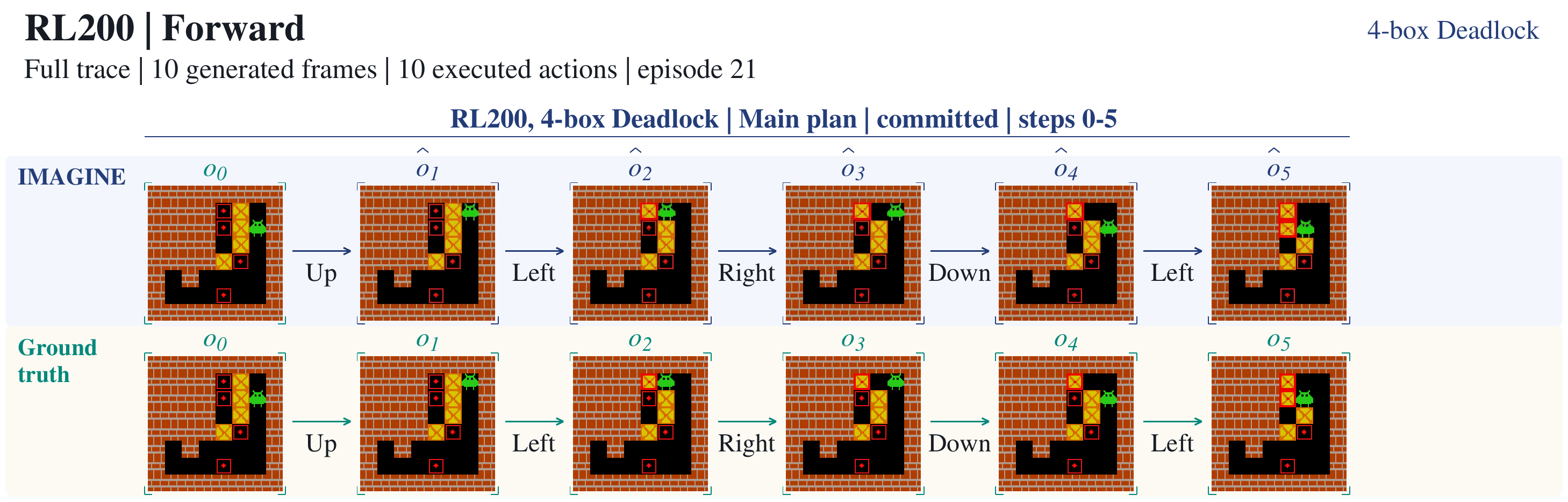}
\end{figure}

\begin{figure}[H]
    \centering
    \includegraphics[page=4,trim=0 0bp 0 448bp,clip,width=\linewidth]{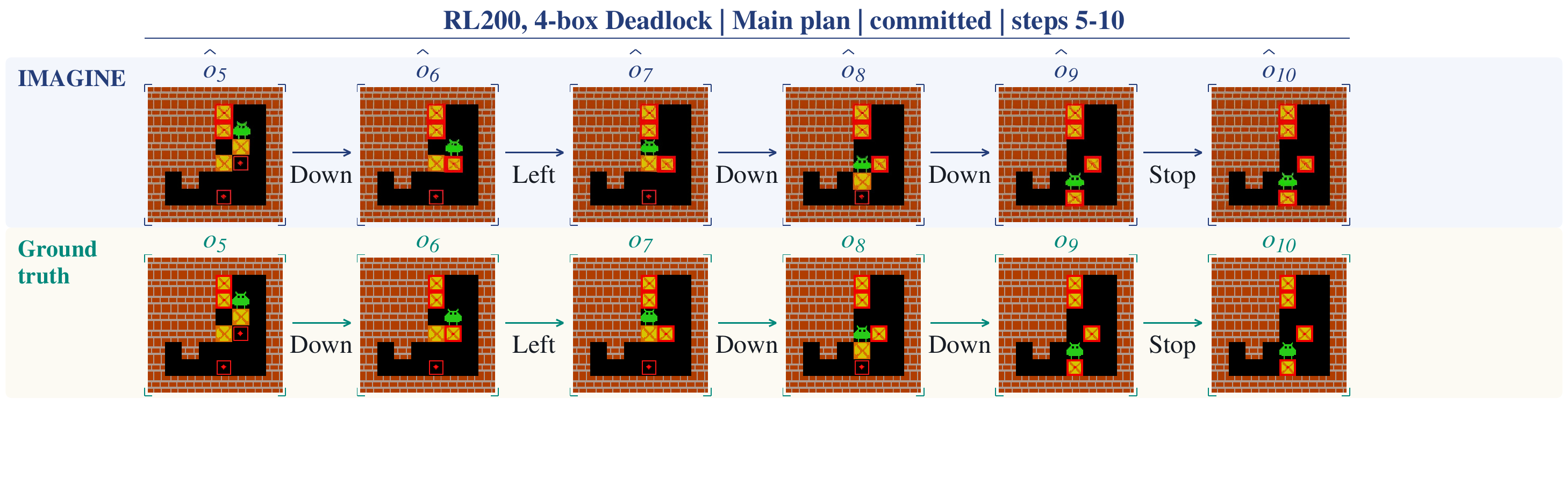}
    \caption{\textbf{Forward at RL200 on four-box Deadlock.} Complete trace for episode 21; recorded execution ends in success.}
    \label{fig:rl_4dead_fwd3k_rl200}
\end{figure}

\begin{figure}[H]
    \centering
    \includegraphics[page=1,trim=0 668bp 0 0bp,clip,width=\linewidth]{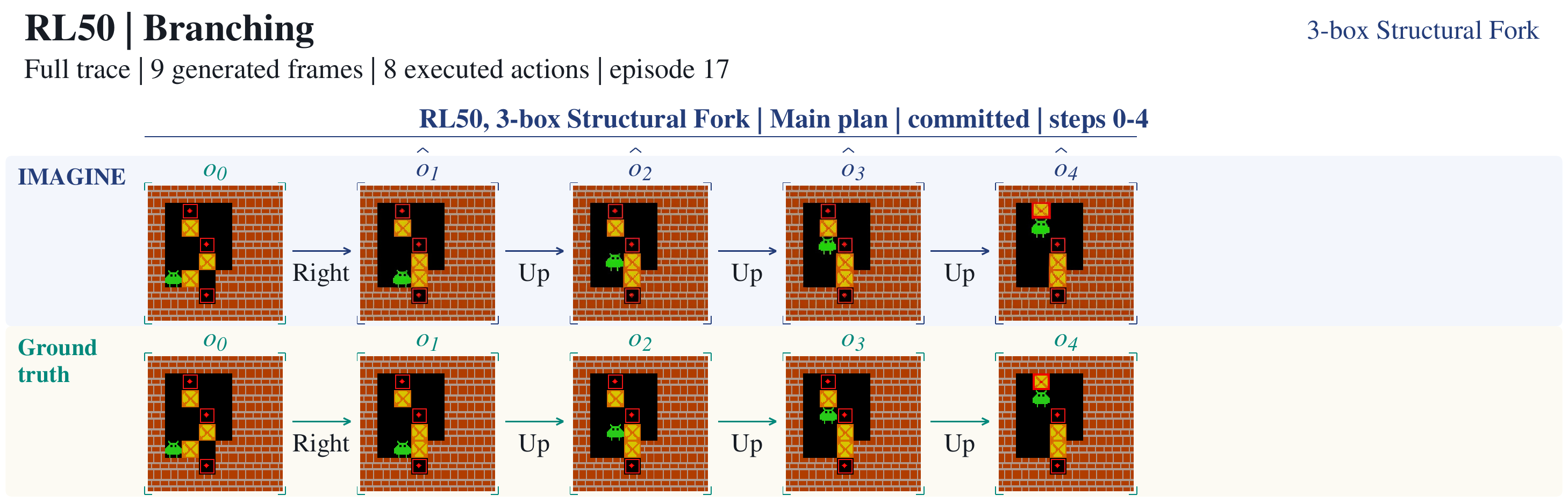}
\end{figure}

\begin{figure}[H]
    \centering
    \includegraphics[page=1,trim=0 308bp 0 448bp,clip,width=\linewidth]{figures/appendix_rl_3fork_traces.pdf}
\end{figure}

\begin{figure}[H]
    \centering
    \includegraphics[page=1,trim=0 0bp 0 808bp,clip,width=\linewidth]{figures/appendix_rl_3fork_traces.pdf}
    \caption{\textbf{Branching at RL50 on three-box Structural Fork.} Complete trace for episode 17; recorded execution ends in failure.}
    \label{fig:rl_3fork_bra3k_rl50}
\end{figure}

\begin{figure}[H]
    \centering
    \includegraphics[page=2,trim=0 1028bp 0 0bp,clip,width=\linewidth]{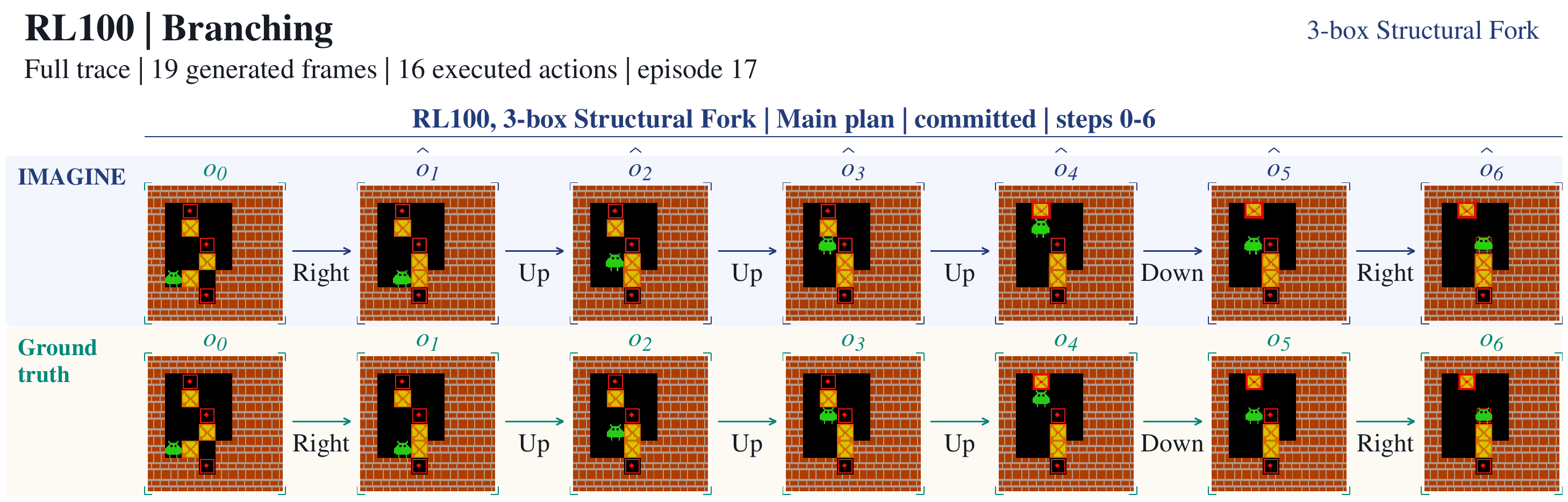}
\end{figure}

\begin{figure}[H]
    \centering
    \includegraphics[page=2,trim=0 668bp 0 448bp,clip,width=\linewidth]{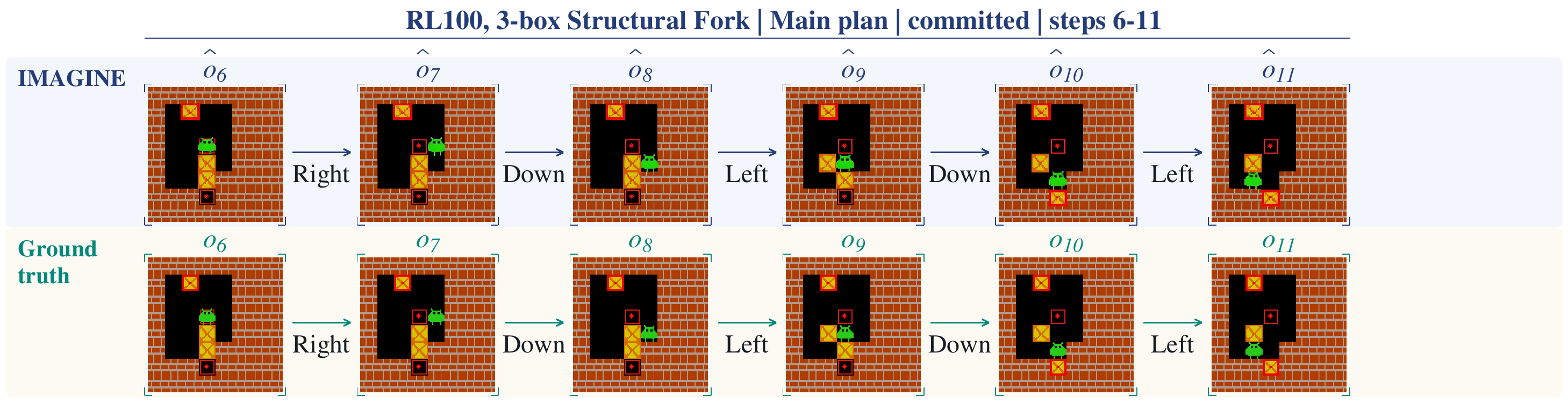}
\end{figure}

\begin{figure}[H]
    \centering
    \includegraphics[page=2,trim=0 308bp 0 808bp,clip,width=\linewidth]{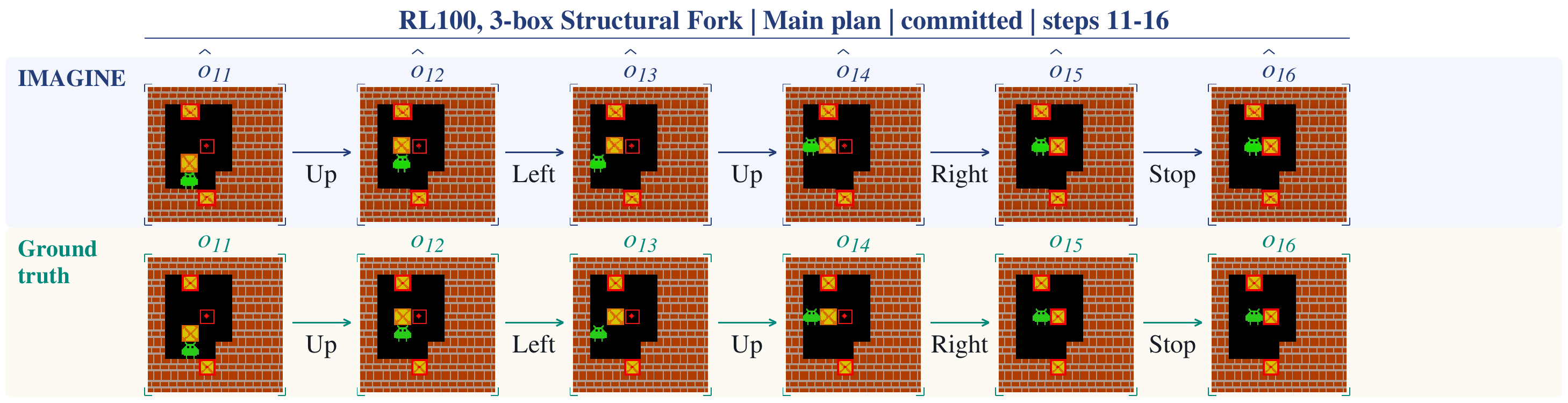}
\end{figure}

\begin{figure}[H]
    \centering
    \includegraphics[page=2,trim=0 0bp 0 1168bp,clip,width=\linewidth]{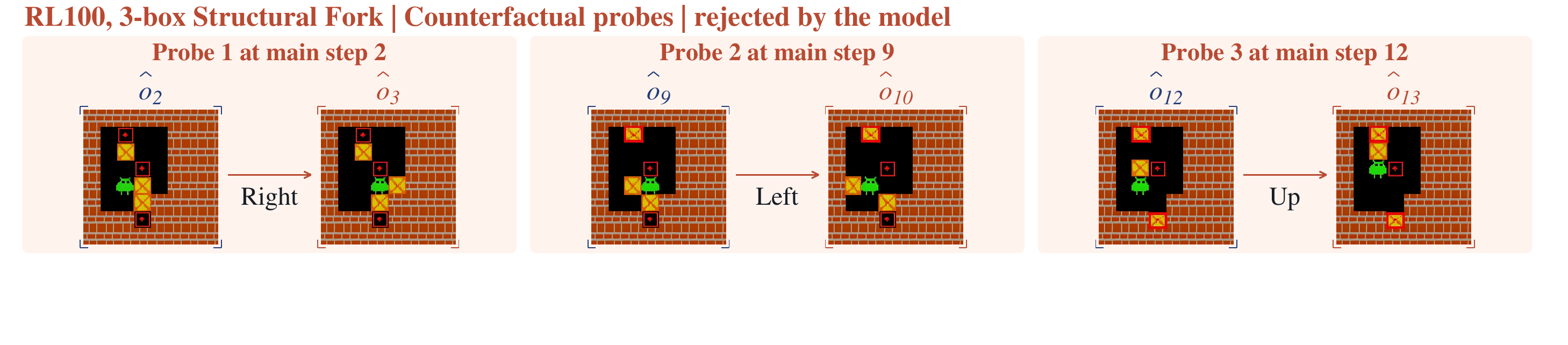}
    \caption{\textbf{Branching at RL100 on three-box Structural Fork.} Complete trace for episode 17; recorded execution ends in success.}
    \label{fig:rl_3fork_bra3k_rl100}
\end{figure}

\begin{figure}[H]
    \centering
    \includegraphics[page=3,trim=0 1028bp 0 0bp,clip,width=\linewidth]{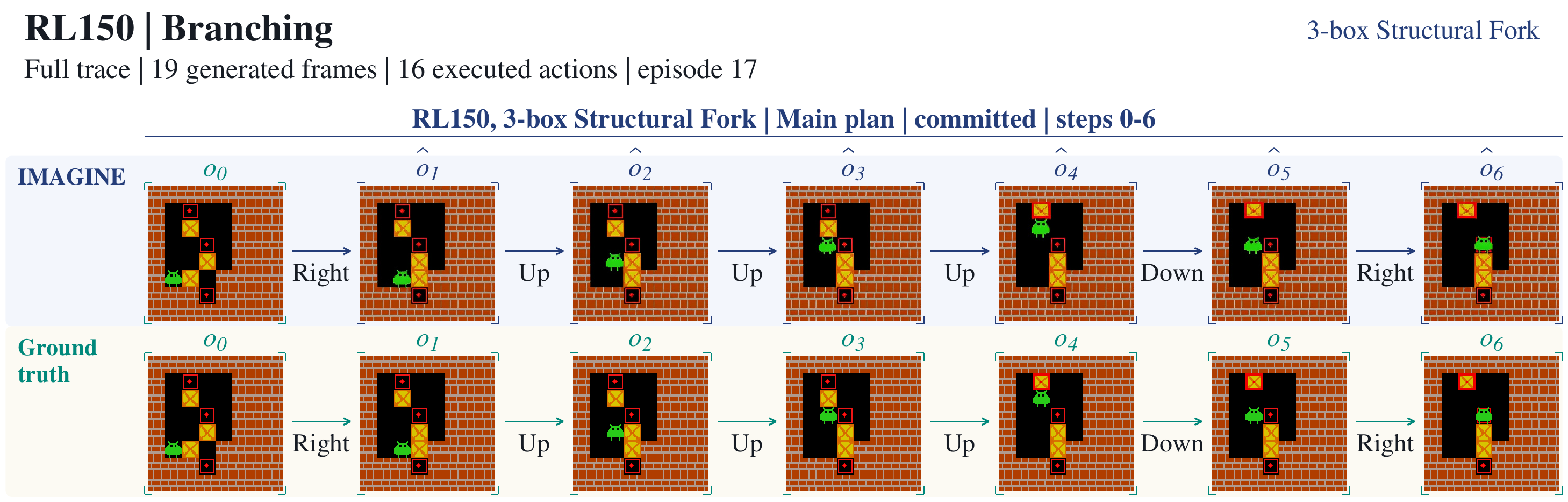}
\end{figure}

\begin{figure}[H]
    \centering
    \includegraphics[page=3,trim=0 668bp 0 448bp,clip,width=\linewidth]{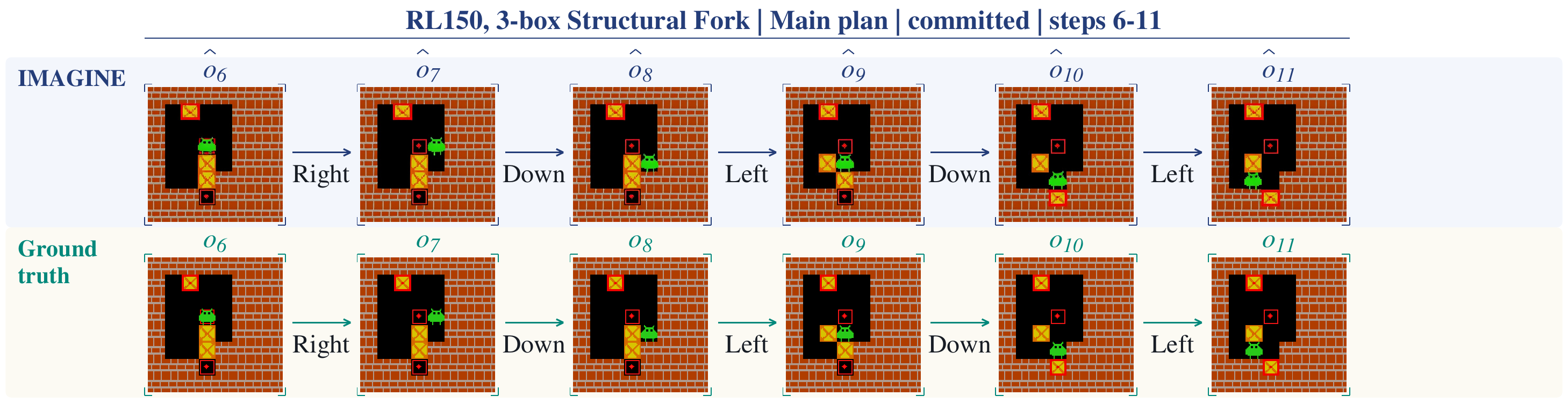}
\end{figure}

\begin{figure}[H]
    \centering
    \includegraphics[page=3,trim=0 308bp 0 808bp,clip,width=\linewidth]{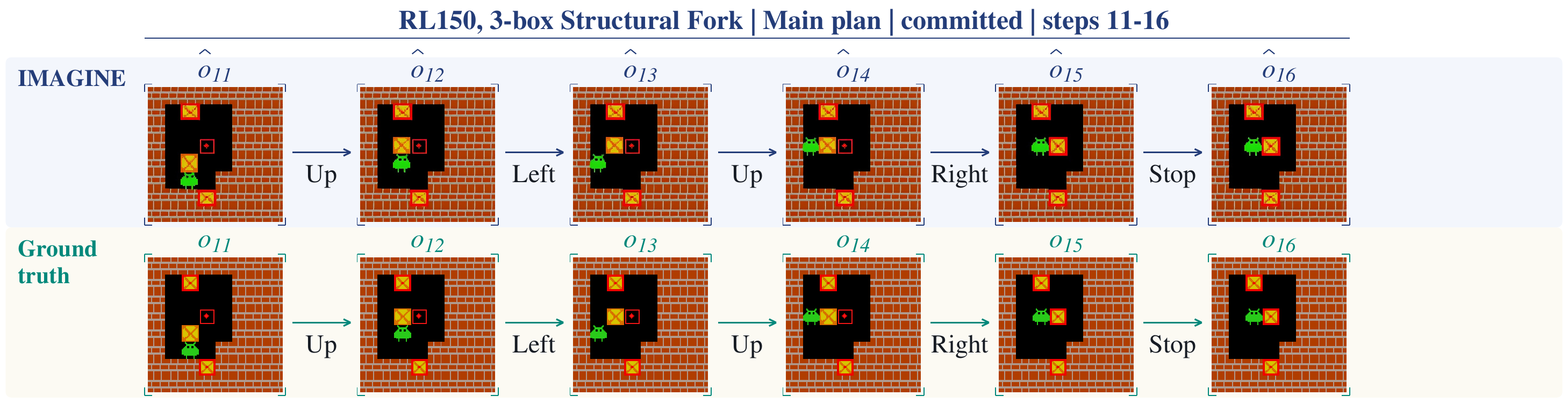}
\end{figure}

\begin{figure}[H]
    \centering
    \includegraphics[page=3,trim=0 0bp 0 1168bp,clip,width=\linewidth]{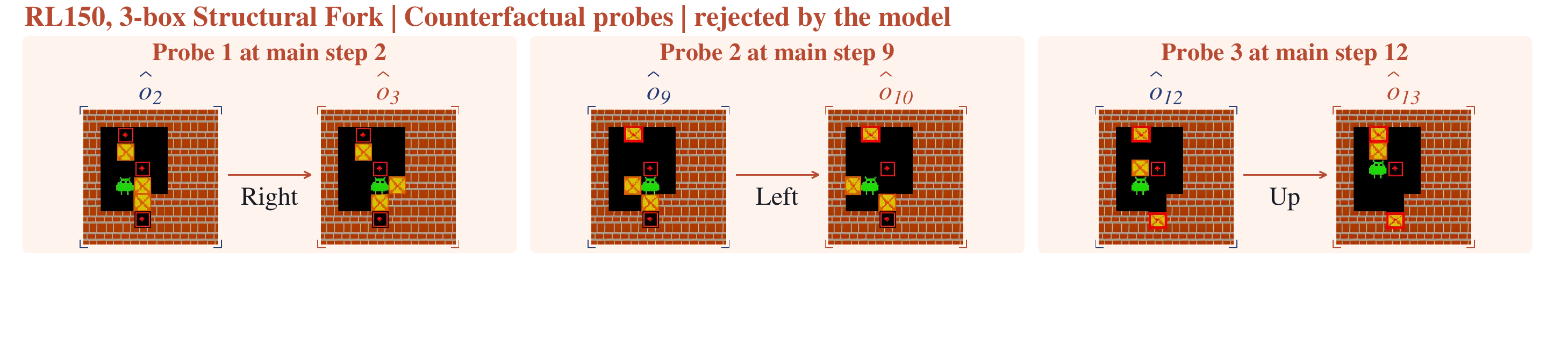}
    \caption{\textbf{Branching at RL150 on three-box Structural Fork.} Complete trace for episode 17; recorded execution ends in success.}
    \label{fig:rl_3fork_bra3k_rl150}
\end{figure}

\begin{figure}[H]
    \centering
    \includegraphics[page=4,trim=0 1028bp 0 0bp,clip,width=\linewidth]{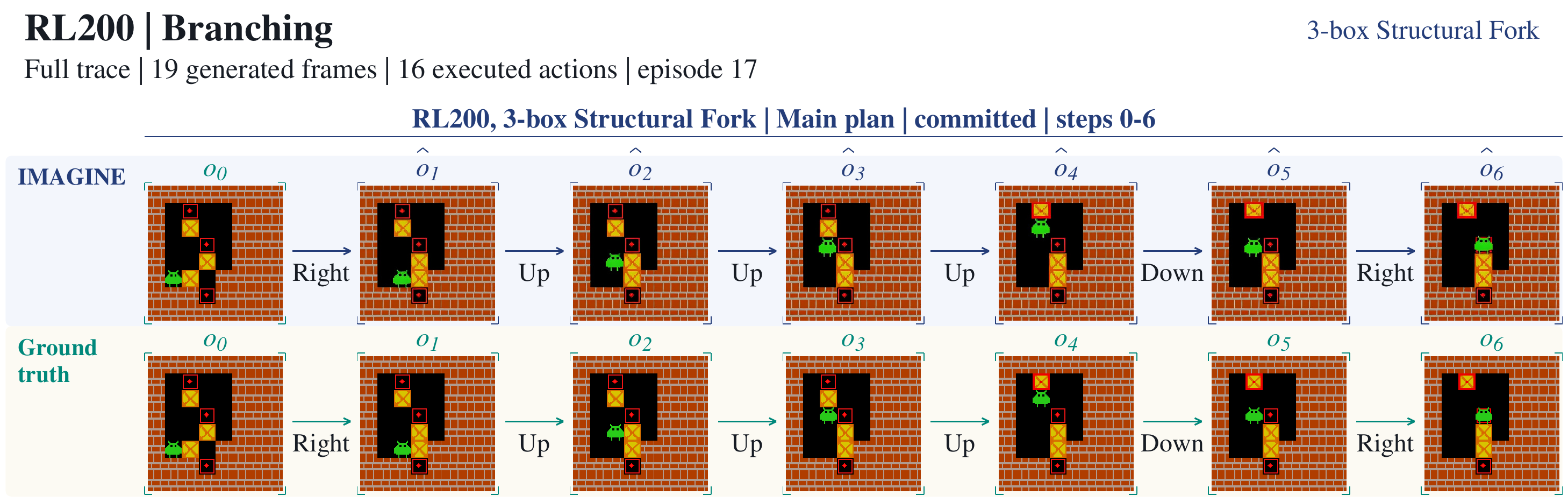}
\end{figure}

\begin{figure}[H]
    \centering
    \includegraphics[page=4,trim=0 668bp 0 448bp,clip,width=\linewidth]{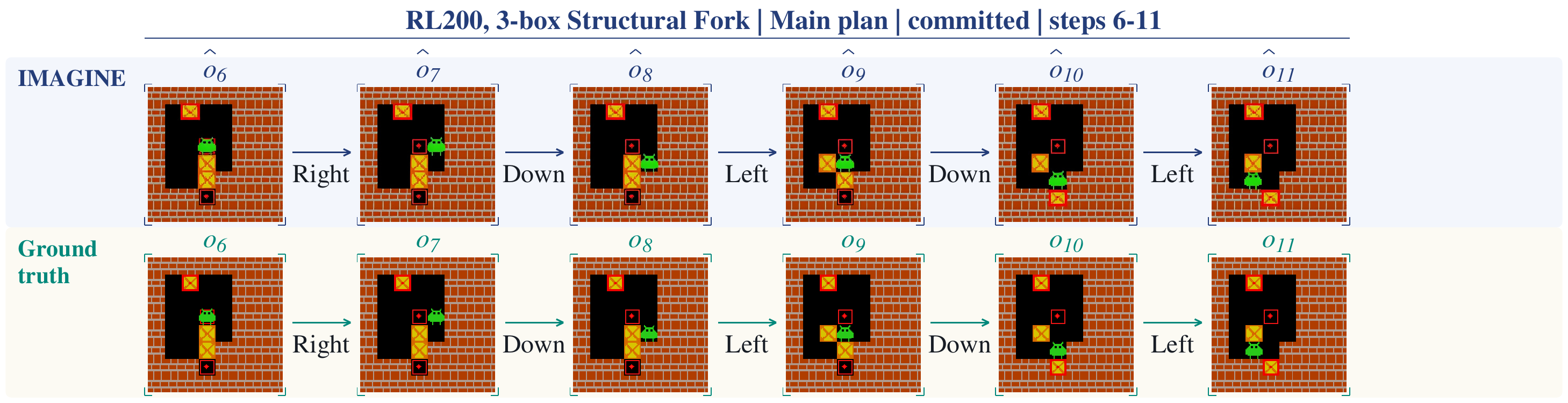}
\end{figure}

\begin{figure}[H]
    \centering
    \includegraphics[page=4,trim=0 308bp 0 808bp,clip,width=\linewidth]{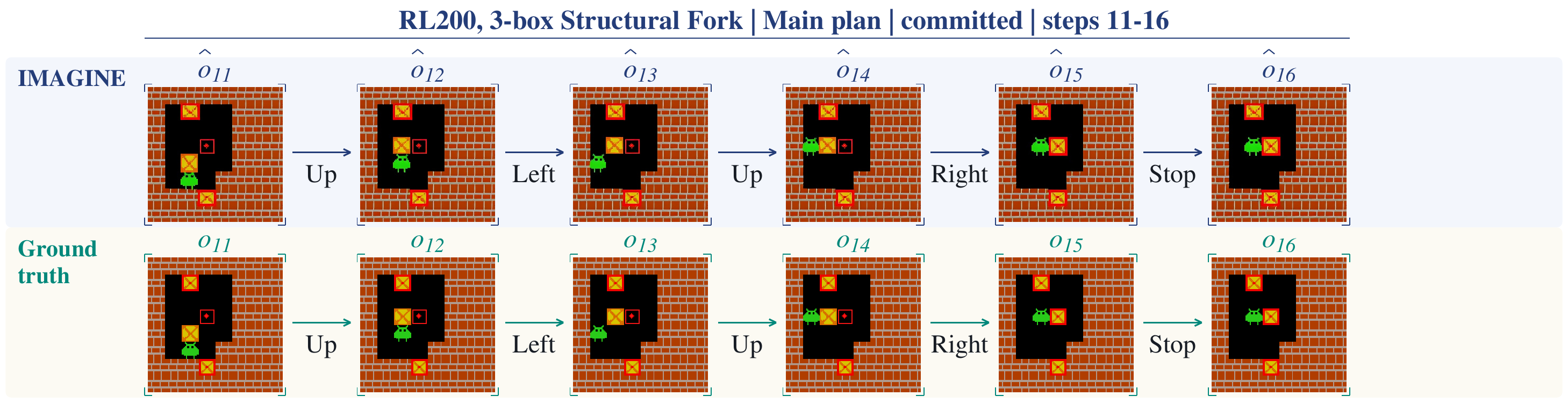}
\end{figure}

\begin{figure}[H]
    \centering
    \includegraphics[page=4,trim=0 0bp 0 1168bp,clip,width=\linewidth]{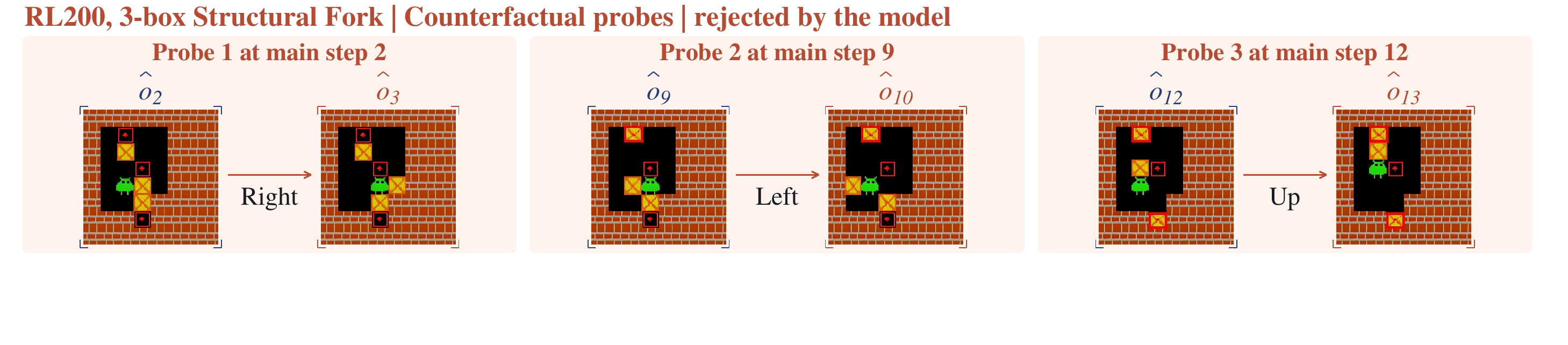}
    \caption{\textbf{Branching at RL200 on three-box Structural Fork.} Complete trace for episode 17; recorded execution ends in success.}
    \label{fig:rl_3fork_bra3k_rl200}
\end{figure}

\begin{figure}[H]
    \centering
    \includegraphics[page=1,trim=0 668bp 0 0bp,clip,width=\linewidth]{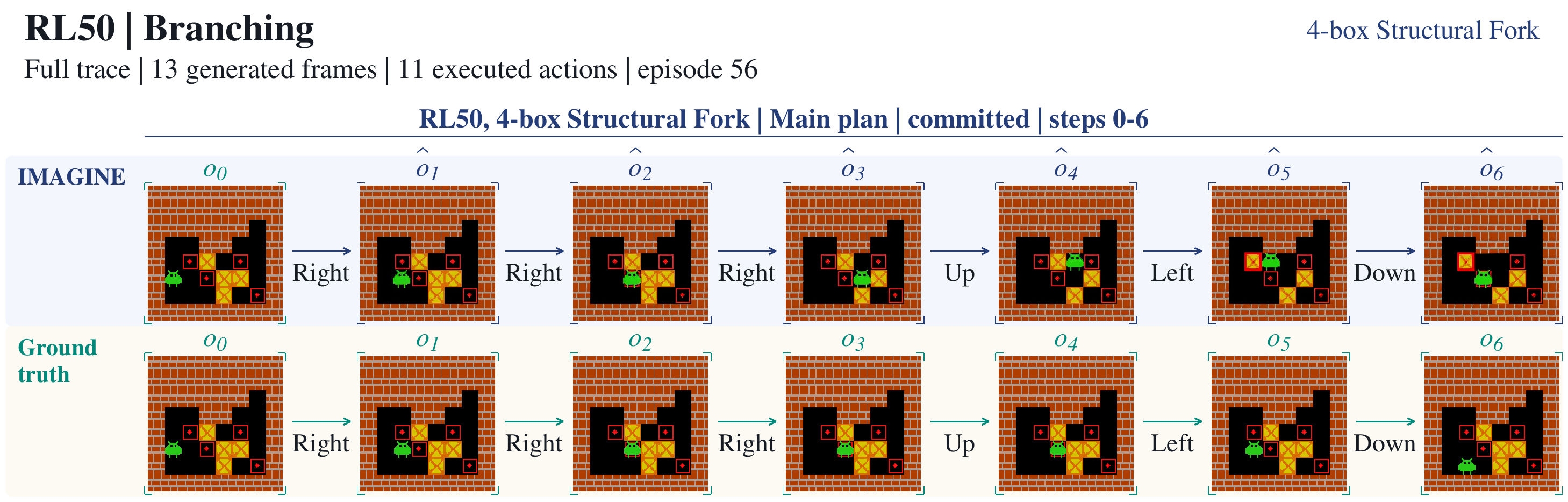}
\end{figure}

\begin{figure}[H]
    \centering
    \includegraphics[page=1,trim=0 308bp 0 448bp,clip,width=\linewidth]{figures/appendix_rl_4fork_traces.pdf}
\end{figure}

\begin{figure}[H]
    \centering
    \includegraphics[page=1,trim=0 0bp 0 808bp,clip,width=\linewidth]{figures/appendix_rl_4fork_traces.pdf}
    \caption{\textbf{Branching at RL50 on four-box Structural Fork.} Complete trace for episode 56; recorded execution ends in failure.}
    \label{fig:rl_4fork_bra3k_rl50}
\end{figure}

\begin{figure}[H]
    \centering
    \includegraphics[page=2,trim=0 668bp 0 0bp,clip,width=\linewidth]{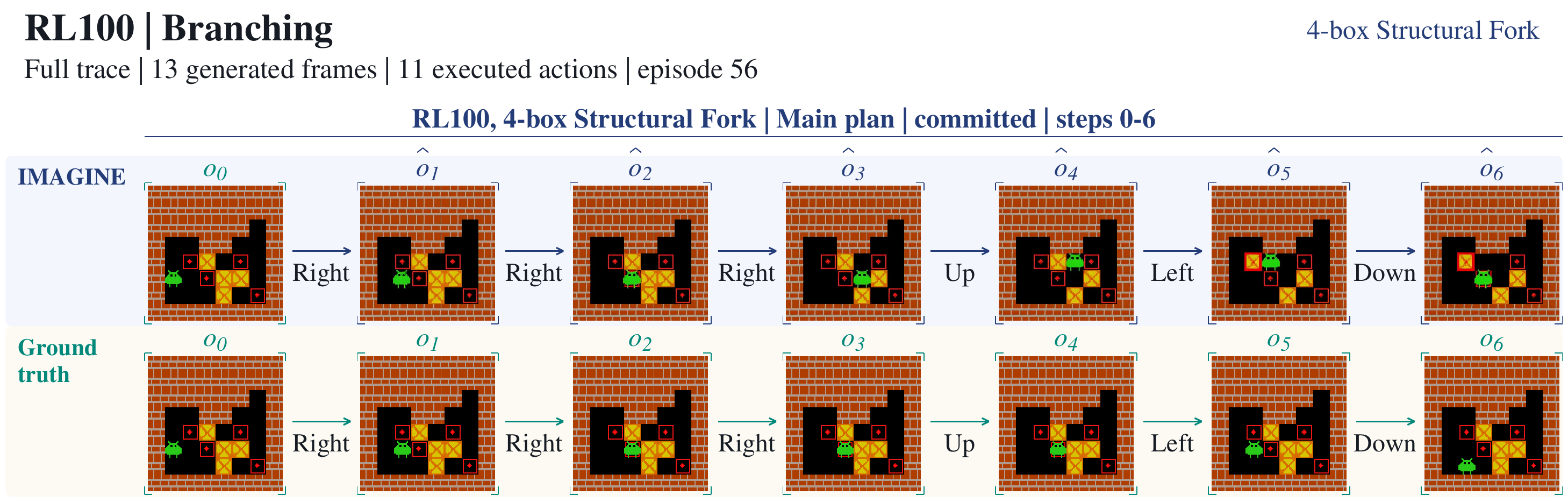}
\end{figure}

\begin{figure}[H]
    \centering
    \includegraphics[page=2,trim=0 308bp 0 448bp,clip,width=\linewidth]{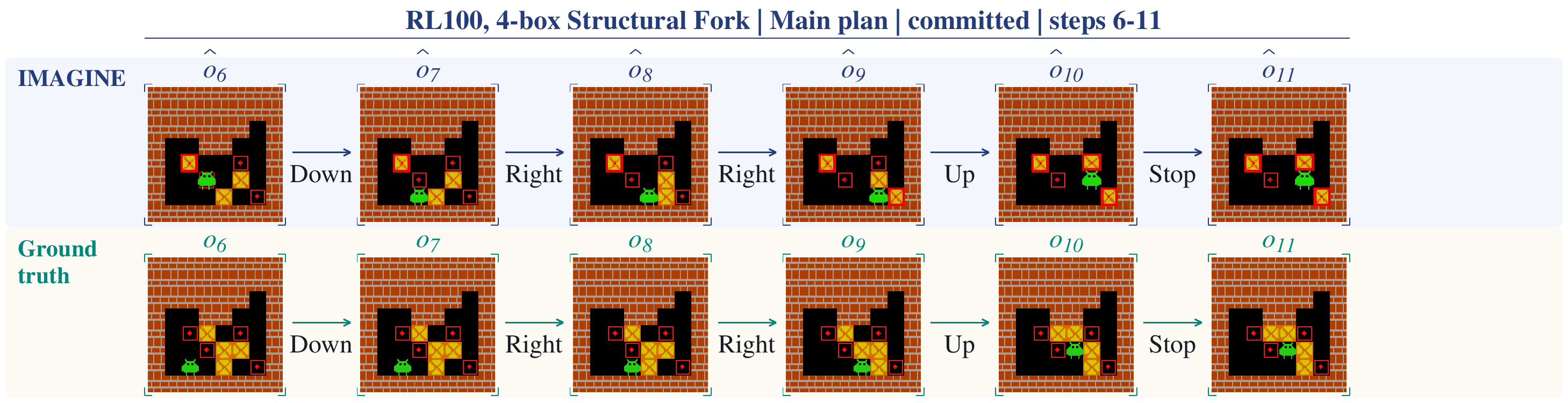}
\end{figure}

\begin{figure}[H]
    \centering
    \includegraphics[page=2,trim=0 0bp 0 808bp,clip,width=\linewidth]{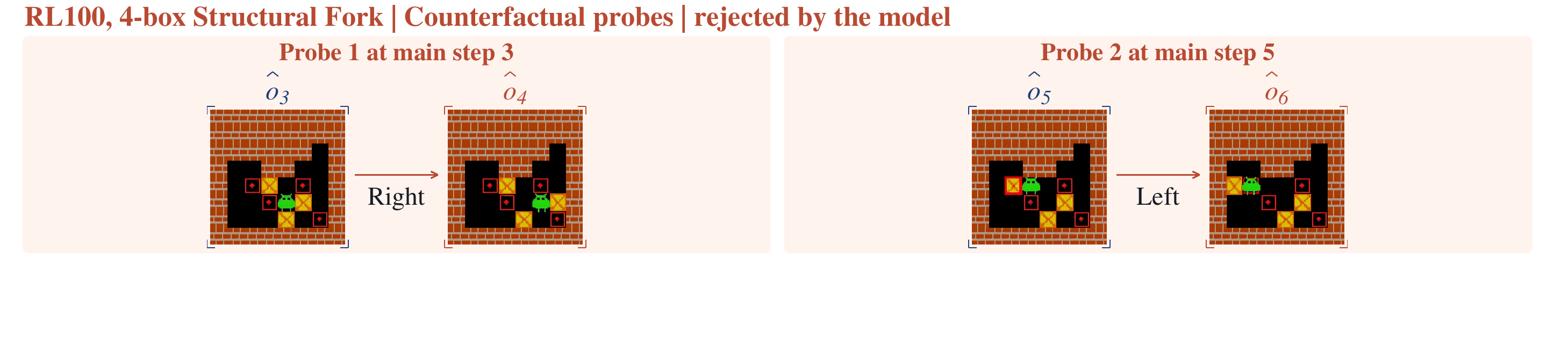}
    \caption{\textbf{Branching at RL100 on four-box Structural Fork.} Complete trace for episode 56; recorded execution ends in failure.}
    \label{fig:rl_4fork_bra3k_rl100}
\end{figure}

\begin{figure}[H]
    \centering
    \includegraphics[page=3,trim=0 668bp 0 0bp,clip,width=\linewidth]{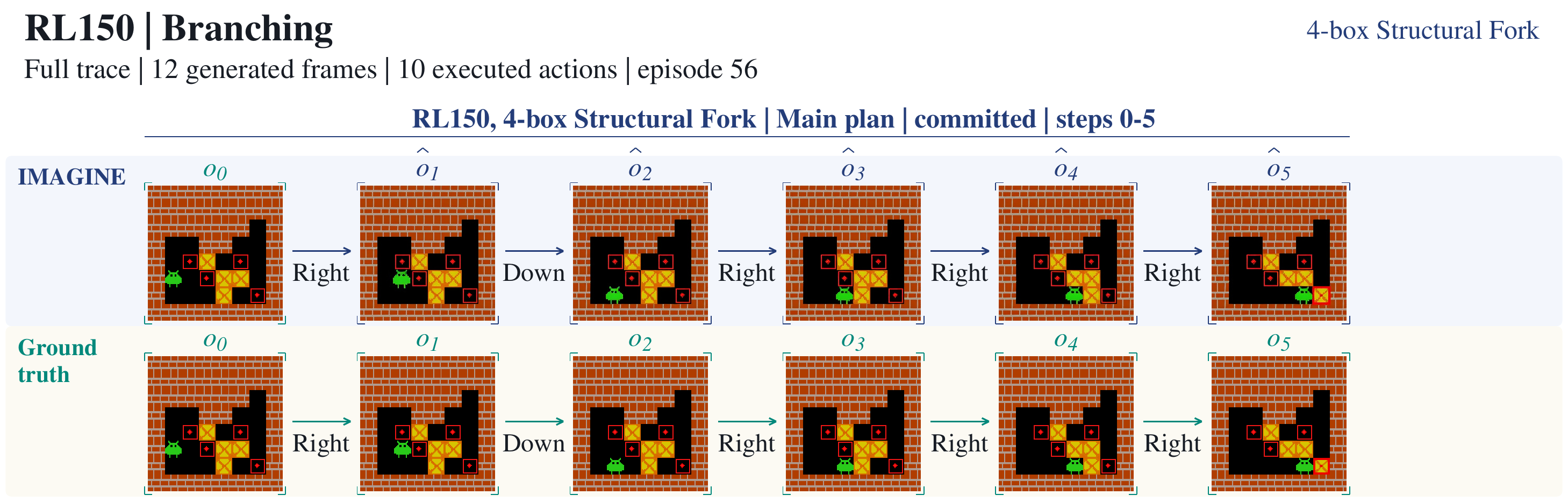}
\end{figure}

\begin{figure}[H]
    \centering
    \includegraphics[page=3,trim=0 308bp 0 448bp,clip,width=\linewidth]{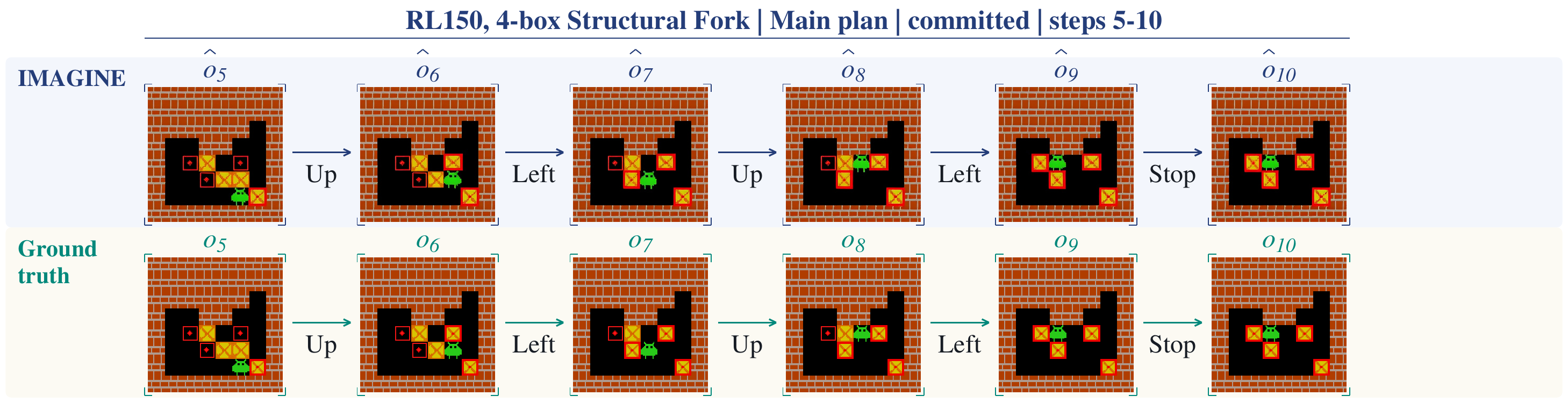}
\end{figure}

\begin{figure}[H]
    \centering
    \includegraphics[page=3,trim=0 0bp 0 808bp,clip,width=\linewidth]{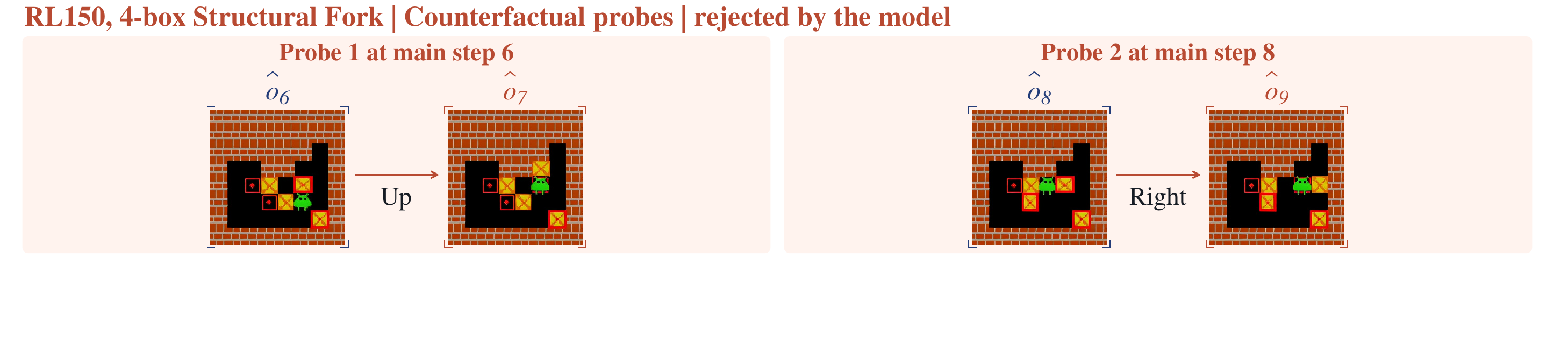}
    \caption{\textbf{Branching at RL150 on four-box Structural Fork.} Complete trace for episode 56; recorded execution ends in success.}
    \label{fig:rl_4fork_bra3k_rl150}
\end{figure}

\begin{figure}[H]
    \centering
    \includegraphics[page=4,trim=0 668bp 0 0bp,clip,width=\linewidth]{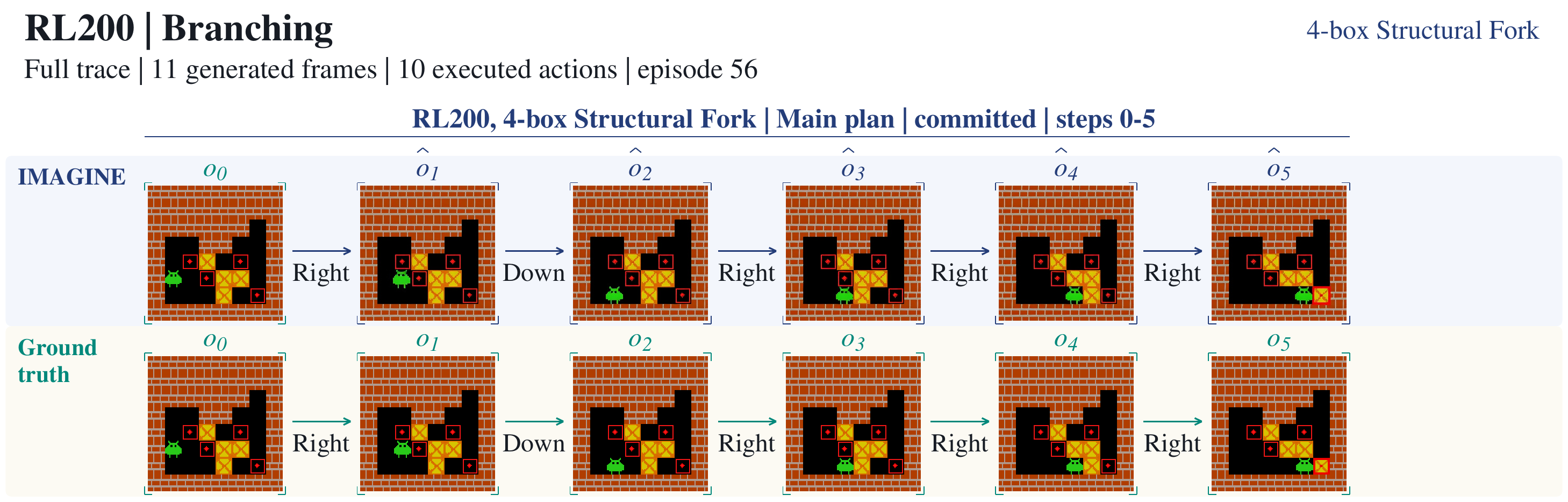}
\end{figure}

\begin{figure}[H]
    \centering
    \includegraphics[page=4,trim=0 308bp 0 448bp,clip,width=\linewidth]{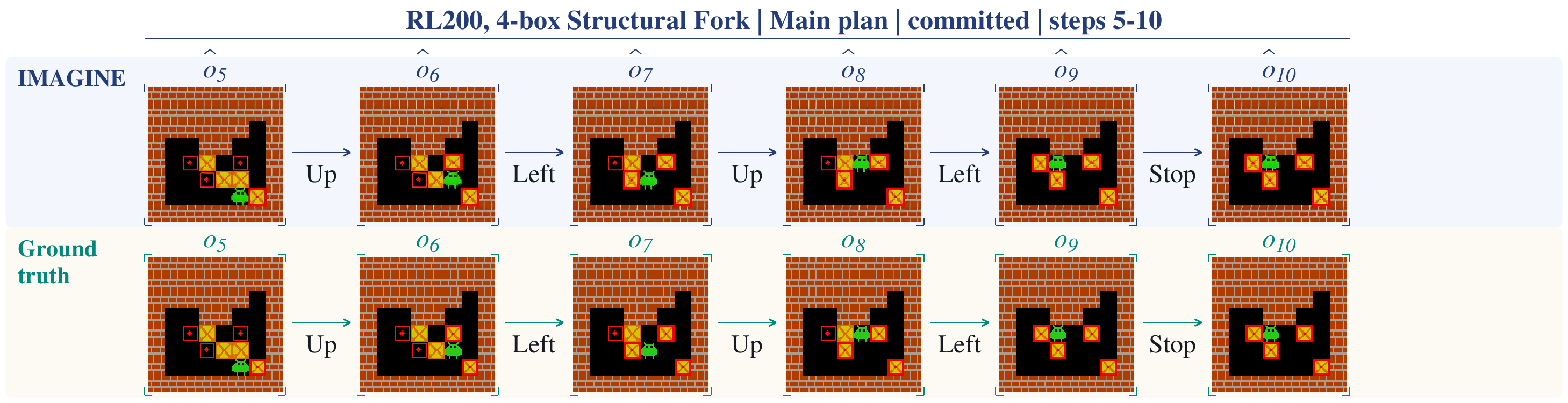}
\end{figure}

\begin{figure}[H]
    \centering
    \includegraphics[page=4,trim=0 0bp 0 808bp,clip,width=\linewidth]{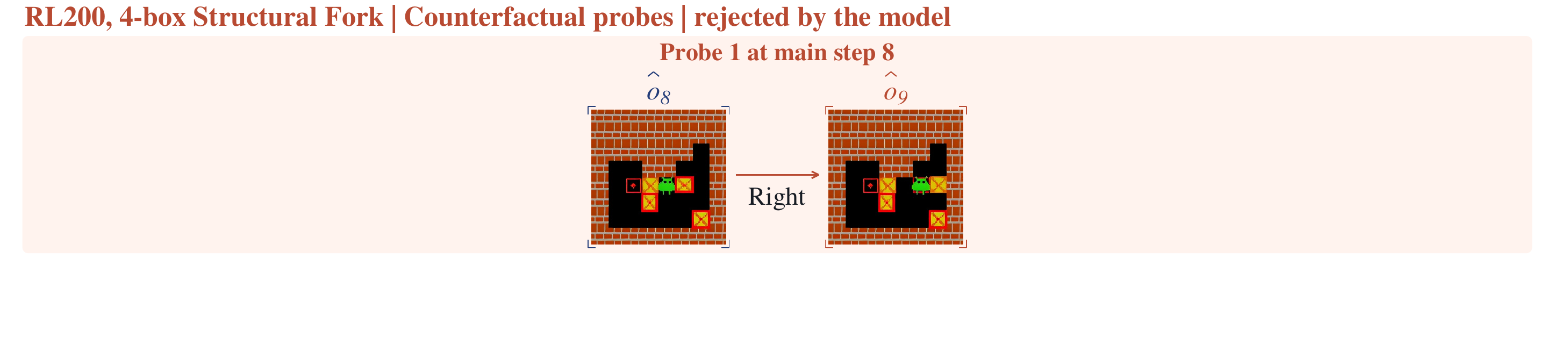}
    \caption{\textbf{Branching at RL200 on four-box Structural Fork.} Complete trace for episode 56; recorded execution ends in success.}
    \label{fig:rl_4fork_bra3k_rl200}
\end{figure}

\endgroup

\subsection{Failure Cases}
\label{app:failure_cases}
The following recorded $K=\infty$ episodes show failures in all four
environments: stopping before reaching the goal in Maze2D, leaving boxes
off their goals in Sokoban, being caught by the ghost in Pacman, and stopping
with insufficient target coverage in PushT. These examples use the same
task settings and checkpoints as the $K=\infty$ evaluations in
Section~\ref{sec:planning_evaluation}.

The real execution can also disagree with the model's outcome assessment.
For example, the RL50 Forward model describes its chosen plan as successful
in Figure~\ref{fig:rl_3dead_fwd3k_rl50}, but the executed plan fails.
In the Pacman example below, execution ends at the collision. Later predicted
frames remain visible, with the corresponding real states marked as not executed.

\begingroup
\setlength{\intextsep}{2pt plus 1pt minus 1pt}
\setlength{\abovecaptionskip}{4pt}
\begin{figure}[H]
    \centering
    \includegraphics[page=1,trim=0 438bp 0 0bp,clip,width=0.96\linewidth]{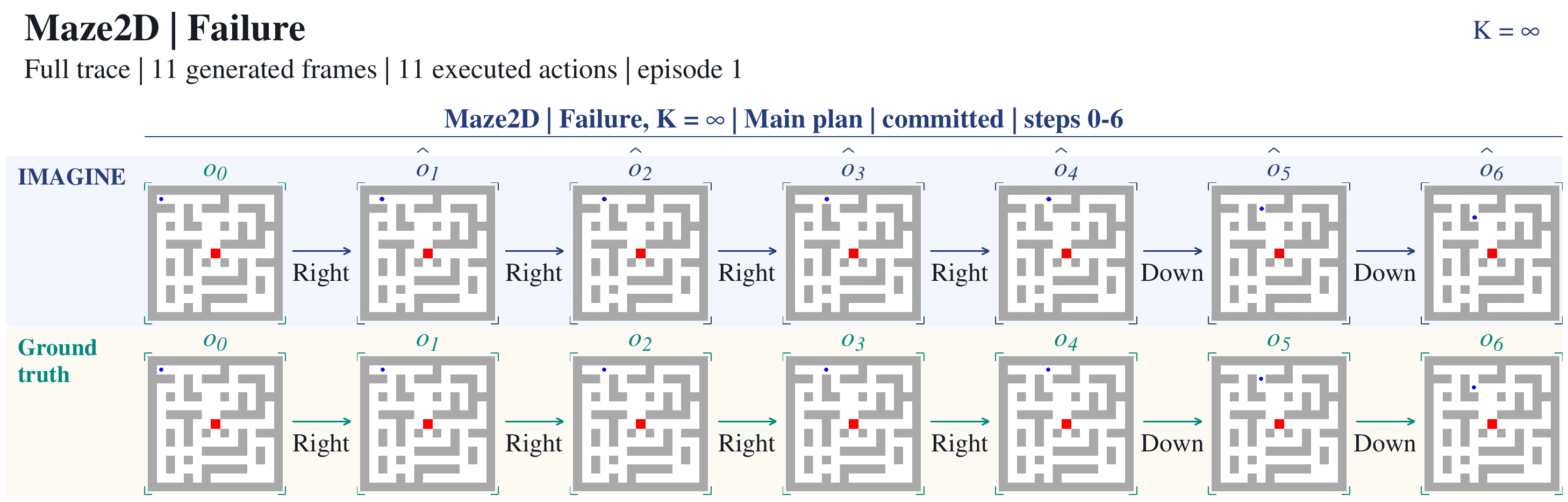}
\end{figure}

\begin{figure}[H]
    \centering
    \includegraphics[page=1,trim=0 0bp 0 448bp,clip,width=0.96\linewidth]{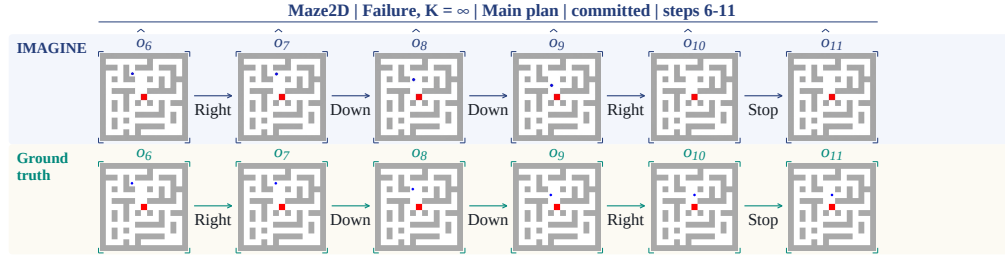}
    \caption{\textbf{Maze2D failure at $K=\infty$.} The agent issues \texttt{STOP} before reaching the goal.}
    \label{fig:failure_maze2d}
\end{figure}

\begin{figure}[H]
    \centering
    \includegraphics[page=2,trim=0 798bp 0 0bp,clip,width=0.96\linewidth]{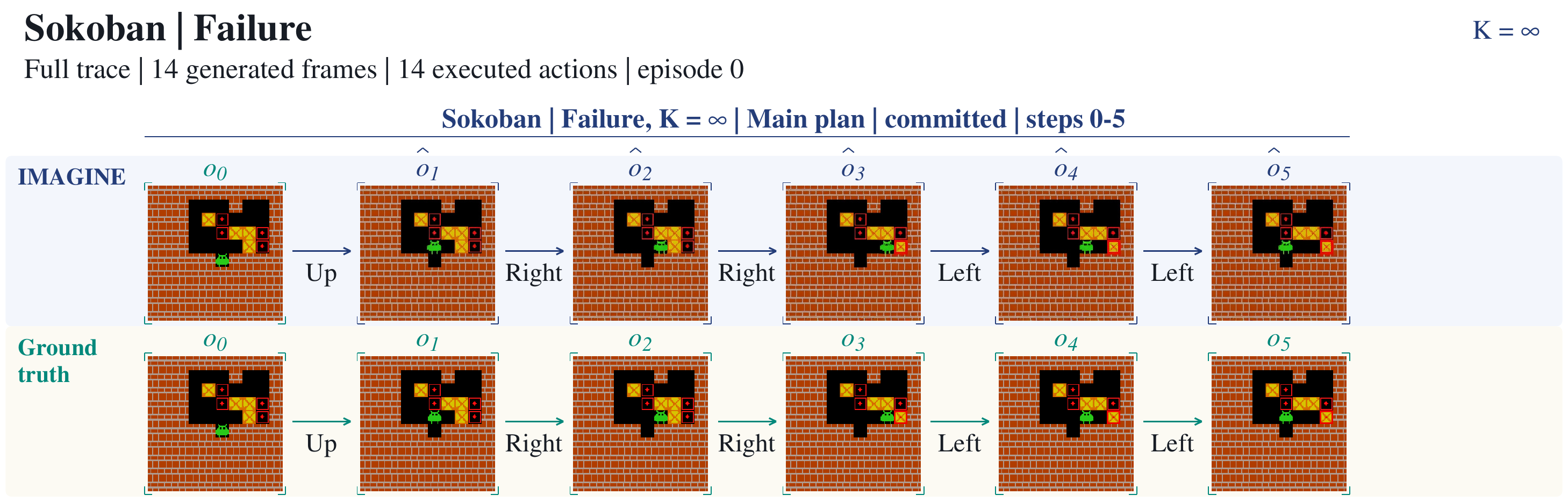}
\end{figure}

\begin{figure}[H]
    \centering
    \includegraphics[page=2,trim=0 438bp 0 448bp,clip,width=0.96\linewidth]{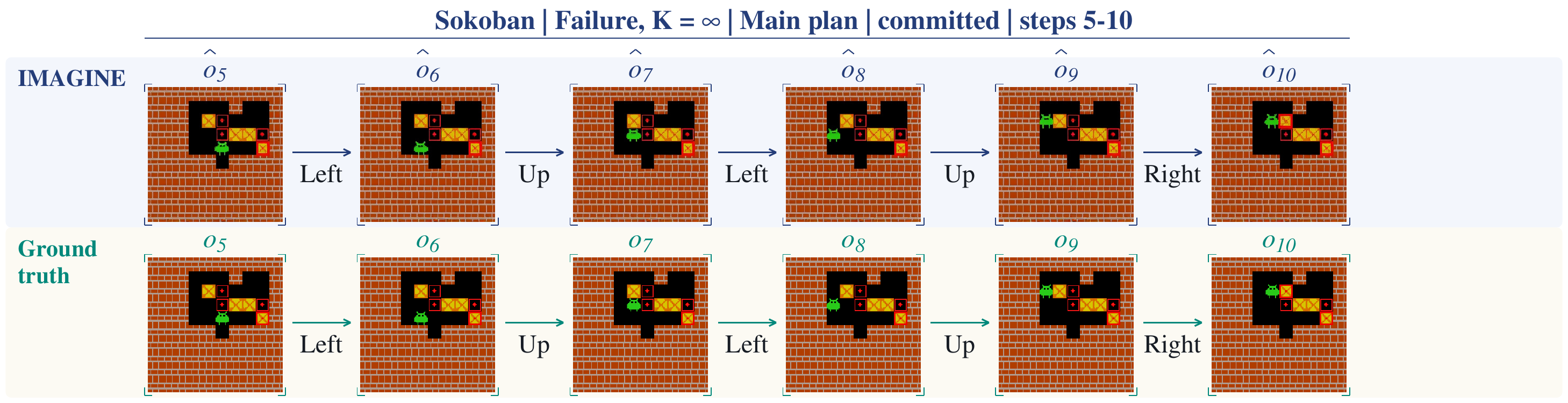}
\end{figure}

\begin{figure}[H]
    \centering
    \includegraphics[page=2,trim=0 0bp 0 808bp,clip,width=0.96\linewidth]{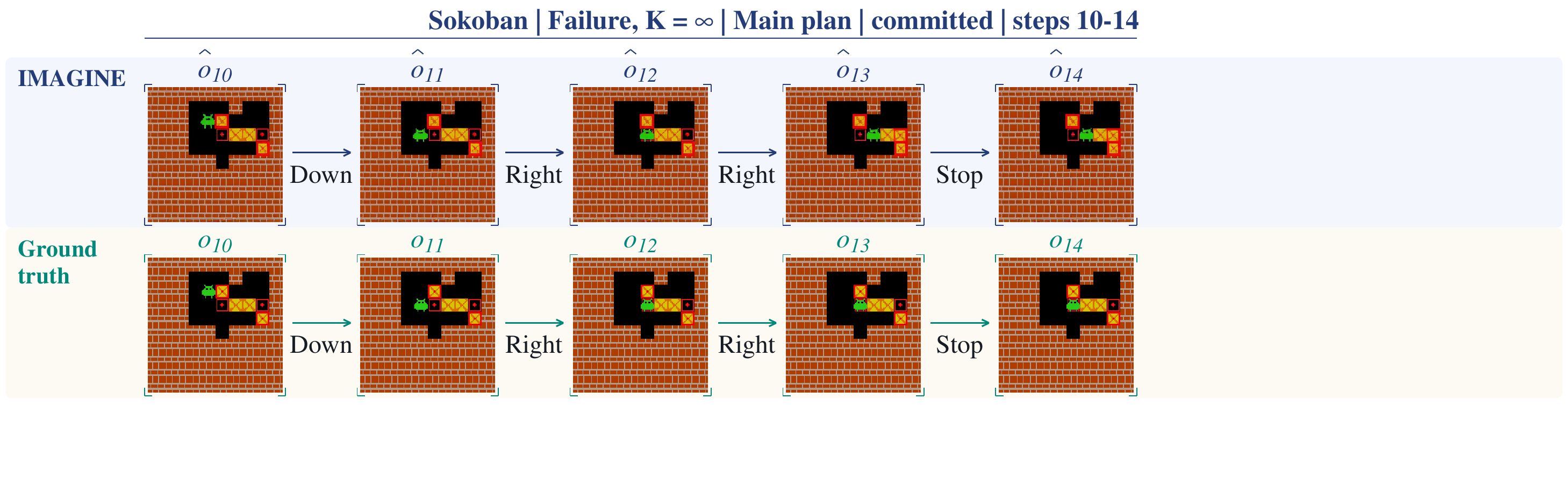}
    \caption{\textbf{Sokoban failure at $K=\infty$.} The agent stops with 2 of 4 boxes on goals.}
    \label{fig:failure_sokoban}
\end{figure}

\begin{figure}[H]
    \centering
    \includegraphics[page=3,trim=0 1158bp 0 0bp,clip,width=0.96\linewidth]{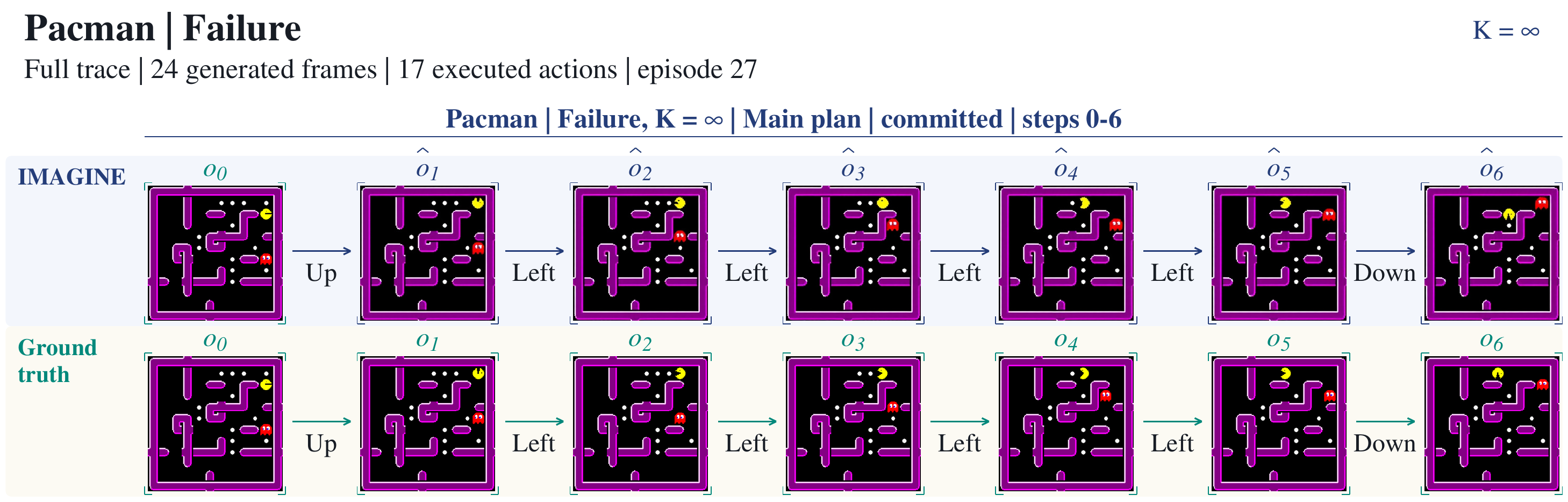}
\end{figure}

\begin{figure}[H]
    \centering
    \includegraphics[page=3,trim=0 798bp 0 448bp,clip,width=0.96\linewidth]{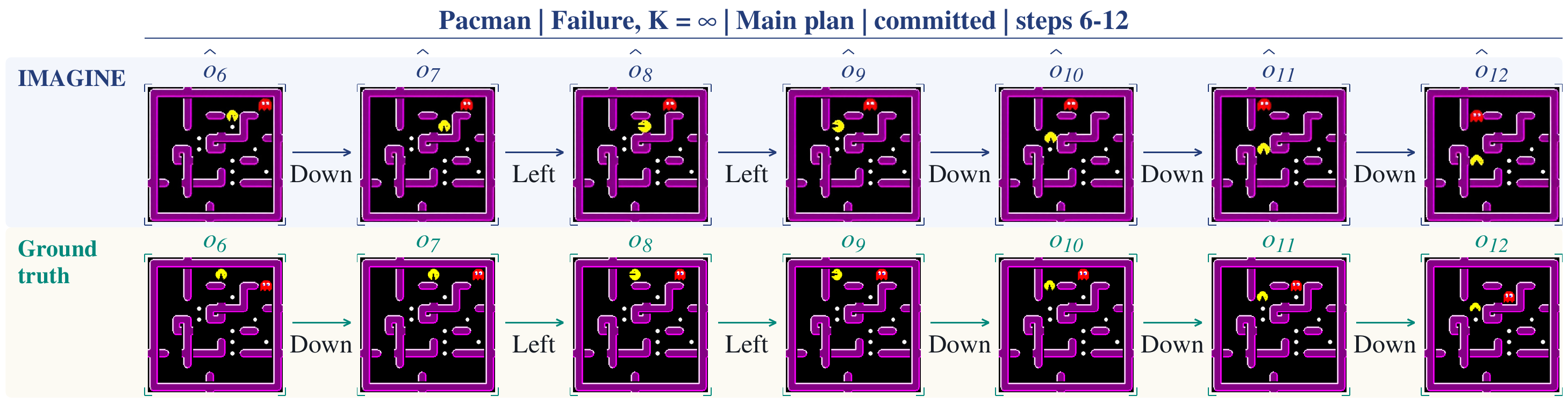}
\end{figure}

\begin{figure}[H]
    \centering
    \includegraphics[page=3,trim=0 438bp 0 808bp,clip,width=0.96\linewidth]{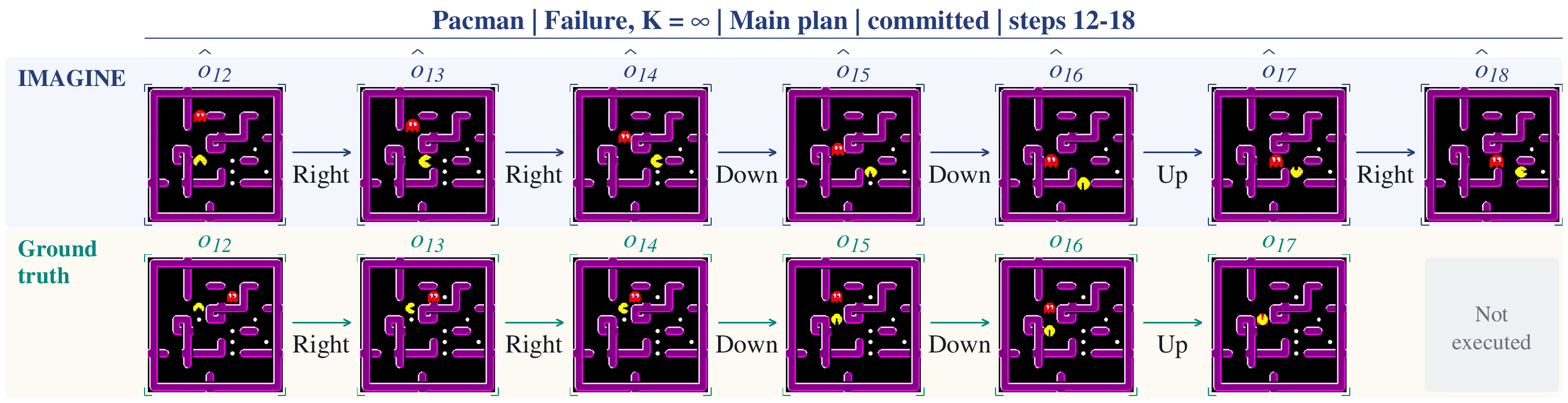}
\end{figure}

\begin{figure}[H]
    \centering
    \includegraphics[page=3,trim=0 0bp 0 1168bp,clip,width=0.96\linewidth]{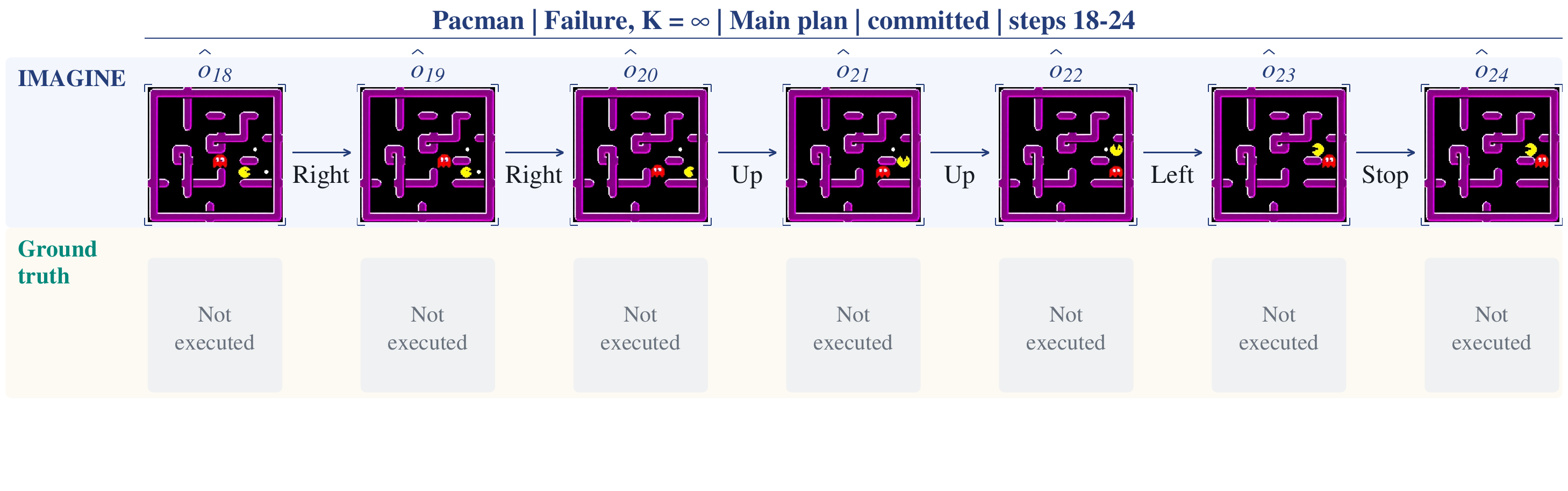}
    \caption{\textbf{Pacman failure at $K=\infty$.} The ghost catches Pacman at action 17. The remaining seven predicted states are never executed.}
    \label{fig:failure_pacman}
\end{figure}

\begin{figure}[H]
    \centering
    \includegraphics[page=4,trim=0 798bp 0 0bp,clip,width=0.96\linewidth]{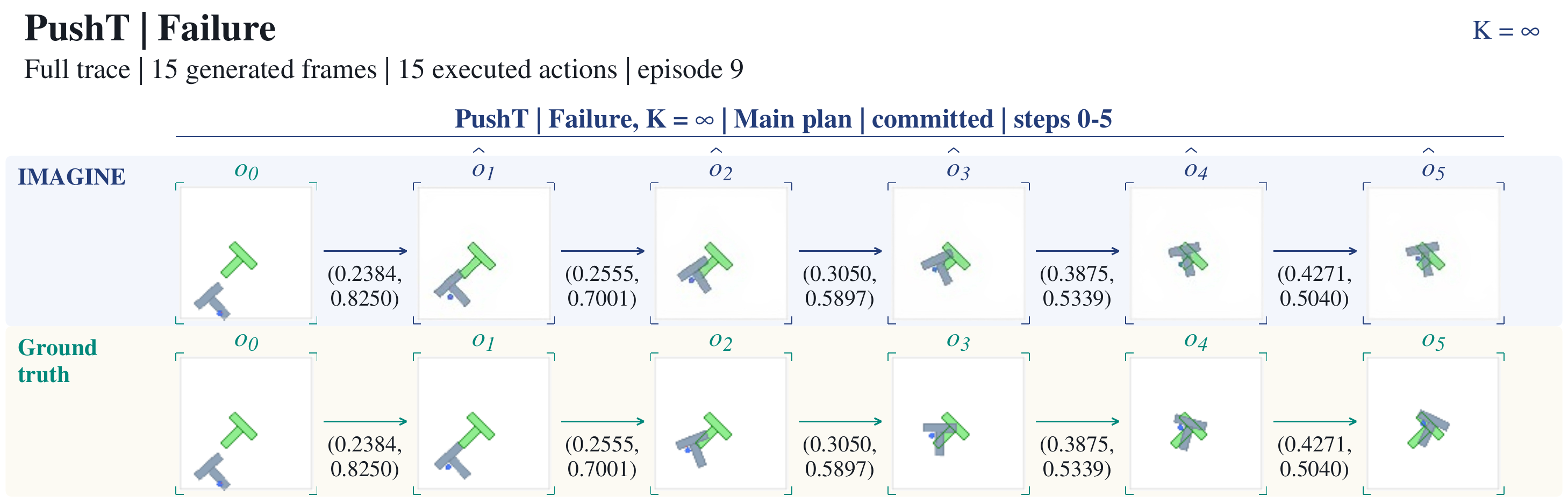}
\end{figure}

\begin{figure}[H]
    \centering
    \includegraphics[page=4,trim=0 438bp 0 448bp,clip,width=0.96\linewidth]{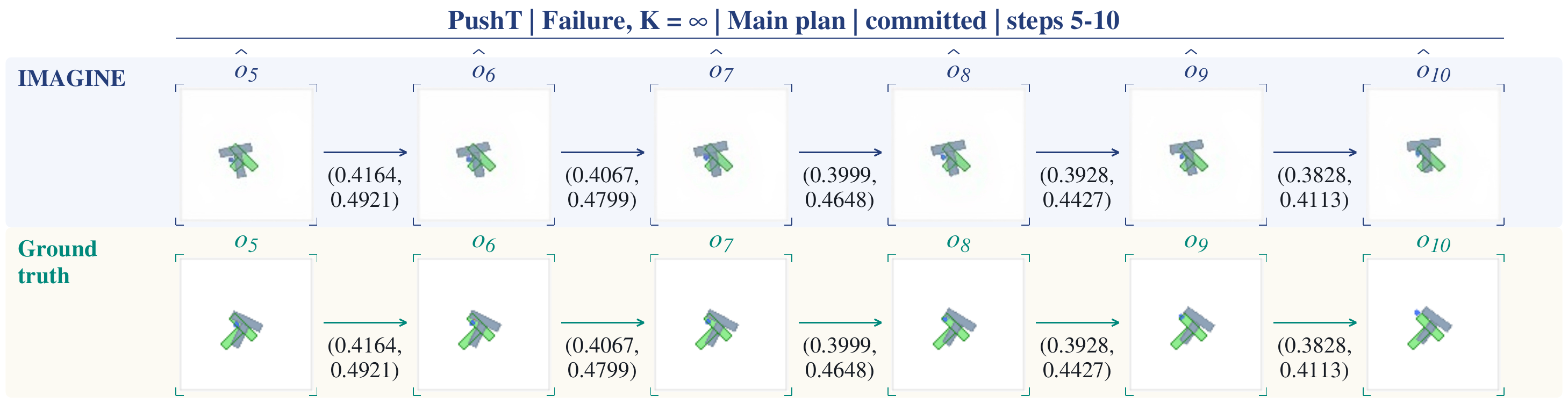}
\end{figure}

\begin{figure}[H]
    \centering
    \includegraphics[page=4,trim=0 0bp 0 808bp,clip,width=0.96\linewidth]{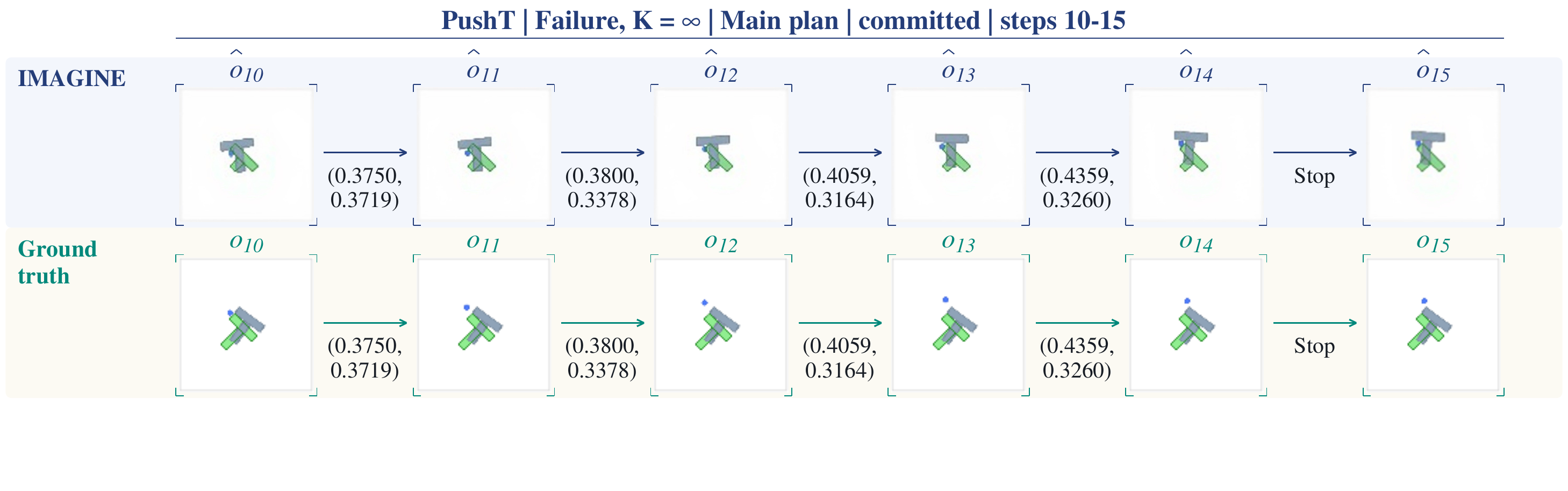}
    \caption{\textbf{PushT failure at $K=\infty$.} The agent stops with 32.4\% final target coverage, below the paper's $80\%$ success threshold.}
    \label{fig:failure_pusht}
\end{figure}

\endgroup

\end{document}